\documentclass{article}
\usepackage[a4paper, total={6in, 9in}]{geometry}
\usepackage{graphicx}
\usepackage[numbers]{natbib}
\usepackage{caption}
\usepackage{amsfonts}
\usepackage{multirow}
\usepackage{booktabs}
\usepackage{bbm}
\usepackage{xcolor}
\usepackage{subcaption}
\usepackage{cleveref}
\usepackage{adjustbox}
\usepackage{authblk}
\usepackage{chngcntr}
\usepackage{wrapfig}

\title{Multi-Label Contrastive Learning for Abstract Visual Reasoning}
\author{
Miko\l aj Ma\l ki\'nski, Jacek Ma\'ndziuk
}
\affil{
Faculty of Mathematics and Information Science \\
Warsaw University of Technology, Warsaw, Poland\\
mikolaj.malkinski@gmail.com, mandziuk@mini.pw.edu.pl
}
\date{}

\begin{document}

    \maketitle

    \begin{abstract}
        For a long time the ability to solve abstract reasoning tasks was considered one of the hallmarks of human intelligence. Recent advances in application of deep learning (DL) methods led, as in many other domains, to surpassing human abstract reasoning performance, specifically in the most popular type of such problems - the Raven's Progressive Matrices (RPMs). While the efficacy of DL systems is indeed impressive, the way they approach the RPMs is very different from that of humans. State-of-the-art systems solving RPMs rely on massive pattern-based training and sometimes on exploiting biases in the dataset, whereas humans concentrate on identification of the rules / concepts underlying the RPM (or generally a visual reasoning task) to be solved. Motivated by this cognitive difference, this work aims at combining DL with human way of solving RPMs and getting the best of both worlds. Specifically, we cast the problem of solving RPMs into multi-label classification framework where each RPM is viewed as a multi-label data point, with labels determined by the set of abstract rules underlying the RPM. For efficient training of the system we introduce a generalisation of the Noise Contrastive Estimation algorithm to the case of multi-label samples. Furthermore, we propose a new sparse rule encoding scheme for RPMs which, besides the new training algorithm, is the key factor contributing to the state-of-the-art performance. The proposed approach is evaluated on two most popular benchmark datasets (Balanced-RAVEN and PGM) and on both of them demonstrates an advantage over the current state-of-the-art results. Contrary to applications of contrastive learning methods reported in other domains, the state-of-the-art performance reported in the paper is achieved with no need for large batch sizes or strong data augmentation.
    \end{abstract}

    \section{Introduction}
\begin{figure}[t]
        \centering
        \includegraphics[width=0.3\columnwidth]{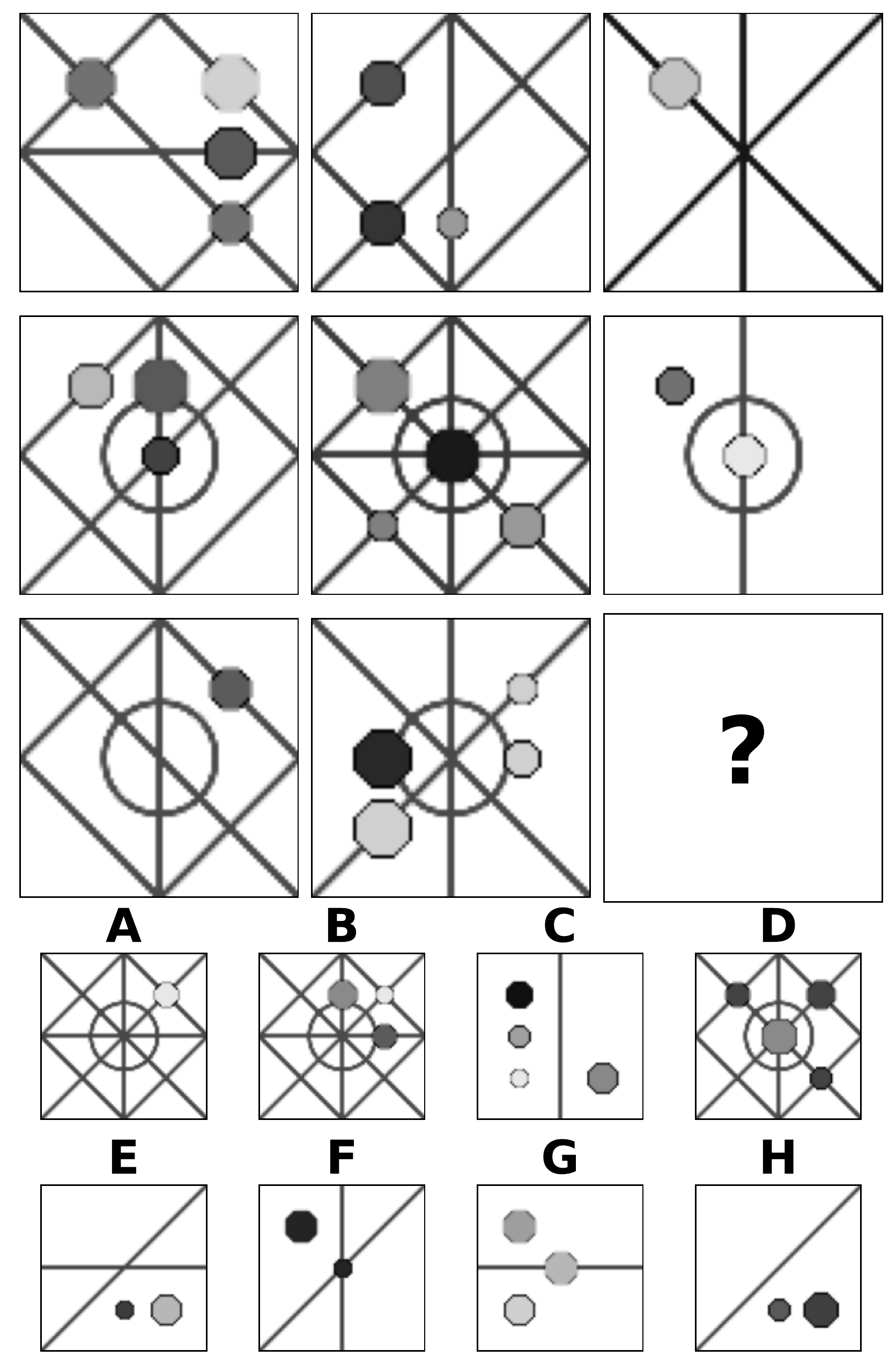}
        ~
        ~
        \includegraphics[width=0.3\columnwidth]{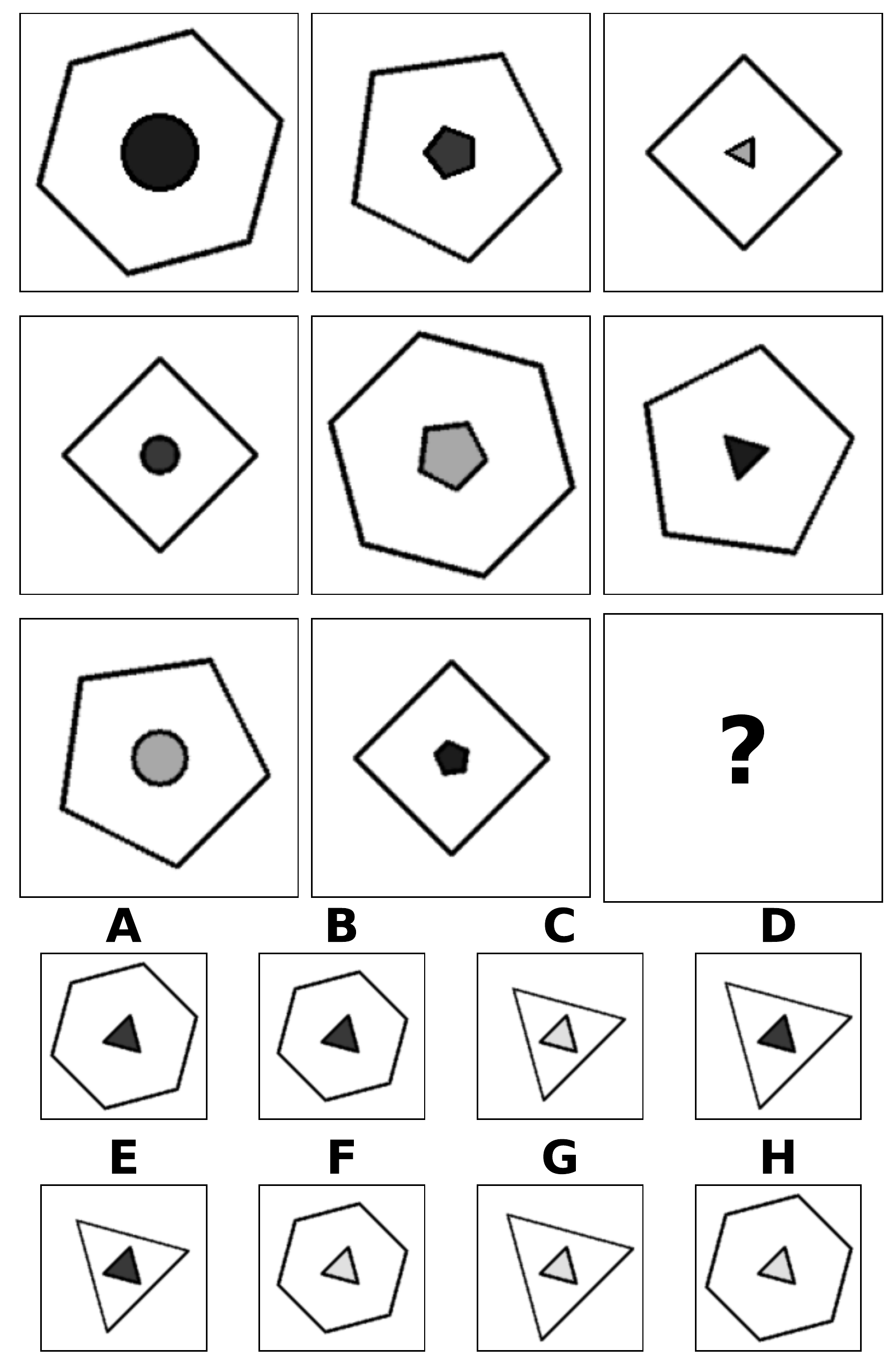}
        \caption{
        Solving RPMs requires to identify abstract relationships hidden behind random visual distractors and contrast possible answers to select the one which fits best.
        Left RPM, although has more visual details, is governed by only a single rule (row-wise AND applied to shape position), whereas perceptually simpler right RPM contains 8 distinct relations applied to both outer and inner structures. The examples come from the PGM and Balanced-RAVEN datasets, respectively and in both cases the correct answer, which is to be placed in the bottom-right panel, is A.
        }
        \label{fig:rpm}
    \end{figure}

    Abstract visual reasoning tasks are considered a widely-accepted way of measuring human intelligence.
    The most popular example of such task are Raven's Progressive Matrices (RPMs)~\cite{raven1936mental,raven1998raven}, where one is required to identify abstract relations between visually simple objects and their attributes (see Figure~\ref{fig:rpm}).
    The importance of RPMs in measuring human intelligence is justified by the fact that their solving requires an incremental strategy for inducing regularities in each problem~\cite{carpenter1990one}.
    Previous works identified a major gap between the performance of Machine Learning (ML) algorithms and humans~\cite{santoro2018measuring,Zhang_2019_CVPR}, which sparked interest in these problems within the deep learning (DL) community.
    In effect, human performance advantage has quickly vanished along with development of recent DL models~\cite{wu2020scattering}.

    However, the way ML algorithms solve RPMs leaves something to be desired.
    In order to solve these tasks, humans are required to come up with a strategy which correctly identifies all the underlying rules and differentiates them from distracting features, while the goal of the vast majority of ML approaches is to focus on selecting a correct answer, without explicit explanation for the output.

    Such a direct optimization problem formulation encourages neural models to simplify the solution process and rely on biases instead of understanding the internal problem structure.
    Indeed, such pathology was identified for the RAVEN dataset~\cite{Zhang_2019_CVPR}, where networks were able to arrive at a correct answer just by comparing the set of possible choices~\cite{hu2020hierarchical}.
    Similar problems were noted throughout the visual reasoning literature~\cite{zhang2016yin,johnson2017clevr}.

    A large body of cognitive literature identified contrastive mechanisms and the ability to create analogies as key ingredients for adaptive problem solving~\cite{gentner1983structure,hofstadter1995fluid,smith2014role}, which allow to apply previous experiences to novel domains.
    Naturally, there have been various approaches to replicate such behaviour in neural models either in the form of an explicit contrastive module~\cite{zhang2019learning} or through the way the data is presented during training~\cite{hill2018learning}, which in both cases resulted in notable improvements.
    Moreover, it was demonstrated that when models were additionally trained to predict a symbolic explanation for their answers (so-called \emph{auxiliary training}), their generalisation capabilities increased substantially~\cite{santoro2018measuring}.
    Nonetheless, despite notable positive impact of the auxiliary training, the topic remains underexplored.

    \paragraph{Motivation and Contribution.}
    In order to build models which \emph{understand} how to solve RPMs, we look for alternative training approaches.
    Encouraged by the effectiveness of both contrastive learning methods and symbolic explanations, we seek to develop a novel auxiliary training method for abstract visual reasoning tasks.
    We aim to exploit two recurring themes present in human approaches: the contrastive mechanism which differentiates between correct and wrong answers to the RPMs and the inherent ability to first identify all the abstract relations defining a given RPM and then use them to select the answer.
    The main contribution of this work is four-fold:
    \begin{itemize}
        \item Approaching the problem of solving RPMs by casting it into a multi-label classification framework, where labels are determined by the underlying abstract rules. This viewpoint allows us to explore a novel training method for abstract visual reasoning tasks.
        \item Devising a new formulation of the Noise Contrastive Estimation (NCE) learning algorithm for the case of multi-label samples, whose application imitates human approach to solving RPMs.
        \item Proposition of a new sparse rule encoding scheme for RPMs which provides a more explicit rule representation compared to the method used in prior works.
        \item Integration of both contrastive and auxiliary training into a novel ML approach to solving abstract visual reasoning tasks, which sets new state-of-the-art results on two major benchmark datasets -- Balanced-RAVEN~\cite{hu2020hierarchical} and PGM~\cite{santoro2018measuring}.
    \end{itemize}
%
%   The rest of this work is structured as follow. In Section~\ref{sec:related-work} we discuss prior related works on solving RPMs and on contrastive representation learning.
%    Our training method and its empirical evaluation are presented in Sections~\ref{sec:multi-label-contrastive-learning} and~\ref{sec:experiments}, respectively. In Section~\ref{sec:conclusion} we conclude the paper.

    \section{Related work}\label{sec:related-work}

    \paragraph{Raven's Progressive Matrices.}
    Although RPMs are characterised by rather simple visual representation, solving them is often a challenging task, as it requires to correctly identify all abstract relations between the component RPM images.
    In order to measure generalisation ability of neural modules in relational reasoning problems, \citeauthor{santoro2018measuring} (\citeyear{santoro2018measuring}) introduced the dataset of Procedurally Generated Matrices (PGM), which contains RPM problems divided by the authors into train and test sets.
    An important part of the PGM dataset are meta-target annotations, which encode the relations between objects and their attributes in a given RPM.
    Models trained to additionally predict these meta-targets by means of an auxiliary training, were shown to posses stronger generalisation capabilities compared to those that did not employ such an auxiliary training.

    The topic of meta-annotations was further extended in the RAVEN dataset~\cite{Zhang_2019_CVPR}, which contains additional structural annotations and RPMs with highly compositional structure.
    However, as reported in previous works, in the case of RAVEN data the auxiliary training seems to have little to no positive impact~\cite{Zhang_2019_CVPR,zhang2019learning}.
    Our investigations on Balanced-RAVEN dataset show that a more explicit rule encoding scheme can mitigate this problem.

    A recent paper~\cite{hu2020hierarchical} put in question the abstract visual reasoning capabilities of reported models, by highlighting a major flaw in the process of generating candidate answers in RAVEN. Due to an unintended bias in the generated set of possible choices, models trained purely on answers were still able to select a correct solution.
    To mitigate this problem, the dataset Balanced-RAVEN was proposed, in which the answers are generated in an unbiased way~\cite{hu2020hierarchical}.

    Initial reports of machine RPM solving demonstrated a notable gap between humans and ML algorithms, which stimulated active research in this area.
    A number of methods were designed for solving RPMs, which aim to identify relations between sets of objects~\cite{santoro2018measuring,zheng2019abstract} using a Relation Network~\cite{santoro2017simple}, reason with multi-layer multiplex graph neural network~\cite{Wang2020Abstract} or discover compositional representations using a scattering transformation~\cite{wu2020scattering}. Other works investigate human-inspired approaches, by measuring feature differences~\cite{mandziuk2019deepiq} or exploiting the hierarchical structure of RPMs~\cite{hu2020hierarchical}.
%    All proposed learning algorithms have access to large training sets, whereas humans aren't required to be familiar with the tests before attempting to solve them, as they are supposed to be a measure of adaptive problem solving skills, a problem which was tackled in~\cite{kim2020few}.
%

    \paragraph{Contrastive representation learning.}
    Previous studies shown that the ability of making analogies and contrasting experiences is a key ingredient of human intelligence~\cite{gentner1983structure,hofstadter1995fluid,hummel1997distributed,lovett2017modeling}.
    Consequently, these concepts were also implemented in ML systems solving abstract reasoning problems.
    \citeauthor{hill2018learning}~(\citeyear{hill2018learning}) discussed how to make analogies by contrasting abstract relational structure, while~\citeauthor{zhang2019learning}~(\citeyear{zhang2019learning}) incorporated contrast directly in the model architecture.
    Motivated by these findings, we propose another approach to abstract reasoning, by incorporating contrastive mechanism directly in the objective function, similarly to contrastive learning implementations in other domains.
    %
    %Besides the bias toward simple solutions, there exist other weaknesses of the cross-entropy loss identified in the literature, such as poor margins~\cite{cao2019learning}, vulnerability to adversarial %examples~\cite{nar2019cross,elsayed2018large} or susceptibility to noisy labels~\cite{wei2020combating,zhang2018generalized}.

    Among contrastive learning algorithms, the most related to our work is the Noise Contrastive Estimation (NCE) method~\cite{gutmann2010noise,mnih2013learning}, which was successfully utilized across various domains, such as image recognition~\cite{chen2020simple}, natural language processing~\cite{mikolov2013distributed} or reinforcement learning~\cite{Kipf2020Contrastive,srinivas2020curl}.
    NCE aims at building congruent representations for semantically related samples (positive pairs) and dissimilar representations for unrelated observations (negative pairs).
    It was shown that the concept of learning representations with NCE (as well as with other contrastive learning methods) is related to mutual information (MI) maximization principles~\cite{oord2018representation,hjelm2018learning,tian2020makes}.

    Contrastive learning is particularly useful in semi-supervised pre-training methods, where the availability of labeled data in downstream tasks is scarce~\cite{chen2020simple,chen2020big,chen2020improved,he2020momentum}.
    A recent work~\cite{khosla2020supervised} has shown that contrastive learning can outperform classical approaches also in fully supervised setting (with cross-entropy) and is characterised by greater robustness and stability to hyperparameter selection.
    This superior performance is a result of employing multiple positive pairs, which was further shown to increase MI lower bound by considering MI estimation as multi-label classification~\cite{song2020multi}.
    Despite this multi-label viewpoint, the framework proposed by~\citeauthor{song2020multi}, as well as the vast majority of prior works, focus on multi-class classification and don't discuss applicability of contrastive methods to multi-label problems.
    Our work aims at bridging this gap by defining a general contrastive learning framework which supports multi-label samples.

    \section{Multi-Label Contrastive Learning}\label{sec:multi-label-contrastive-learning}
    Motivated by successful applications of contrastive learning methods in other domains, we aim to extend it for multi-label setting and investigate its usage in learning representations of abstract visual reasoning problems. We start by proposing a general multi-label learning framework and then discuss its applicability to solving RPMs.

    \subsection{Preliminaries}\label{sec:preliminaries}
    Our method builds on the foundations of Supervised Contrastive Learning proposed by~\citeauthor{khosla2020supervised} (\citeyear{khosla2020supervised}) and extends it to support multi-label data. Given a randomly sampled batch $\{x_i, y_i\}_{i=1 \ldots N}\in \{\mathcal{X} \times \mathcal{Y}\}$ of size $N$, the base method consists of the following integral components:
    \begin{itemize}
        \item A data augmentation module, which transforms image $x_i$ into two randomly augmented views $\tilde{x}_{2i}$ and  $\tilde{x}_{2i-1}$, leading to an extended batch $\{\tilde{x}_i, \tilde{y}_i\}_{i=1 \ldots 2N}$ such that $\tilde{y}_{2i} = \tilde{y}_{2i-1} = y_i$.
        It should be noted that while this step is critical for Supervised Contrastive Learning its usage in our framework is optional.
        \item An encoder network $f_\theta$ which forms latent representations of the augmented views, defined as $h_i = f_\theta(\tilde{x}_i)$.
        The representations obtained from the encoder are $\ell_2$-normalized, which encourages learning from hard negatives and hard positives~\cite{khosla2020supervised} and simplifies the final linear classification task by aligning the features from positive pairs and uniformly distributing them on the hypersphere~\cite{wang2020understanding}.
        \item A projection network $g_\phi$, which maps feature representation into a lower-dimensional vector $z_i = g_\phi(h_i)$ suitable for computation of the contrastive loss.
        The representations obtained from the projection network are $\ell_2$-normalized, which this time allows to measure similarity between two vectors based on their dot-product.
        The projection network is realized by a non-linear function, an MLP with a single hidden layer, which was shown to be of critical importance~\cite{chen2020simple,chen2020big,chen2020improved}.
    \end{itemize}
    Both $f_\theta$ and $g_\phi$ are optimized jointly with respect to a Supervised Contrastive Loss, defined as follows:
    \begin{equation}
        \mathcal{L}^{\mathrm{sup}} = \sum_{i=1}^{2N} \mathcal{L}_i^{\mathrm{sup}}
    \end{equation}
    \begin{equation}
        \label{eq:supervised-contrastive-loss}
        \mathcal{L}_i^{\mathrm{sup}} = \frac{1}{2N_{\tilde{y}_i} - 1} \sum_{j=1}^{2N} \mathbbm{1}_{i \neq j} \cdot \mathbbm{1}_{\tilde{y}_i = \tilde{y}_j} \cdot \mathcal{L}_{i,j}^{\mathrm{sup}}
    \end{equation}
    \begin{equation}
        \mathcal{L}_{i,j}^{\mathrm{sup}} = -\log\frac{\exp(\textrm{sim}(z_i,z_j) / \tau)}{\sum_{k=1}^{2N} \mathbbm{1}_{i \neq k} \cdot \exp(\textrm{sim}(z_i,z_k) / \tau)}
    \end{equation}
    where $N_{\tilde{y}_i}$ is the number of samples in a given mini-batch with the same label as the anchor $i$, $\mathbbm{1}_\mathrm{B} \in \{0, 1\}$ is an indicator which evaluates to $1$ iff B is true, $\mathrm{sim}(z_i, z_j) = z_i \cdot z_j$ is a similarity measure defined as dot-product and $\tau > 0$ is a constant temperature parameter.

    After this pre-training stage, the weights of the encoder $f_\theta$ are frozen and the projection network $g_\phi$ is replaced with a randomly initialized linear classification head, which is trained with cross entropy on the downstream task.
    This procedure, commonly referred to as the linear evaluation protocol~\cite{oord2018representation,bachman2019learning}, provides a simple way to measure the quality of learned representations.

    \subsection{Multi-Label Contrastive Loss}
    In the default setting, the Supervised Contrastive Loss supports multi-class samples, i.e. $y_i \in \mathcal{Y}$.
    In order to extend it to multi-label samples $\{x_i, Y_i\}$ such that $Y_i \subset \mathcal{Y}$, we propose a novel objective function, the Multi-Label Contrastive Loss, which is defined as follows:
    \begin{equation}
        \label{eq:multi-label-contrastive-loss-sum}
        \mathcal{L}^{\mathrm{mlc}} = \sum_{i=1}^{2N} \mathcal{L}_i^{\mathrm{mlc}}
    \end{equation}
    \begin{equation}
        \label{eq:multi-label-contrastive-loss}
        \mathcal{L}_i^{\mathrm{mlc}} = \frac{1}{2N_{\widetilde{Y}_i} - 1} \sum_{j=1}^{N} \mathbbm{1}_{i \neq j} \cdot \mathbbm{1}_{\widetilde{Y}_i \cap \widetilde{Y}_j \neq \emptyset} \cdot \mathcal{L}_{i,j}^{\mathrm{mlc}}
    \end{equation}
    where $N_{\widetilde{Y}_i} = |\{\widetilde{Y}_j \mid \widetilde{Y}_i \cap \widetilde{Y}_j \neq \emptyset\}_{j=1 \ldots 2N}|$ is the number of samples in a given mini-batch which share at least one label with the anchor $i$ and $\mathcal{L}_{i,j}^{\mathrm{mlc}}=\mathcal{L}_{i,j}^{\mathrm{sup}}$.

    The difference when compared to eq.~(\ref{eq:supervised-contrastive-loss}) lies in the definition of positive pairs for an anchor $i$.
    While the base formulation defines the set of positives as those samples with exactly the same label, eq.~(\ref{eq:multi-label-contrastive-loss}) defines it as those samples which share at least one label.
    Our formulation preserves key properties of the base objective, i.e. 1) aggregates an arbitrary number of positive samples in the numerator and 2) increases contrastive strength by using all negative samples in the denominator.
    At the same time, the modified definition of positive pairs allows to handle samples with multiple labels.
    Analogously to the base method, our framework relies on the inner product of $\ell_2$-normalized vectors as a measure of similarity.

    \subsection{Adaptation to RPMs}\label{subsec:adaptation-to-rpms}
    In order to utilise the above-proposed Multi-Label Contrastive Learning (MLCL) framework for solving RPMs, let us first observe that each RPM can be naturally viewed as a multi-label sample, where labels correspond to the rules governing the RPM\@.
    Let us consider a mini-batch $\{\mathcal{M}_i\}_{i=1 \ldots N}$ of size $N$, where $\mathcal{M}_i = \{X_i, Y_i, k_i\}$ represents the whole RPM instance composed of a set of 16 images $X_i$, i.e. 8 context panels and 8 choice panels (out of which only one correctly completes the matrix).
    $Y_i \subset \mathcal{Y}$ is a set of associated rules such that $1 \leq |Y_i| \leq N_\mathcal{Y}$ and $k_i \in \{1 \ldots 8\}$ is an index of the correct answer.
    Here, $\mathcal{Y}$ is the set of all possible rules and $N_\mathcal{Y}$ is a dataset-dependent maximal number of rules for a single RPM.
%    The value of $N_R$ is $1 \leq |r_i| \leq 4$ for PGM and $1 \leq |r_i| \leq 8$ for both RAVEN and Balanced-RAVEN\@.

\begin{figure}[t]
        \centering
        \includegraphics[width=0.3\columnwidth]{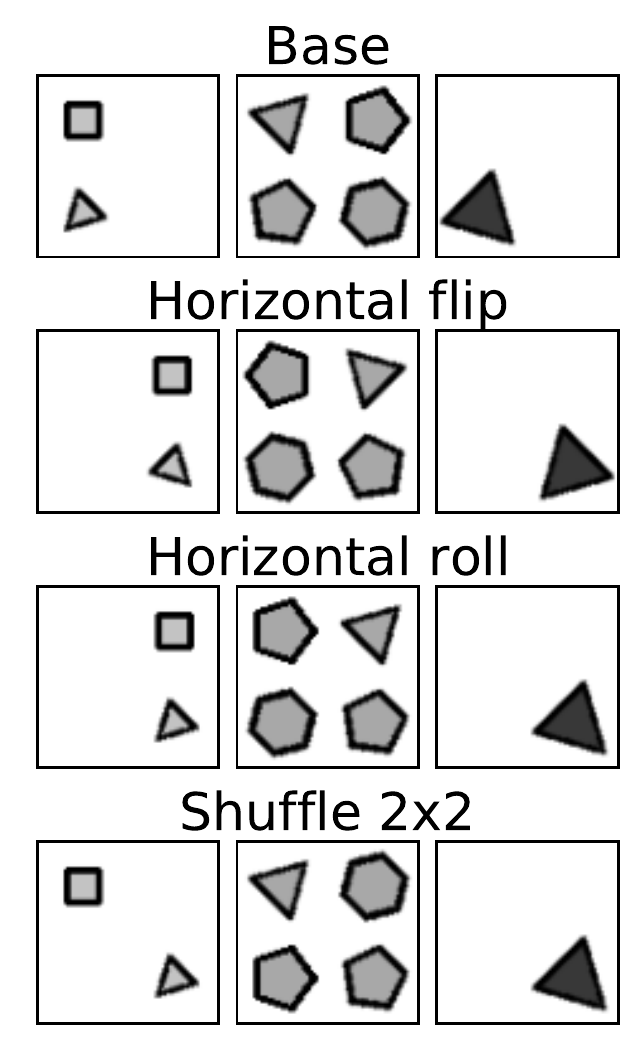}
        ~
        \includegraphics[width=0.3\columnwidth]{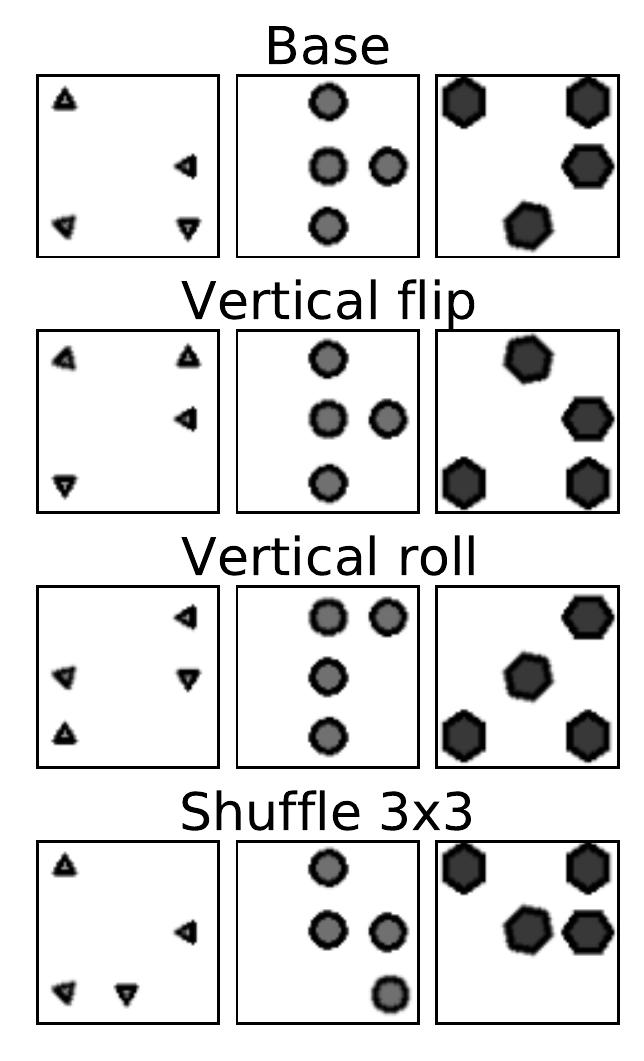}
        \caption{
        Augmentation of RPMs from the Balanced-RAVEN dataset.
        Selected transformation is applied in the same way to all images in a given RPM.
        The data augmentation module applies a randomly selected combination of presented methods with additional random rotation and transposition, using implementation from~\cite{buslaev2020albumentations}.
        For clarity, we only depict different views of a single row for two RPMs belonging to configurations \texttt{2x2Grid} (left part) and \texttt{3x3Grid} (right part), respectively.
        }
        \label{fig:augmentation}
    \end{figure}

    \paragraph{Data augmentation.}
    Previous works on solving RPMs do not report the usage of data augmentation.
    On the other hand, augmentation was highlighted as a fundamental component of NCE-based learning in other domains, e.g. by~\citeauthor{chen2020simple} (\citeyear{chen2020simple}).
    In order to conduct a meaningful comparison with prior literature, in the experiments both setups \emph{with} and \emph{without} augmentation are considered.
    In the former case (with augmentation), for each RPM in the mini-batch we apply two randomly selected augmentations, which gives a mini-batch of $2N$ RPMs $\widetilde{\mathcal{M}}_i = \{\widetilde{X}_i, \widetilde{Y}_i, \tilde{k}_i\}_{i=1 \ldots 2N}$, where $\widetilde{X}_{2i}$ and $\widetilde{X}_{2i-1}$\ are both obtained by augmenting $X_i$.
    All augmentations preserve both the underlying rules and index of the correct answer, hence: $\widetilde{Y}_{2i-1} = \widetilde{Y}_{2i} = Y_i$ and $\tilde{k}_{2i-1} = \tilde{k}_{2i} = k_i$.
    Since RPMs consist of greyscale images, we cannot rely on augmentation methods popular in image recognition.
    Instead, simple transformations which rearrange or rotate images are used, as showcased in Figure~\ref{fig:augmentation}.
    If augmentation is not applied a base mini-batch $\{\mathcal{M}_i\}_{i=1 \ldots N}$ is used.

    \paragraph{Contrastive pre-training.}
    For each RPM, we construct 8 different matrices by filling in the remaining context panel with each of the choice panels.
    For each sample $\widetilde{\mathcal{M}}_i$, this gives us a single RPM which satisfies the rules $\widetilde{Y}_i$ and 7 RPMs which satisfy fewer rules, because of the incorrectly chosen answer. We arrive at a batch divided into two components, one containing correct RPMs with their rules $\{\tilde{x}_i, \widetilde{Y}_i\}_{i=1 \ldots 2N}$ and the other one composed of incorrect RPMs $\{\tilde{x}_{i,l}'\}_{i=1 \ldots 2N, l=1 \ldots 7}$. This division allows us to consider the incorrectly completed RPMs as additional negative samples, for which:
    \begin{equation}
        \label{eq:multi-label-contrastive-loss-additional-negatives}
        \mathcal{L}_{i,j}^{\mathrm{mlc}} = -\log\frac{\exp(\textrm{sim}(z_i,z_j) / \tau)}{\sum_{k=1}^{2N} \mathbbm{1}_{i \neq k} \cdot \exp(\textrm{sim}(z_i,z_k) / \tau) + \Sigma_{i,k}}
    \end{equation}
    \begin{equation}
        \label{eq:multi-label-contrastive-loss-additional-negatives-sum}
        \Sigma_{i,k} = \sum_{l=1}^{7} \exp(\textrm{sim}(z_i,z_{k,l}') / \tau)
    \end{equation}
    We optimize both $f_\theta$ and $g_\phi$ with respect to the Multi-Label Contrastive Loss defined in eqs.~(\ref{eq:multi-label-contrastive-loss-sum})-(\ref{eq:multi-label-contrastive-loss-additional-negatives-sum}).

    \paragraph{Auxiliary training.}
    We support the contrastive pre-training with an auxiliary loss $\mathcal{L}^{\mathrm{aux}}$ introduced in~\cite{santoro2018measuring}.
    For this purpose, we employ a rule discovery network $\rho$ (implemented as an MLP with a single hidden layer), which transforms outputs of the encoder $f_\theta$ into a meta-target prediction $\pi_i = \rho(\{h_i\} \cup \{h_{i,l}'\}_{l=1 \ldots 7}) \in \mathbb{R}^d$, where $d$ is the dimension of meta-target encoding.
    We compare two setups for the rule encoding:
    \begin{itemize}
        \item \textit{Dense} - multi-hot encoding scheme introduced in~\cite{santoro2018measuring} and~\cite{Zhang_2019_CVPR}.
        It encodes each rule as a binary string of fixed length ($d=12$ for PGM and $d=9$ for Balanced-RAVEN) and performs logical OR operation on the whole set of rules.
        \item \textit{Sparse} - our proposed scheme, which encodes each rule as a one-hot vector of fixed length equal to the number of unique relations in the dataset ($d=50$ for PGM and $d=38$ for Balanced-RAVEN). Similarly to dense encoding it performs logical OR operation on the whole set of rules.
    \end{itemize}
    Intuitively, the advantage of proposed sparse encoding over the dense encoding used in prior works stems from information lossless OR operation, i.e. from sparse representation one can recover which rules were encoded, which is not always the case for dense encoding (e.g. when a single object is governed by multiple rules, or the same relation is applied to different objects). The difference between the two encodings is further discussed in supplementary material.
    Since sparse encoding provides a more explicit training signal it is the default encoding method for MLCL.
    Rule predictions are activated using a sigmoid unit and the loss is calculated using binary cross-entropy.
    We define the Multi-Label Contrastive Loss for solving RPMs as:
    \begin{equation}
        \label{eq:contrastive-auxiliary-loss}
        \mathcal{L} = \gamma\mathcal{L}^{\mathrm{mlc}} + \beta\mathcal{L}^{\mathrm{aux}}
    \end{equation}
    where $\gamma$ and $\beta$ are balancing factors.
    For simplicity, we set $\gamma = 1$ and $\beta = 10$ in all main experiments. Additional results are presented in supplementary material.

    \paragraph{Linear evaluation.}
    After the contrastive pre-training step, we discard the projection head $g_\phi$, freeze the parameters of the encoder $f_\theta$ and attach a simple linear scoring head $s_\psi$ with a single output neuron.
    For each RPM problem $\mathcal{M}_i$, we employ $f_\theta$ to generate the matrix representations $\{h_i\} \cup \{h_{i,l}'\}_{l=1 \ldots 7}$, calculate a score using the scoring head $s_\psi$ and apply softmax to produce a probability distribution over the set of possible answers.
    Using the estimated probability, we optimize the scoring head with a standard cross-entropy loss and keep weights of the encoder network frozen.
    \begin{figure}[t]
        \centering
        \includegraphics[width=0.8\columnwidth]{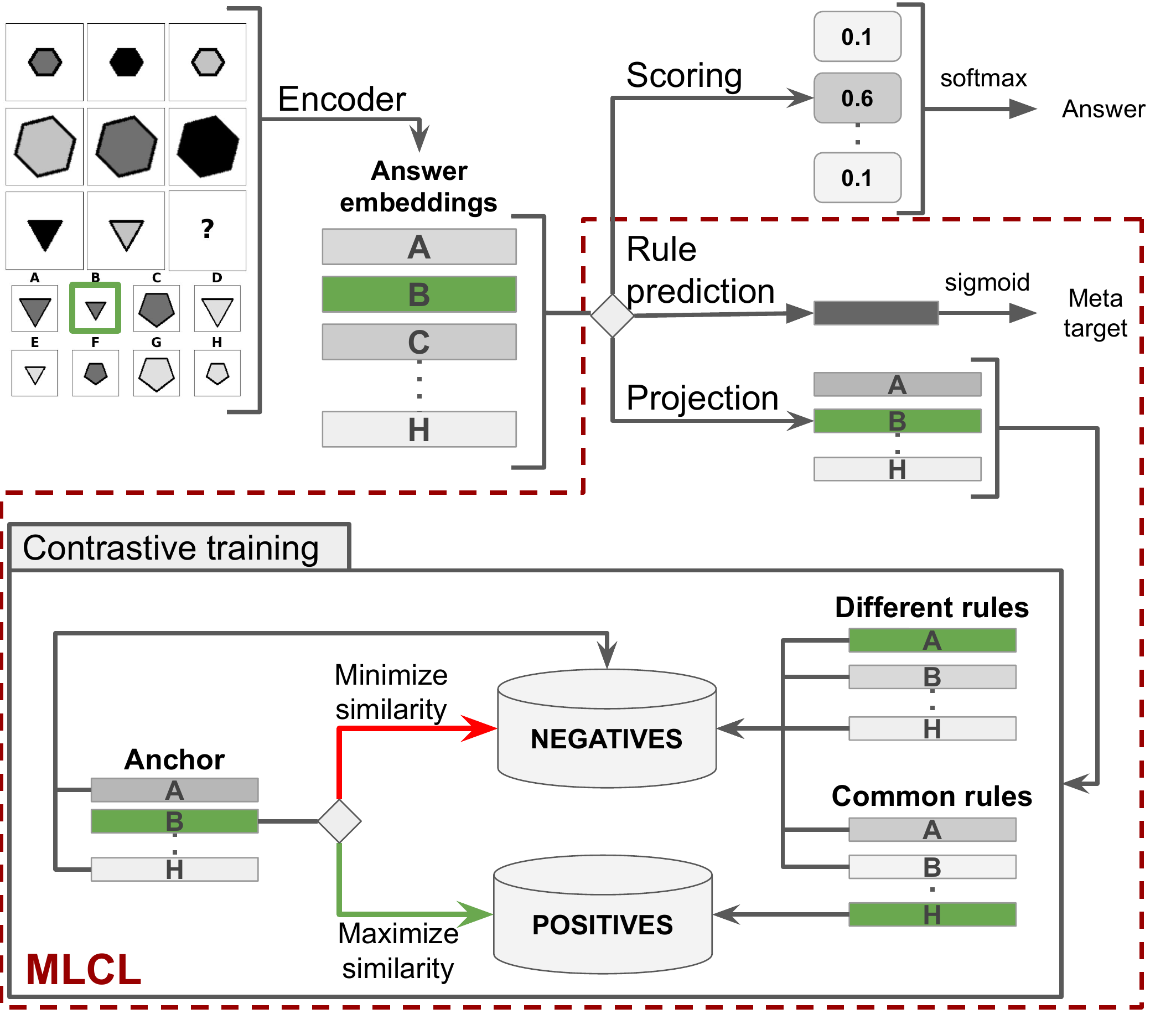}
        \caption{
        In all training setups, we start by filling in the context panels with each choice panel. In each case, we generate an embedding with an abstract reasoning encoder network (SCL, HriNet or CoPINet). The embeddings are used as an input to 1) a scoring module which predicts the answer (supervised training, denoted as CE), 2) a rule prediction module which predicts rules as encoded meta-target (auxiliary training, denoted as AUX) and 3) a projection network which maps them into a lower-dimensional representation suitable for computing contrastive loss.
        The proposed approach combines both auxiliary and contrastive training into a joint learning framework -- MLCL.
        }
        \label{fig:method}
    \end{figure}

    \section{Experiments}\label{sec:experiments}
    We compare our method (MLCL) with a fully supervised framework for solving RPMs used throughout the literature~\cite{santoro2018measuring,Zhang_2019_CVPR}, referred to as CE. In CE both the encoder network $f_\theta$ and $s_\psi$ are used and optimized in an end-to-end manner with respect to the cross-entropy objective function.
    Since our contrastive training method is inherently based on the description of the RPM rules, we support the supervised training with the same auxiliary loss $\mathcal{L}^{\mathrm{aux}}$ and rule discovery network $\rho$, obtaining:
    \begin{equation}
        \mathcal{L} = \mathcal{L}^{\mathrm{ce}} + \beta\mathcal{L}^{\mathrm{aux}}\label{eq:CE-base}
    \end{equation}
    where $\beta$ is a balancing coefficient. This enhanced training is referred to as CE+AUX. Following the setup from~\cite{santoro2018measuring}, we set $\beta = 10$ in all experiments.
    In order to comprehensively compare our method with the baseline, in the experimental evaluation we consider three different state-of-the-art models for abstract visual reasoning and use two abstract reasoning RPM benchmarks.

    \paragraph{Datasets.}
    Our main findings are demonstrated on the newest RPM benchmark set -- Balanced-RAVEN~\cite{hu2020hierarchical}, which contains visually diverse RPMs with highly compositional structure.
    Balanced-RAVEN is a modification of the original RAVEN dataset~\cite{Zhang_2019_CVPR} which fixes the defect of biased choice panels.
    It contains novel relation types which are not present in PGM. Each RPM in Balanced-RAVEN can be governed by up to 8 different rules.
    Balanced-RAVEN is divided into seven distinct visual configurations. In the default setting each of them contains 10K problem instances.

    Additionally, we support our studies with the experiments on large-scale PGM dataset~\cite{santoro2018measuring}, containing RPM problems with between 1 and 4 abstract rules.
    The main purpose of PGM is to test generalisation skills of ML models across various regimes. Each regime is characterised by explicitly defined differences between training and test data, which address various types of generalisation, e.g. interpolation or ability to reason about relations not seen in the training set.
    The dataset contains 1.42M RPM problems per regime.
    %    In total, there exist 50 unique rule configurations, which determine the relations between objects and their attributes.

    \paragraph{Models.}
    We use three different state-of-the-art models for solving RPMs as the encoder network $f_\theta$: SCL~\cite{wu2020scattering}, HriNet~\cite{hu2020hierarchical} and CoPINet~\cite{zhang2019learning}.
    In order to reduce the training time and memory consumption of HriNet, we replaced its ResNet backbone~\cite{he2016deep} in all three hierarchies with a simpler CNN architecture, analogous to the one used in the Wild Relation Network~\cite{santoro2018measuring}.
    The details are presented in supplementary material.

    \paragraph{Implementation details.}
    We devoted a maximum of 50 training epochs with batch size of 256 and learning rate $0.003$ for each model on the PGM dataset and 100 epochs with batch size of 128 and learning rate $0.002$ for Balanced-RAVEN. Model parameters were optimized using the ADAM optimizer~\cite{kingma2014adam} with $\beta_1=0.9$, $\beta_2=0.999$, $\epsilon=10^{-8}$ until the maximal number of epochs was reached or the loss stopped improving on a validation set.
    All experiments were conducted using mixed-precision training on a single worker with 4 NVIDIA Tesla P100 GPUs, each with 16Gb of memory.
    %We use our own implementation of SCL, modified implementation of HriNet~\footnote{https://github.com/husheng12345/RPM} and original implementation of %CoPINet~\footnote{https://github.com/WellyZhang/CoPINet}.
    %Following best practices for reproducible research, we will distribute our code in an open-source repository as PyTorch Lightning modules~\cite{falcon2019pytorch}.
%
    \begin{table*}[t]
%        \small
        \centering
        \begin{adjustbox}{max width=\textwidth}
            \begin{tabular}{cl|c|ccccccc}
                \toprule
                \multicolumn{2}{c|}{\multirow{2}{*}{Method}} & \multicolumn{8}{c}{Test accuracy (\%)} \\
    %            \cmidrule{3-10} \\
                &                           & Average $\pm$ Std       & \texttt{Center} & \texttt{2x2Grid} & \texttt{3x3Grid} & \texttt{L-R} & \texttt{U-D} & \texttt{O-IC} & \texttt{O-IG} \\
                \midrule
                \multirow{6}{*}{SCL}     & CE                        & $82.8 \pm 0.7$          & $99.0$          & $80.3$           & $74.7$           & $83.9$       & $84.4$       & $87.3$        & $69.8$        \\
                & CE+AUX-dense              & $86.9 \pm 3.5$          & $98.9$          & $86.9$           & $80.6$           & $88.4$       & $88.8$       & $92.0$        & $72.7$        \\
                & CE+AUX-sparse             & $95.9 \pm 0.3$          & $99.8$          & $97.0$           & $91.8$           & $99.1$       & $99.1$       & $98.7$        & $85.8$        \\
                & MLCL                      & $95.7 \pm 0.5$          & $99.6$          & $96.4$           & $90.7$           & $99.2$       & $99.1$       & $99.1$        & $85.9$        \\
                & MLCL+AUG                  & $\textbf{96.8} \pm 0.4$ & $99.8$          & $97.6$           & $93.8$           & $99.3$       & $99.4$       & $99.1$        & $88.9$        \\
    %            \cmidrule{3-10}
                & \cite{wu2020scattering}   & 95.0                    & 99.0            & 96.2             & 89.5             & 97.9         & 97.1         & 97.6          & 87.7          \\
                \midrule
                \multirow{6}{*}{HriNet}  & CE                        & $50.6 \pm 1.3$          & $66.3$          & $41.0$           & $35.6$           & $56.0$       & $54.9$       & $58.5$        & $41.8$        \\
                & CE+AUX-dense              & $51.6 \pm 1.5$          & $75.7$          & $41.2$           & $36.5$           & $55.0$       & $55.4$       & $57.0$        & $40.2$        \\
                & CE+AUX-sparse             & $57.0 \pm 1.6$          & $84.5$          & $43.2$           & $36.9$           & $65.4$       & $66.1$       & $60.5$        & $42.6$        \\
                & MLCL                      & $56.2 \pm 1.5$          & $78.9$          & $44.5$           & $38.2$           & $64.4$       & $64.1$       & $60.5$        & $42.9$        \\
                & MLCL+AUG                  & $\textbf{65.7} \pm 2.4$ & $94.4$          & $57.3$           & $46.8$           & $75.3$       & $76.0$       & $64.2$        & $45.8$        \\
    %            \cmidrule{3-10}
                & \cite{hu2020hierarchical} & 63.9                    & 80.1            & 53.3             & 46.0             & 72.8         & 74.5         & 71.0          & 49.6          \\
                \midrule
                \multirow{6}{*}{CoPINet} & CE                        & $44.8 \pm 0.8$          & $53.6$          & $35.9$           & $30.2$           & $51.0$       & $52.1$       & $51.1$        & $39.9$        \\
                & CE+AUX-dense              & $32.2 \pm 1.9$          & $37.1$          & $30.1$           & $27.6$           & $33.4$       & $35.8$       & $33.3$        & $28.0$        \\
                & CE+AUX-sparse             & $46.8 \pm 2.3$          & $52.7$          & $37.6$           & $34.2$           & $52.6$       & $55.7$       & $55.5$        & $39.0$        \\
                & MLCL                      & $57.1 \pm 0.4$          & $68.5$          & $42.8$           & $36.3$           & $68.8$       & $71.4$       & $68.7$        & $42.9$        \\
                & MLCL+AUG                  & $\textbf{58.0} \pm 0.4$ & $68.3$          & $44.6$           & $38.9$           & $69.0$       & $71.7$       & $69.0$        & $44.4$        \\
    %            \cmidrule{3-10}
                & \cite{wu2020scattering}   & 46.3                    & 54.4            & 33.4             & 30.1             & 56.8         & 55.6         & 54.3          & 39.0          \\
                \bottomrule
            \end{tabular}
        \end{adjustbox}
        \caption{
        Test accuracy on the Balanced-RAVEN dataset averaged across 4 random seeds.
        Results are reported for three different encoder networks and the following training setups: supervised with cross-entropy (CE), supervised augmented with auxiliary training with \emph{dense} (CE+AUX-dense) or \emph{sparse} (CE+AUX-sparse) encoding and Multi-Label Contrastive Learning \emph{without} (MLCL) or \emph{with} (MLCL+AUG) data augmentation.
        Average - denotes the mean accuracy for all configurations, \texttt{L-R} denotes (\texttt{Left-Right}), \texttt{U-D} (\texttt{Up-Down}), \texttt{O-IC} (\texttt{Out-InCenter}), \texttt{O-IG} (\texttt{Out-InGrid}). For each model, the last row presents the best result reported in the literature.
        }
        \label{tab:balanced-raven}
    \end{table*}

    \subsection{Results}
    We start experimental evaluation by comparing the proposed MLCL method with other training setups for three different ML abstract reasoning models on the Balanced-RAVEN dataset (see Table~\ref{tab:balanced-raven}).
    For all three models MLCL with data augmentation excels the best results from the literature, and in the case of CoPINet by a significant margin. Furthermore, for all three models MLCL significantly outperforms the base setup (CE), which does not utilize the rule-related information in the training signal. This observation confirms that MLCL is able to absorb and efficiently utilize this additional structural information, which is not a common property, as shown by the CoPINet's performance deterioration when trained with dense encoding.

    For SCL and HriNet, MLCL (without augmentation) matches the performance of supervised training supported by auxiliary loss with sparse encoding and excels it by above 10 p.p. for CoPINet. We hypothesize that contrastive nature of our objective function (eqs.~(\ref{eq:multi-label-contrastive-loss-sum})-(\ref{eq:multi-label-contrastive-loss-additional-negatives-sum})) amplifies the benefits of architectural contrastive mechanisms in CoPINet.

    Moreover, auxiliary training with sparse encoding (CE+AUX+sparse) always improves performance over the baseline CE setup (with $\beta=0$ in eq. (\ref{eq:CE-base})). At the same time, CE+AUX+dense results in worse than CE performance for CoPINet, which aligns with the outcomes reported in~\cite{zhang2019learning}, and yields only a slight improvement for both SCL and HriNet. Hence, for the Balanced-RAVEN dataset, the more explicit, sparse rule encoding scheme is overall beneficial, regardless of a particular ML model.

    Table~\ref{tab:pgm} presents results on the PGM dataset. Due to huge number of RPM instances, we compare results of the best-performing encoder network -- SCL, with the best MLCL configuration -- with augmentation.
    In the most demanding regime (H.O. Shape-Colour) all training methods achieve close-to-random results. MLCL+AUG achieves superior results in 2 regimes and in the remaining 5 cases best outcomes are accomplished by SCL supervised training supported with an auxiliary loss with either dense or sparse encoding.
    MLCL outperforms the base setup (CE) in 5 regimes.
    The results show that for the PGM dataset, none of the evaluated training setups clearly stands out.
    On the contrary -- each method seems to present different kind of generalisation abilities.
    It should be noted, however, that MLCL is tested under linear evaluation protocol, without fine-tuning of the encoder network $f_\theta$ and therefore reaching the results comparable to fully supervised setups (or even outperforming them in 2 regimes) proves its strength and potential.
    \begin{table}[t]
        \centering
        \begin{adjustbox}{max width=\textwidth}
            \begin{tabular}{l|ccc|ccc|ccc|ccc}
                \toprule
                \multirow{3}{*}{Regime} & \multicolumn{12}{c}{Accuracy (\%)} \\
                & \multicolumn{3}{|c}{CE} & \multicolumn{3}{|c}{CE+AUX-dense} & \multicolumn{3}{|c}{CE+AUX-sparse} & \multicolumn{3}{|c}{MLCL+AUG} \\
                & Val. & Test. & Diff. & Val. & Test.         & Diff. & Val. & Test.         & Diff. & Val. & Test.         & Diff. \\
                \midrule
                Neutral               & 86.2 & 85.6  & -0.6  & 87.6 & \textbf{87.1} & -0.5  & 87.4 & \textbf{87.1} & -0.3  & 71.0 & 71.1          & +0.1  \\
                Interpolation         & 91.2 & 55.8  & -35.4 & 97.9 & 56.0          & -41.9 & 88.1 & 54.1          & -34.0 & 93.2 & \textbf{70.9} & -22.3 \\
                H.O.\ Attribute Pairs & 56.4 & 40.8  & -15.6 & 88.6 & \textbf{79.6} & -9.0  & 80.8 & 63.6          & -17.2 & 79.7 & 66.0          & -13.7 \\
                H.O.\ Triple Pairs    & 78.2 & 64.5  & -13.7 & 88.7 & \textbf{76.6} & -12.1 & 77.5 & 64.0          & -13.5 & 86.1 & 71.7          & -14.4 \\
                H.O.\ Triples         & 78.6 & 27.0  & -51.6 & 88.1 & 23.0          & -65.1 & 86.0 & \textbf{30.8} & -55.2 & 84.0 & 22.1          & -61.9 \\
                H.O.\ Line-Type       & 87.6 & 15.1  & -72.5 & 87.9 & 14.1          & -73.8 & 93.5 & \textbf{17.0} & -76.5 & 86.1 & 16.1          & -70.0 \\
                H.O.\ Shape-Colour    & 96.9 & 12.7  & -84.2 & 98.3 & 12.6          & -85.7 & 98.0 & 12.7          & -85.3 & 94.1 & 12.8          & -81.3 \\
                Extrapolation         & 96.3 & 17.3  & -79.0 & 99.1 & 19.8          & -79.3 & 96.9 & 17.5          & -79.4 & 83.0 & \textbf{21.9} & -61.1 \\
                \bottomrule
            \end{tabular}
        \end{adjustbox}
        \caption{
        Accuracy in all regimes of the PGM dataset with SCL as the encoder network.
        We use the same notation for training methods as in Table~\ref{tab:balanced-raven}.
        For each regime, we report results on the validation set (Val.), test set (Test.) and their difference (Diff.).
        Training and validation sets have the same distribution, whereas the test set has different distribution, specific to a given regime.
%    This shift in distribution allows to test different types of generalisation.
        }
        \label{tab:pgm}
    \end{table}

    \subsection{Ablation study}\label{subsec:ablation-study}
    In the ablation study we further validate the role of MLCL framework as an auxiliary training method and analyze its contrastive properties on the Balanced-RAVEN dataset.

    \paragraph{Data augmentation.}
    Depending on the choice of the encoder network MLCL either achieves comparable performance (SCL and HriNet) or outperforms (CoPINet) supervised approaches even without augmentations (cf. Table~\ref{tab:balanced-raven}).
    However, since most of the contrastive methods heavily rely on data augmentation, we analyze its influence on MLCL for solving abstract visual problems.
    The results of using data augmentation reported in Table~\ref{tab:balanced-raven} (denoted as MLCL+AUG) show consistent improvement for all encoders.
    This aligns with observations reported throughout the contrastive representation learning literature.
    Most notably, combining our contrastive framework for solving RPMs with data augmentation reduces the error rate on Balanced-RAVEN to 3.2\%, which sets the new state-of-the-art result.
    Additionally, we see a large performance gain for HriNet, which surpasses the original result reported in~\cite{hu2020hierarchical} even with a much weaker perceptual backbone.

    \paragraph{Joint optimization.}
    MLCL combines both contrastive and auxiliary losses, as shown in eq.~(\ref{eq:contrastive-auxiliary-loss}).
    We have verified that using either one of its individual components alone is not sufficient for building strong representations.
    When using contrastive loss only, that is with $\beta=0$, we observed a serious performance downgrade for all models: the accuracy on SCL dropped to 35.3\%, on HriNet to 20.7\% and on CoPINet to 19.0\%. We hypothesize that the lack of auxiliary training information makes it difficult to extract abstract rules and encourages to focus too much on the visual similarity of RPMs, which is unprofitable for the final downstream task.
    Similarly, setting $\gamma=0$, decreased the accuracy on SCL, HriNet and CoPINet to 46.7\%, 31.9\% and 24.2\%, respectively.
    This suggests that with purely auxiliary training, models are unable to relate the same abstract relationships to different visual figure configurations.
    Moreover, the absence of contrastive loss significantly hinders the ability to discriminate between correctly and incorrectly completed RPMs.
    These observations stress the importance of using joint loss, which is realized by setting both $\beta > 0$ and $\gamma > 0$.

    \paragraph{Contrast strength.}
    The quality of representations learned with contrastive frameworks benefits from applying bigger contrast to positive samples, which is realized by including additional negative pairs while calculating loss~\cite{chen2020simple}.
    Analogously, methods which support multiple positive samples tend to benefit from higher number of positive pairs in a given batch~\cite{khosla2020supervised}.
    These two properties of contrastive approaches result in the reliance on large batch sizes, which some works tried to tackle by storing additional negative samples in a memory bank~\cite{wu2018unsupervised} or using dynamically updated queue with a moving-average encoder~\cite{he2020momentum,chen2020improved}.
%    In Figure~\ref{fig:batch-size}, we validate the dependence of MLCL on batch sizes by varying the sizes used during the pre-training stage (the linear classifier is trained in default setting).
%    Empirical evaluation shows, that MLCL doesn't require large batch sizes and surpasses the performance of traditional supervised training approaches under the same experimental protocol.
%    In fact, we see a slight drop in the final performance for larger batches across all models, which suggests their negative influence on the auxiliary training.
    Contrary to the above findings, empirical evaluation of MLCL shows that the method does not require large batch sizes and surpasses the performance of traditional supervised training under the same experimental protocol. In fact, for larger batches we observed a slight drop in the final performance across all models, which suggests their negative influence on the auxiliary training. A further analysis of this phenomenon is presented in supplementary material.

    In the final linear classification stage, the main goal is to select a correct answer, which requires to discriminate between correctly and incorrectly completed RPMs.
    We additionally analyze the importance of using RPMs with incorrect answers as additional negative samples, by removing the term $\Sigma_{i,k}$ from the denominator of eq.~(\ref{eq:multi-label-contrastive-loss-additional-negatives}).
    Unavailability of these RPMs results in a notable drop of performance on the downstream task to 58.0\% for SCL, 37.2\% for HriNet and 26.5\% for CoPINet.
    In fact, when this additional signal is ignored, the encoder network only learns how to differentiate between correctly completed RPMs governed by different sets of rules, whereas the final classification task requires to differentiate between a correct RPM and a set of incorrectly completed RPMs.
    We conclude that this additional training signal obtained from RPMs with incorrect answers is mandatory for high downstream task performance.

    \section{Conclusion}\label{sec:conclusion}
    In this work we propose a novel NCE algorithm suitable for multi-label samples and integrate it with an auxiliary training to devise a new ML approach (MLCL) to abstract visual reasoning tasks. The efficacy of MLCL is tested on a challenging task of solving RPMs which is formulated in this paper as a multi-label classification problem with the 1-1 correspondence between labels and abstract rules underlying a given RPM.
    The proposed approach establishes new state-of-the-art results on the Balanced-RAVEN dataset and demonstrates superior performance in 2 regimes from PGM.
%   {\color{red}In some sense MLCL imitates human approach to solving RPMs which rely on seeking commonalities and differences among objects %constituting a visual RPM representation.}
    The MLCL framework is additionally supported by a sparse rule encoding scheme for RPMs introduced in the paper, which consistently outperforms the encoding method used in prior works on the Balanced-RAVEN dataset and is the preferred scheme for half of the PGM regimes.

    \bibliographystyle{plainnat}
    \bibliography{main}
    \vfill
    \eject

    \appendix
    \renewcommand{\thesection}{\Alph{section}}
    \counterwithin{figure}{section}
    \section*{Supplementary Material}

    \section{Extended problem description}
    In the main body of the paper we have introduced the problem of solving RPMs only briefly.
    Here we provide additional details and in particular discuss the differences between rule encoding schemes used in the two considered datasets (Balanced-RAVEN and PGM). In both of them the goal is to complete a $3\times3$ matrix with a missing panel in the bottom-right corner.
    The answer has to be chosen from a set of 8 candidates, such that the chosen image satisfies \emph{all} abstract rules defining a given RPM\@.
    In each problem instance there is only one answer which correctly completes the matrix, while the remaining answers usually satisfy some (but not all) of the underlying rules.

    The sets of possible rules differ between the datasets. In Balanced-RAVEN a set of rules associated to a given RPM (called an \emph{abstract structure}) is defined as a set of pairs $\mathcal{S} = \{[r, a] | r\in\mathcal{R}, a\in\mathcal{A}\}$, where $\mathcal{R} = \{\texttt{Constant}, \texttt{Progression}, \texttt{Arithmetic},\allowbreak \texttt{Distribute Three}\}$ is the set of relation types and $\mathcal{A} = \{\texttt{Number}, \texttt{Position}, \texttt{Type}, \texttt{Size}, \texttt{Color}\}$ is the set of attributes.

    The abstract structure of RPMs from the PGM dataset is defined as a set of triples $\mathcal{S} = \{[r, o, a] | r\in\mathcal{R}, o\in\mathcal{O}, a\in\mathcal{A}\}$, where $\mathcal{R} = \{\texttt{progression}, \texttt{XOR}, \texttt{OR}, \texttt{AND}, \texttt{consistent union}\}$ is the set of relation types, $\mathcal{O} = \{\texttt{shape}, \texttt{line}\}$ is the set of object types and $\mathcal{A} = \{\texttt{size}, \texttt{type}, \texttt{color}, \allowbreak\texttt{position}, \texttt{number}\}$ is the set of attribute types.

    \begin{wrapfigure}{r}{0.5\textwidth}
        \centering
        \includegraphics[width=0.35\columnwidth]{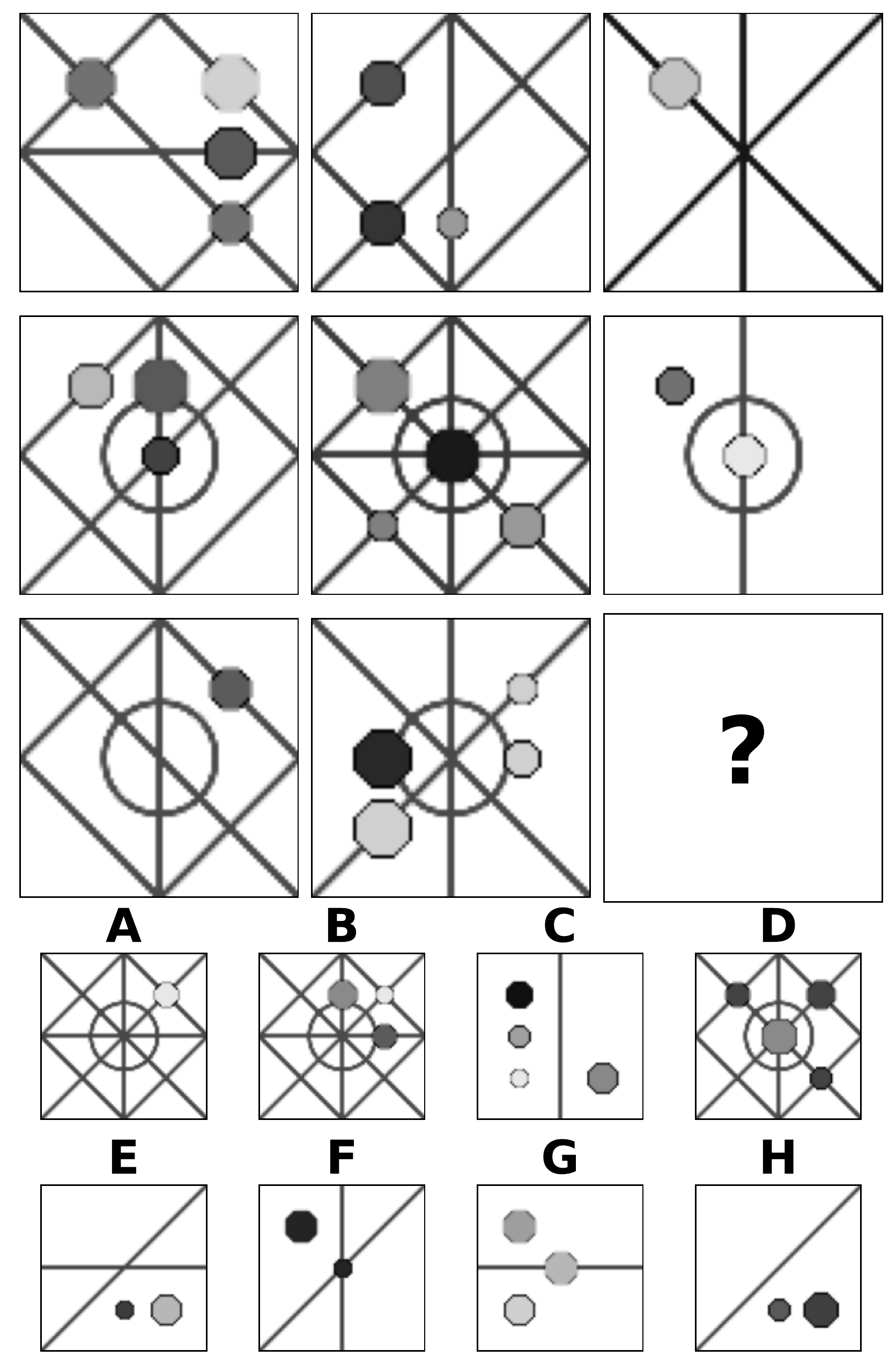}
        \caption{
        RPM example from the PGM dataset presented in the main paper. The matrix is governed by a single rule -- \texttt{AND} applied row-wise to octagon shape position. The correct answer is A.}
        \label{fig:rpm-supp}
    \end{wrapfigure}

    In order to better present the problem, let us consider the first example of an RPM from the main text (Fig.~\ref{fig:rpm-supp}).
    Searching for common patterns among the images leads us to the following observation:
    in both the first (top) and the second (middle) row images, the octagon shapes in the last (right) column images appear only in positions where there are octagons in both the first (left) and the second (middle) column images.
    Consequently, the respective rule ($s=[\texttt{shape},\texttt{position},\texttt{AND}]$) is potentially a part of the underlying abstract structure associated with this RPM, i.e. one may suppose that $s \in \mathcal{S}$.
    After applying this rule to the third (bottom) row, one would expect the missing image to have only a single octagon, located in the top-right corner.

    In fact, only a single image (with label A) from the set of potential candidates matches this description. Furthermore, the remaining shape attributes (\texttt{size}, \texttt{color}, \texttt{number}), as well as various combinations of lines in this RPM do not satisfy any of the allowed rules from $\mathcal{S}$ (actually their role is to \emph{distract} the solver). The above observations lead us to a conclusion that for the considered RPM $\mathcal{S} = \{[\texttt{shape},\texttt{position},\texttt{AND}]\}$ and that the correct answer is A.

    Observe that in the described example, the rule was applied row-wise, however, in general the RPMs from the PGM dataset may have rules applied either row- or column-wise, whereas rules from the Balanced-RAVEN dataset are only applied row-wise.

    Further examples of RPMs together with explanation of their solutions are presented in Figs.~\ref{fig:augmentation-center}-\ref{fig:augmentation-pgm2}. We refer the readers interested in a more detailed description of both datasets to the original works of~\citeauthor{santoro2018measuring} (\citeyear{santoro2018measuring}) and~\citeauthor{Zhang_2019_CVPR} (\citeyear{Zhang_2019_CVPR}).

    \section{\emph{Sparse} vs \emph{dense} rule encoding}
    Let us now revisit the two rule encoding schemes discussed in the paper.
    \citeauthor{santoro2018measuring}~(\citeyear{santoro2018measuring}) proposed to encode the abstract rules as binary strings of length 12 (called \emph{meta-targets}) according to the following syntax: (\texttt{shape}, \texttt{line}, \texttt{color}, \texttt{number}, \texttt{position}, \texttt{size}, \texttt{\texttt{type}}, \texttt{progression}, \texttt{XOR}, \texttt{OR}, \texttt{AND}, \texttt{consistent union}).
    In order to support RPMs with multiple rules, the meta-target for a given RPM is obtained by performing an \texttt{OR} operation on the set of encoded individual rules.

    For example, for an RPM instance with rules $\mathcal{S}$ $=$ $\{[\texttt{OR},\texttt{shape},\texttt{type}],[\texttt{AND},\texttt{line},\texttt{color}]\}$, the method, referred to as \emph{dense} encoding in the paper, yields the following meta-target:
    \[
        \texttt{OR}([100000100100],[011000000010])=[111000100110]
    \]
    Based on the resultant string, one is able to conclude that the underlying structure consists of \texttt{OR} and \texttt{AND} relations, \texttt{shape} and \texttt{line} objects, \texttt{type} and \texttt{color} attributes.
    However, recovering the exact relations governing the considered RPM is not possible.

    The new rule encoding scheme proposed in the paper, referred to as \emph{sparse} encoding, is more explicit and allows to unambiguously retrieve the encoded relations, with the aim of providing a more accurate training signal.
    Since the set of all possible \emph{abstract structures} in PGM is composed of 50 elements, i.e. there are $|\mathcal{R}| \times |\mathcal{O}| \times |\mathcal{A}| = 2 \times 5 \times 5 = 50$ unique rules, our method encodes them as a one-hot vector of length 50.
    Similarly to the dense encoding, we perform an \texttt{OR} operation on the set of all rules for a given RPM.
    However, due to the nature of one-hot encoding, all information about the underlying abstract structure is preserved.

    Analogously to PGM, the rule encoding method for RPMs from the Balanced-RAVEN dataset, proposed by \citeauthor{Zhang_2019_CVPR}~(\citeyear{Zhang_2019_CVPR}), represents each rule as a multi-hot vector of length $|\mathcal{R}| + |\mathcal{A}| = 9$ and combines the set of individual rule encodings by means of an \texttt{OR} operation.
    In effect, similarly to the case of PGM, recovering individual rules constituting the final encoded representation is not possible.
    This problem is further exacerbated by the generally higher numbers of rules in Balanced-RAVEN instances. RPMs in the Balanced-RAVEN dataset contain $6.29$ rules on average, compared to only 1.37 rules (on average) in RPMs from the PGM dataset~\cite{Zhang_2019_CVPR}.
    Consequently, for RPMs from the Balanced-RAVEN dataset, the resultant representation obtained with dense encoding is highly ambiguous.
    % and contains little to none useful information.

    Sparse rule representations of Balanced-RAVEN RPMs are obtained analogously to the PGM ones.
    In this dataset there are $|\mathcal{R}| \times |\mathcal{A}| = 4 \times 5 = 20$ unique configurations of rules and attributes.
    However, \citeauthor{Zhang_2019_CVPR}~(\citeyear{Zhang_2019_CVPR}) pointed out that applying \texttt{Arithmetic} to \texttt{Type} is counterintuitive.
    Additionally, in some configurations the rules can be applied to both component structures.
    Namely, each of the configurations \texttt{L-R}, \texttt{U-D}, \texttt{O-IC}, \texttt{O-IG} consists of two distinct substructures: left/right, up/down, outer/inner, outer/inner, respectively.
    This gives a total of 38 unique rules, which is the length of our one-hot vector representation.
    Again, due to the nature of one-hot encoding, all information related to the abstract structure is preserved after applying the OR operation.

    In summary, although the advantage of sparse encoding for PGM data is demonstrated only in certain regimes, it consistently outperforms the previous (dense) encoding method on Balanced-RAVEN, due to the significantly higher average number of rules per RPM instance in this dataset.

    \begin{figure}[t]
        \centering
        \includegraphics[width=0.99\textwidth]{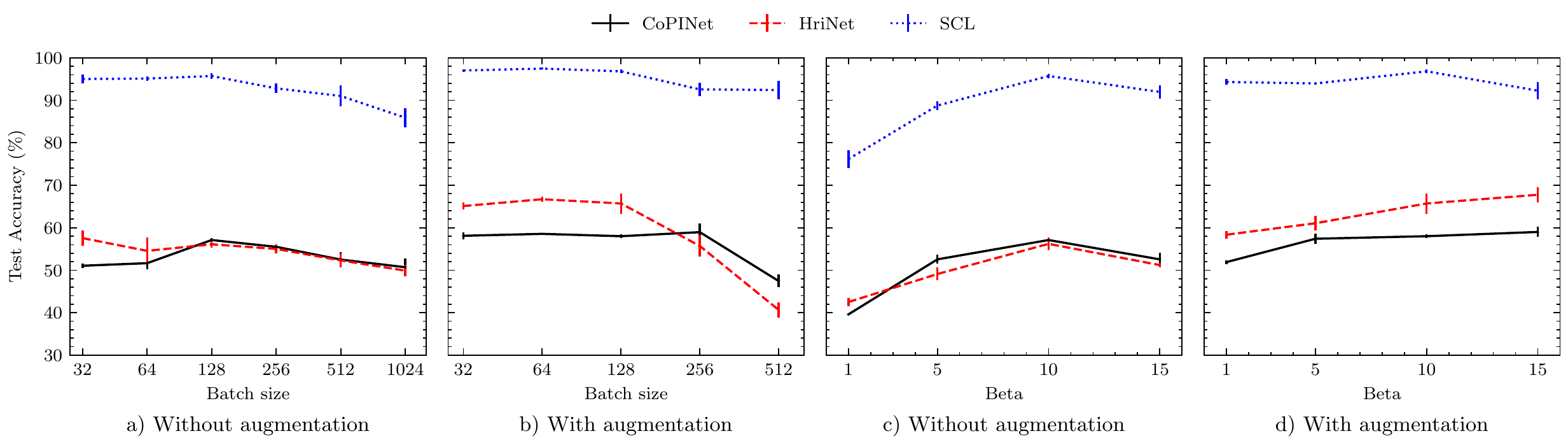}
        \caption{
        Additional ablation studies on the Balanced-RAVEN dataset.
        The figures present variations in the final classification performance averaged across 4 random seeds depending on: 1) a batch size used for the contrastive pre-training stage of MLCL a) \emph{without} and b) \emph{with} data augmentation; 2) balancing factors in the definition of Multi-Label Contrastive Loss c) \emph{without} and d) \emph{with} data augmentation.
        }
        \label{fig:additional-ablations}
    \end{figure}

    \section{Data augmentation}
    To the best of our knowledge this work is the first to investigate the use of data augmentation strategies in solving RPMs.
    For each sample, our data augmentation module chooses a random subset from available transformations and applies them in the same way to all RPM panels (images).
    We consider only image-level augmentations, which do not change the order of RPM panels. The following augmentation methods are used: \emph{vertical flip}, \emph{horizontal flip}, \emph{rotation} by a random angle, \emph{transposition}, \emph{grid shuffle} and \emph{roll}.
    Grid shuffle transformation splits an image into a 2x2 or 3x3 grid and randomly shuffles its components, whereas the roll operation, rolls an image along the vertical axis, horizontal axis or both axes simultaneously.

\begin{figure*}[t]
        \centering
        \begin{subfigure}{.20\textwidth}
            \includegraphics[width=\textwidth]{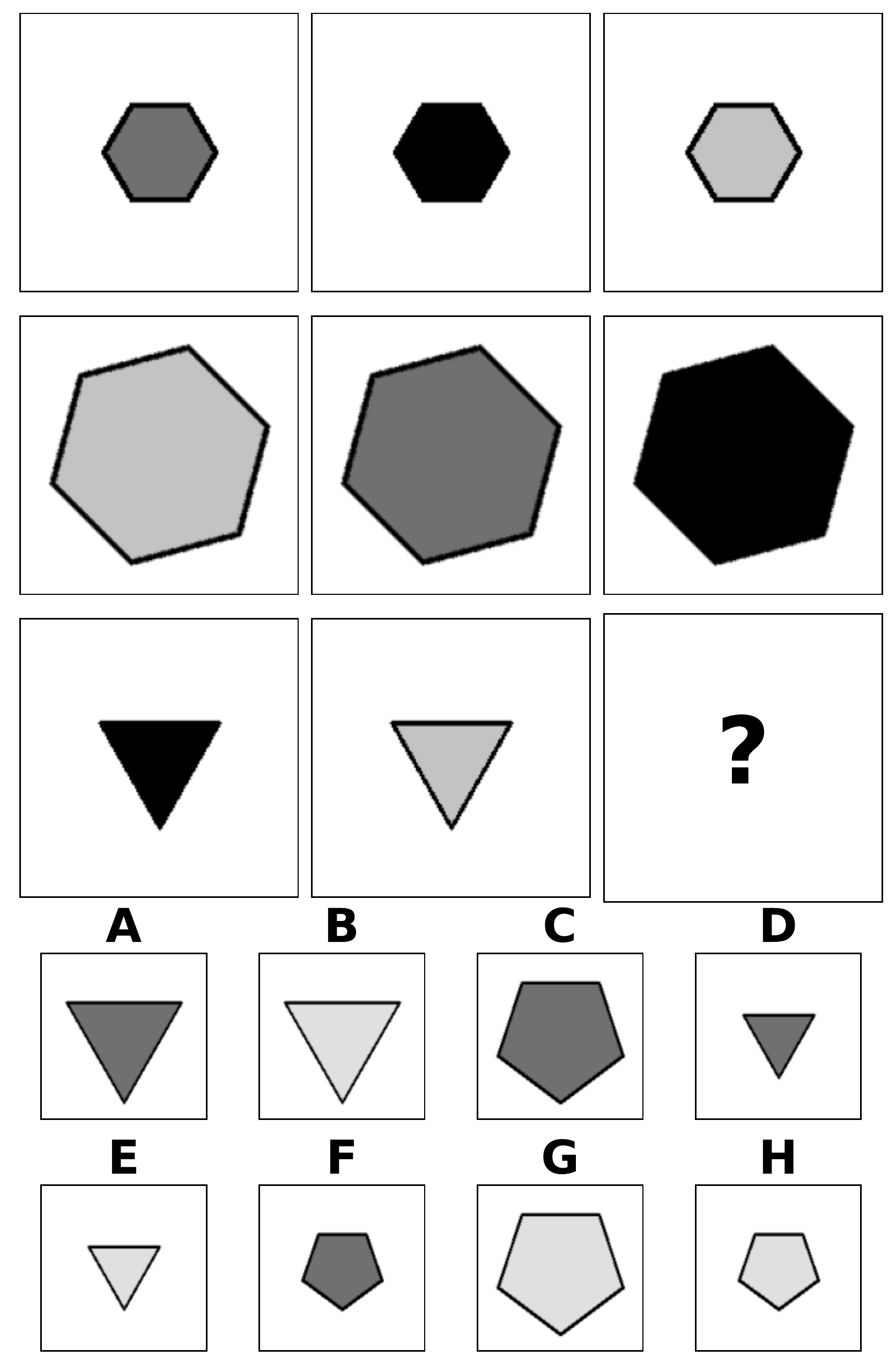}
            \caption{Base}
        \end{subfigure}
        ~
        \begin{subfigure}{.20\textwidth}
            \includegraphics[width=\textwidth]{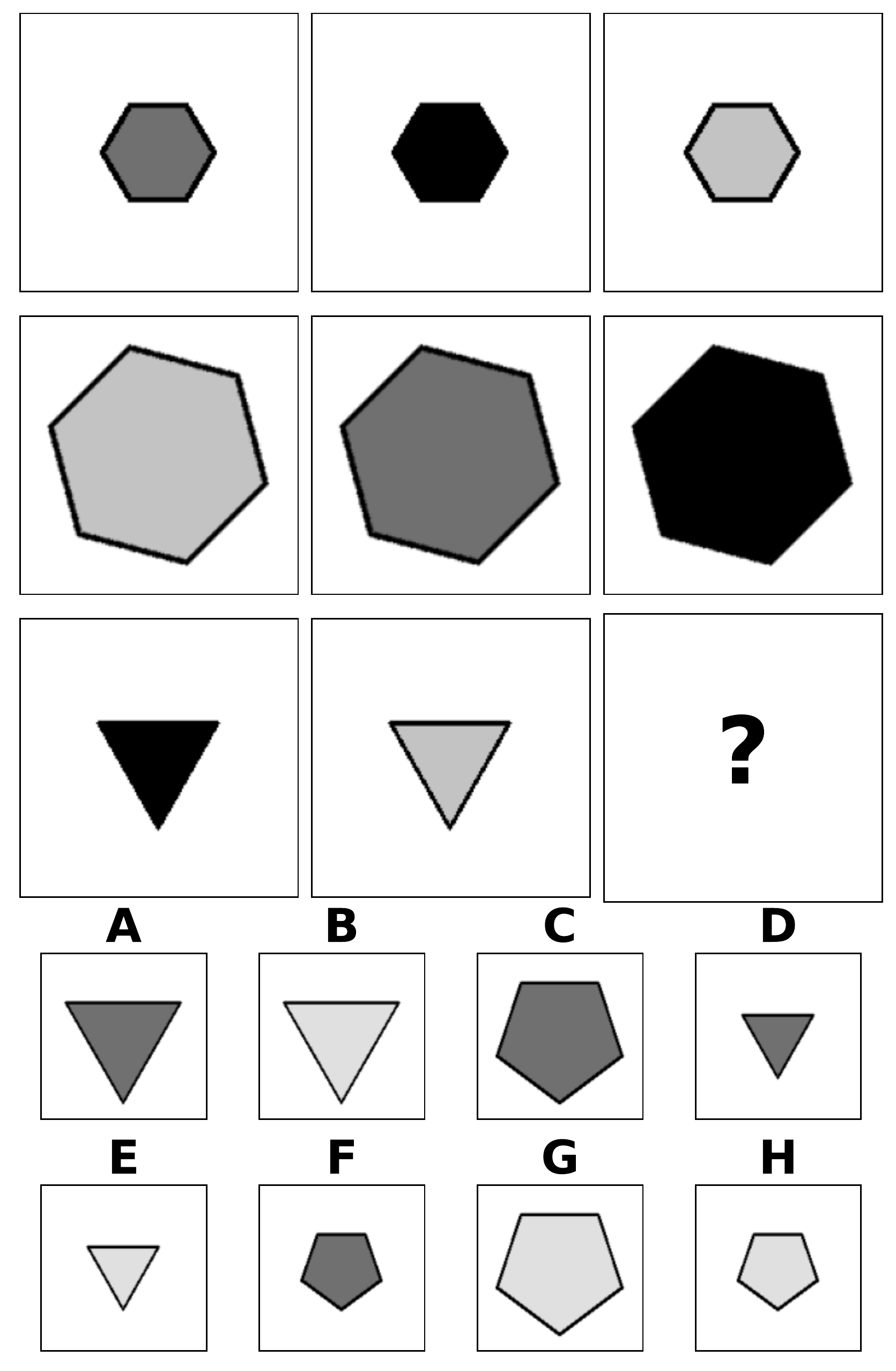}
            \caption{Horizontal flip}
        \end{subfigure}
        ~
        \begin{subfigure}{.20\textwidth}
            \includegraphics[width=\textwidth]{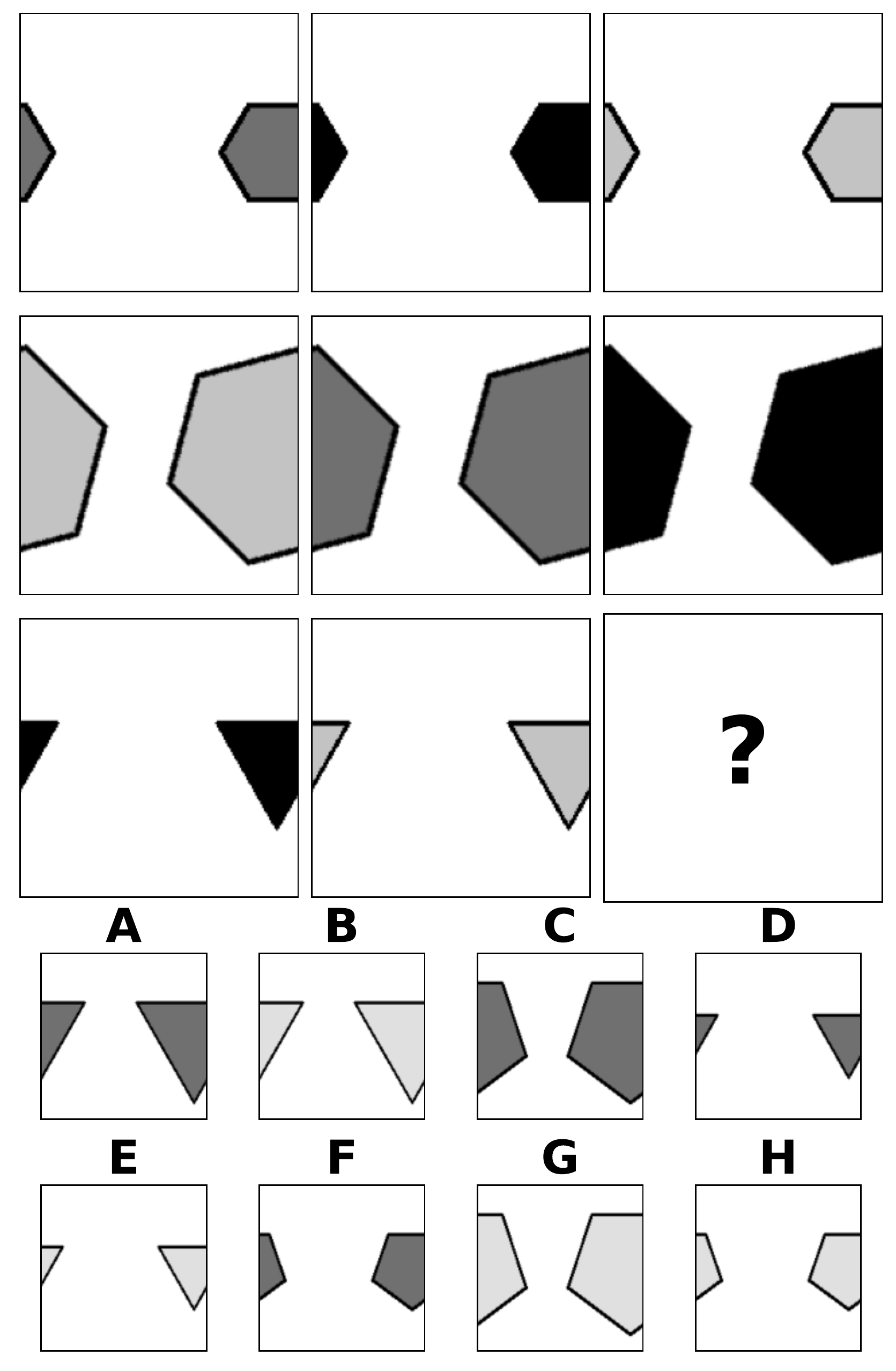}
            \caption{Horizontal roll}
        \end{subfigure}
        ~
        \begin{subfigure}{.20\textwidth}
            \includegraphics[width=\textwidth]{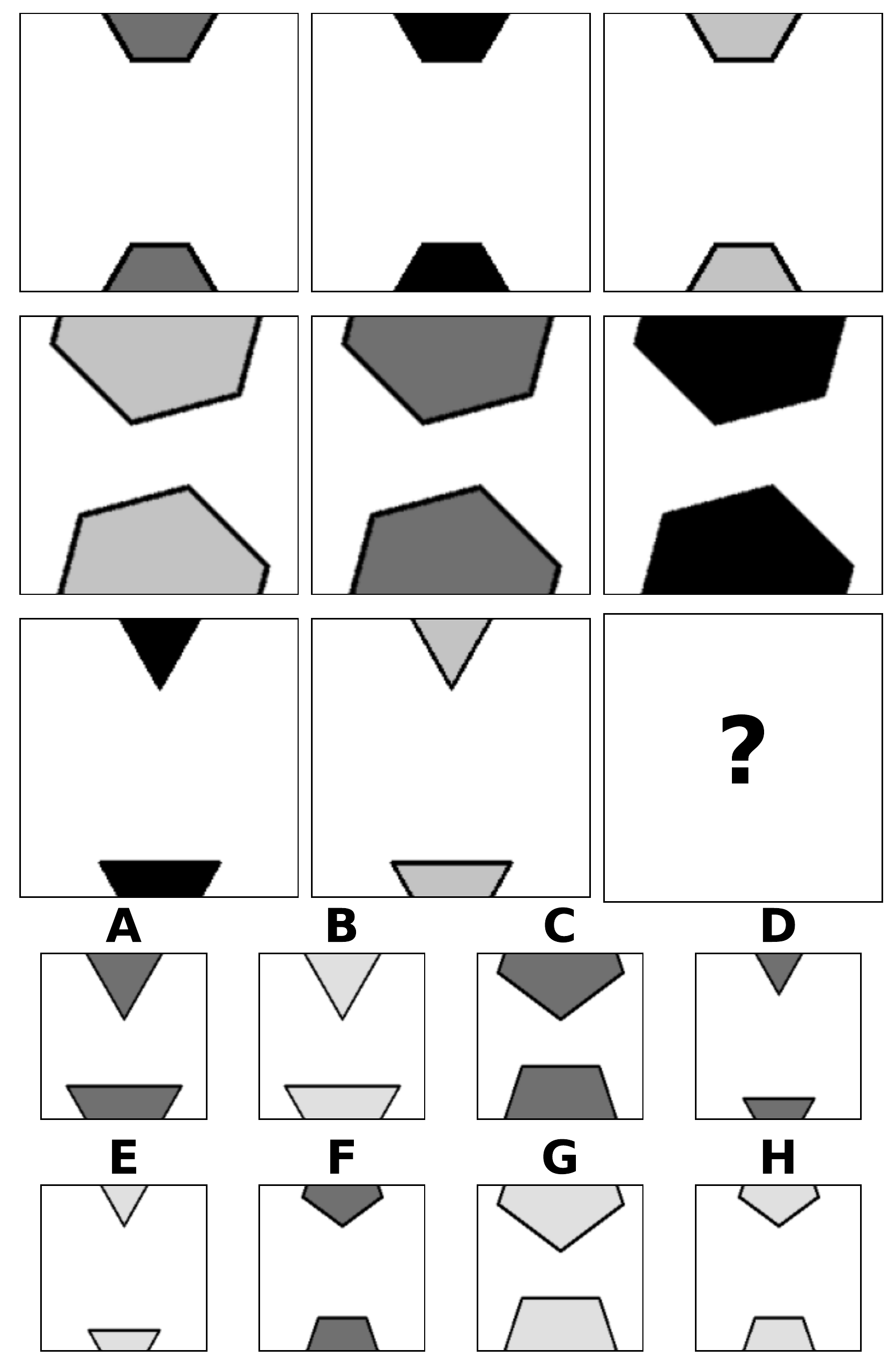}
            \caption{Shuffle 2x2}
        \end{subfigure}
        \caption{
        Augmented RPM from the Balanced-RAVEN dataset with configuration \texttt{Center}.
        In each of two top rows, there is a single shape in each image ($[\texttt{Constant},\texttt{Number}]$), located in the same position ($[\texttt{Constant},\texttt{Position}]$), the shapes are of the same type $[\texttt{Constant},\texttt{Type}]$ and size $[\texttt{Constant},\texttt{Size}]$.
        In each row there are shapes in 3 different colors ($[\texttt{Distribute\_Three},\texttt{Color}]$).
        This leads to an underlying abstract structure $\mathcal{S}$ $=$ $\{[\texttt{Constant},\texttt{Number}],$ $[\texttt{Constant},\texttt{Position}],$ $[\texttt{Constant},\texttt{Type}],$ $[\texttt{Constant},\texttt{Size}],$ $[\texttt{Distribute\_Three},\texttt{Color}]\}$, which is realized by completing the matrix with answer D.
        }
        \label{fig:augmentation-center}
    \end{figure*}

\begin{figure*}[t]
        \centering
        \begin{subfigure}{.20\textwidth}
            \includegraphics[width=\textwidth]{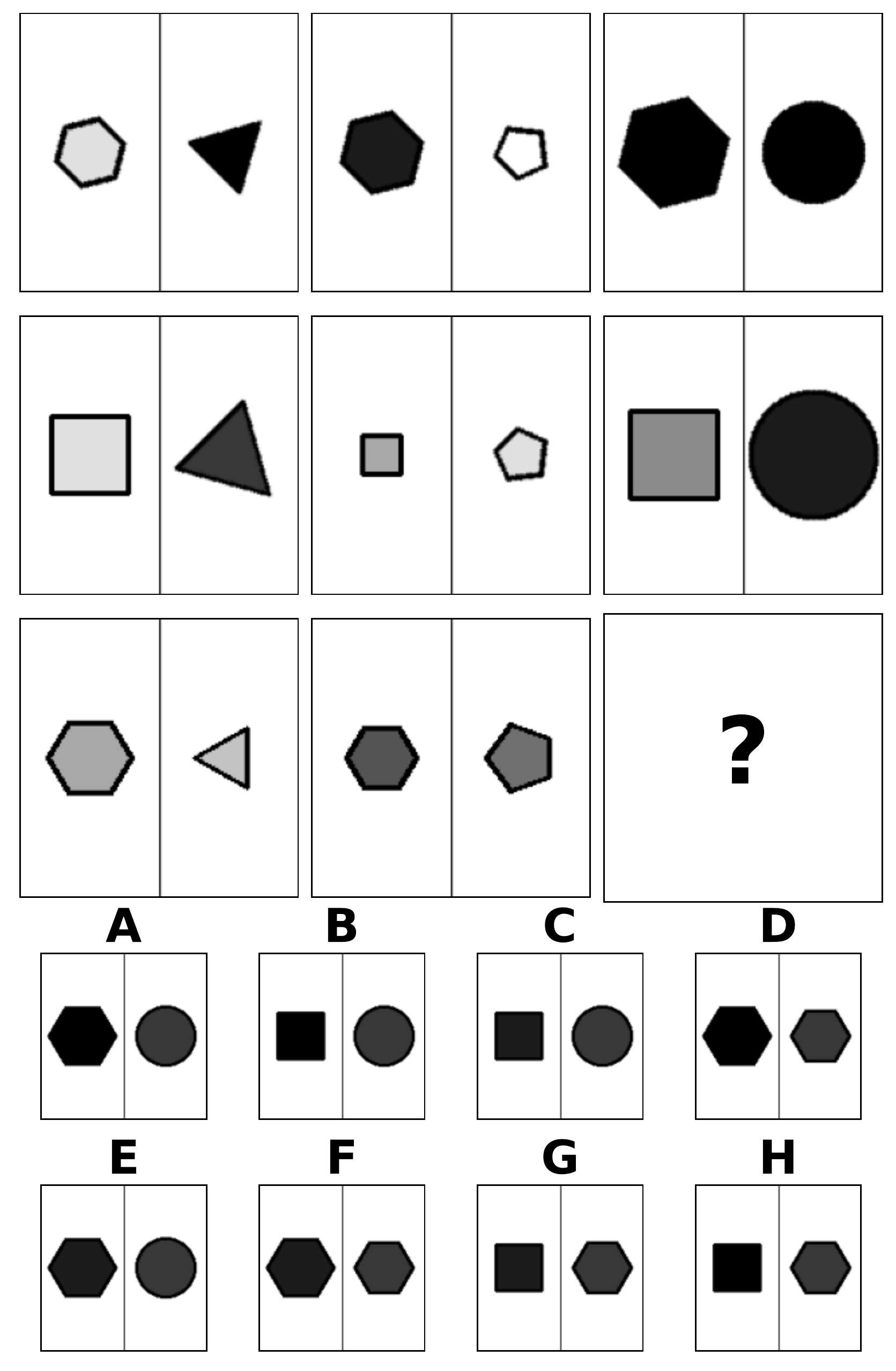}
            \caption{Base}
        \end{subfigure}
        ~
        \begin{subfigure}{.20\textwidth}
            \includegraphics[width=\textwidth]{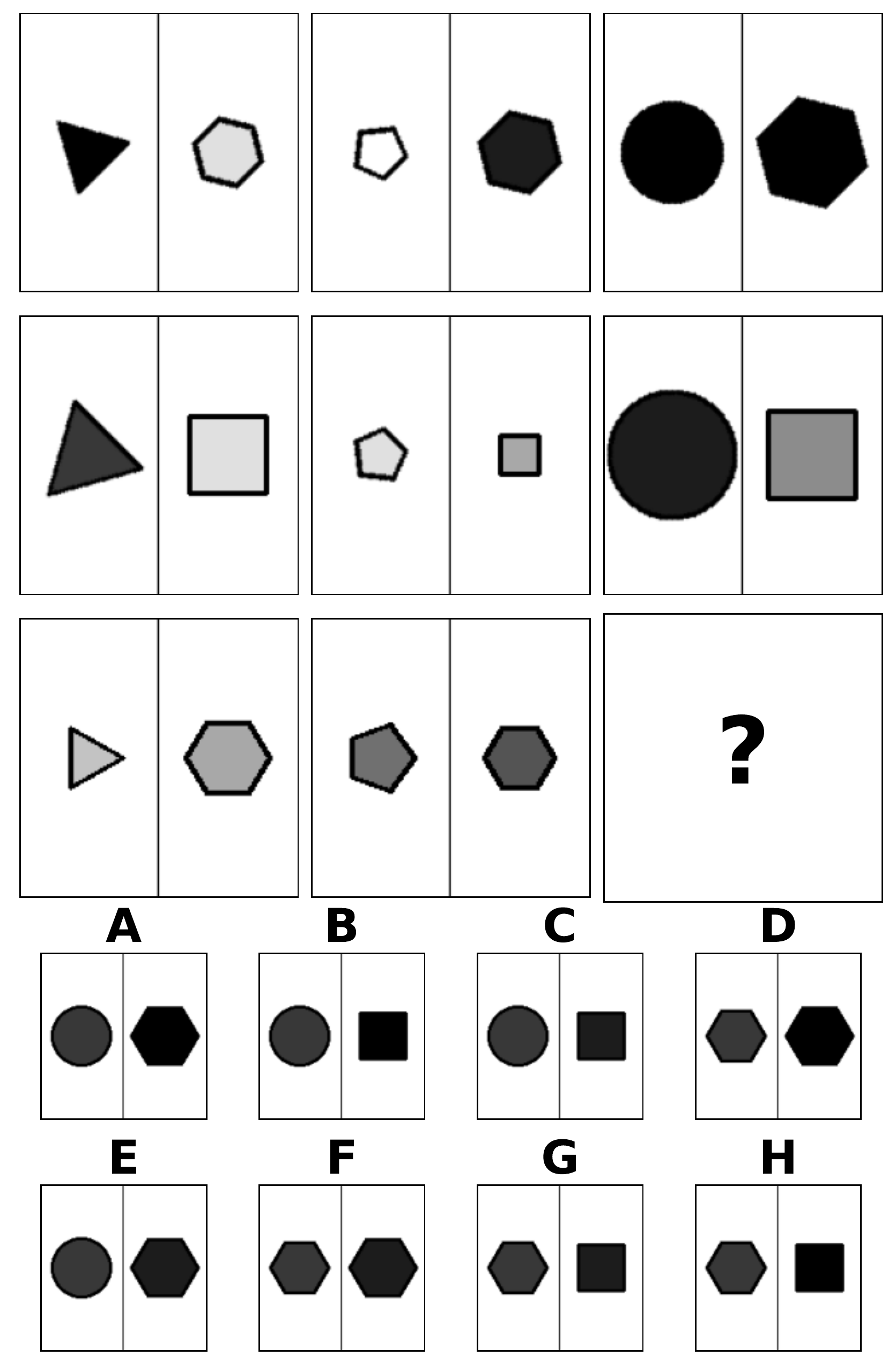}
            \caption{Horizontal flip}
        \end{subfigure}
        ~
        \begin{subfigure}{.20\textwidth}
            \includegraphics[width=\textwidth]{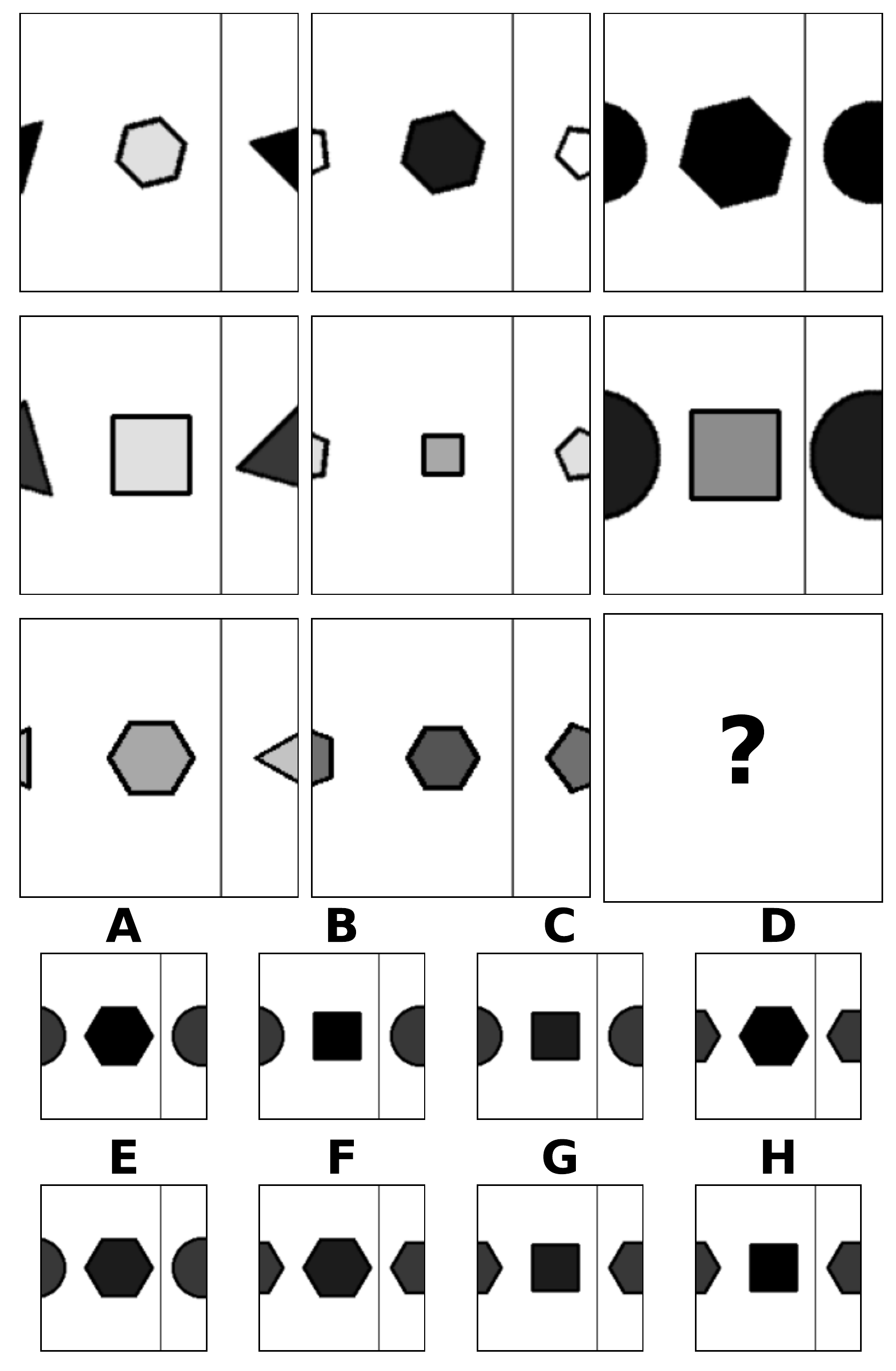}
            \caption{Horizontal roll}
        \end{subfigure}
        ~
        \begin{subfigure}{.20\textwidth}
            \includegraphics[width=\textwidth]{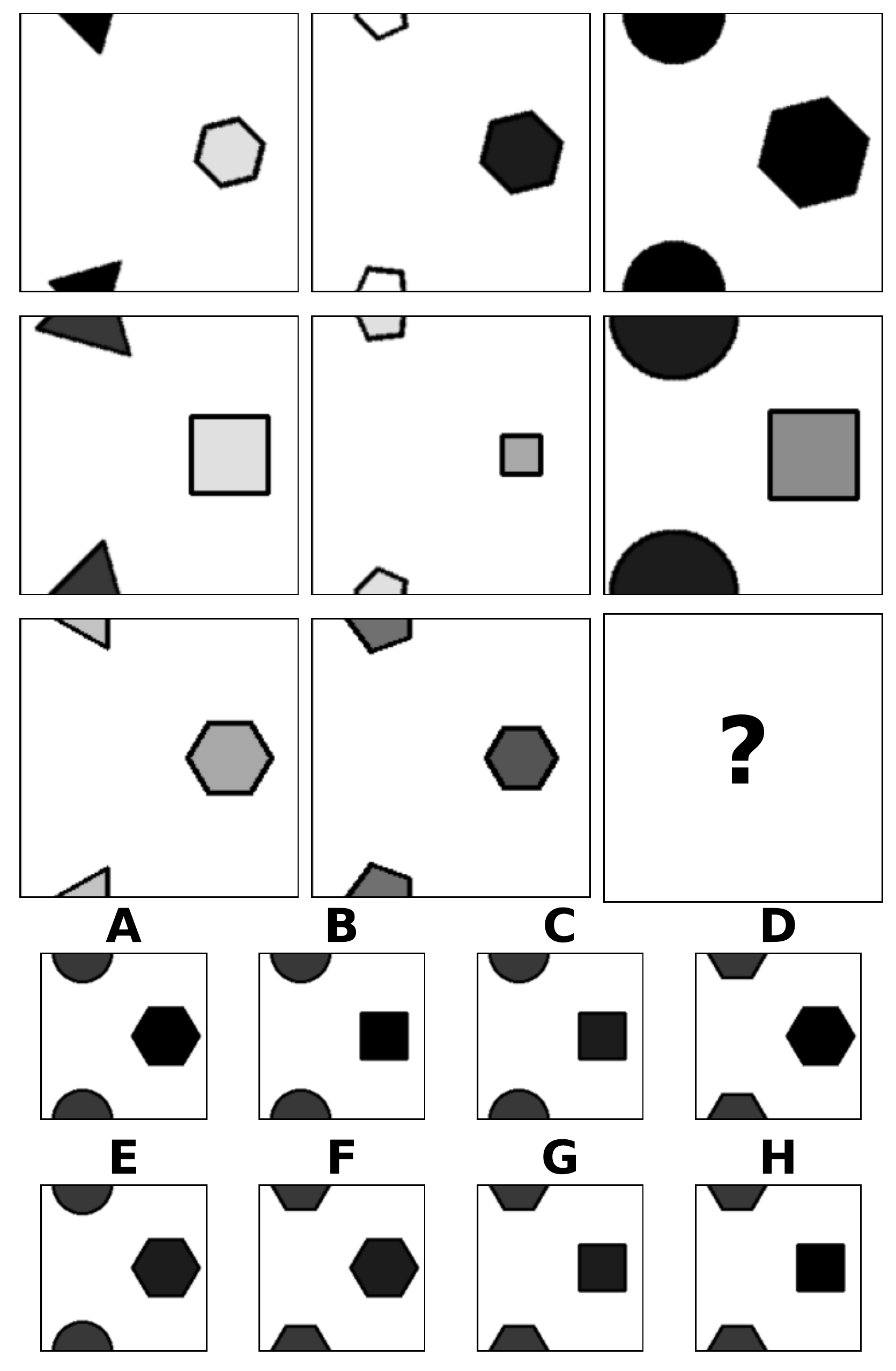}
            \caption{Shuffle 2x2}
        \end{subfigure}
        \caption{
        Augmented RPM from the Balanced-RAVEN dataset with configuration \texttt{L-R}.
        In this configuration, both left and right parts of the images can be governed by different sets of rules.
        Starting from the left parts of the panels, one can see that each part is composed of a single object ($[\texttt{Constant},\texttt{Number}]$), located in the centre ($[\texttt{Constant},\texttt{Position}]$), objects in each row are of the same type -- hexagons in the top row and squares in the middle row ($[\texttt{Constant},\texttt{Type}]$), the sizes and the colors of objects in the third column are equal to the sum of object sizes (respectively colors) from the first and the second column ($[\texttt{Arithmetic},\texttt{Size}]$) and ($[\texttt{Arithmetic},\texttt{Color}]$). These rules constitute the left structure $\mathcal{S}_{\mathrm{left}}$ $=$ $\{[\texttt{Constant},\texttt{Number}],$ $[\texttt{Constant},\texttt{Position}],$ $[\texttt{Constant},\texttt{Type}],$ $[\texttt{Arithmetic},\texttt{Size}],$ $[\texttt{Arithmetic},\texttt{Color}]\}$.
        Next, looking at the right subpanels of each image, it can be observed that each part contains a single object ($[\texttt{Constant},\texttt{Number}]$), located in the center ($[\texttt{Constant},\texttt{Position}]$).
        In each row, types of objects (from right subpanels) follow progression ($[\texttt{Progression},\texttt{Type}]$).
        Similarly to the left subpanels, sizes and colors of object in the third column can be calculated by summing up the respective attribute values of objects in the two preceding columns ($[\texttt{Arithmetic},\texttt{Size}]$ and $[\texttt{Arithmetic},\texttt{Color}]$).
        This leads to the right structure defined as $\mathcal{S}_{\mathrm{right}}$ $=$ $\{[\texttt{Constant},\texttt{Number}],$ $[\texttt{Constant},\texttt{Position}],$ $[\texttt{Progression},\texttt{Type}],$ $[\texttt{Arithmetic},\texttt{Size}],$ $[\texttt{Arithmetic},\texttt{Color}]\}$.
        The final underlying abstract structure of the whole RPM is the sum of both left and right structures $\mathcal{S}$ $=$ $\mathcal{S}_{\mathrm{left}}$ $\cup$ $\mathcal{S}_{\mathrm{right}}$.
        A is the only answer which satisfies all 10 of the above-discussed rules.
        }
        \label{fig:augmentation-leftright}
    \end{figure*}

\begin{figure*}[t]
        \centering
        \begin{subfigure}{.20\textwidth}
            \includegraphics[width=\textwidth]{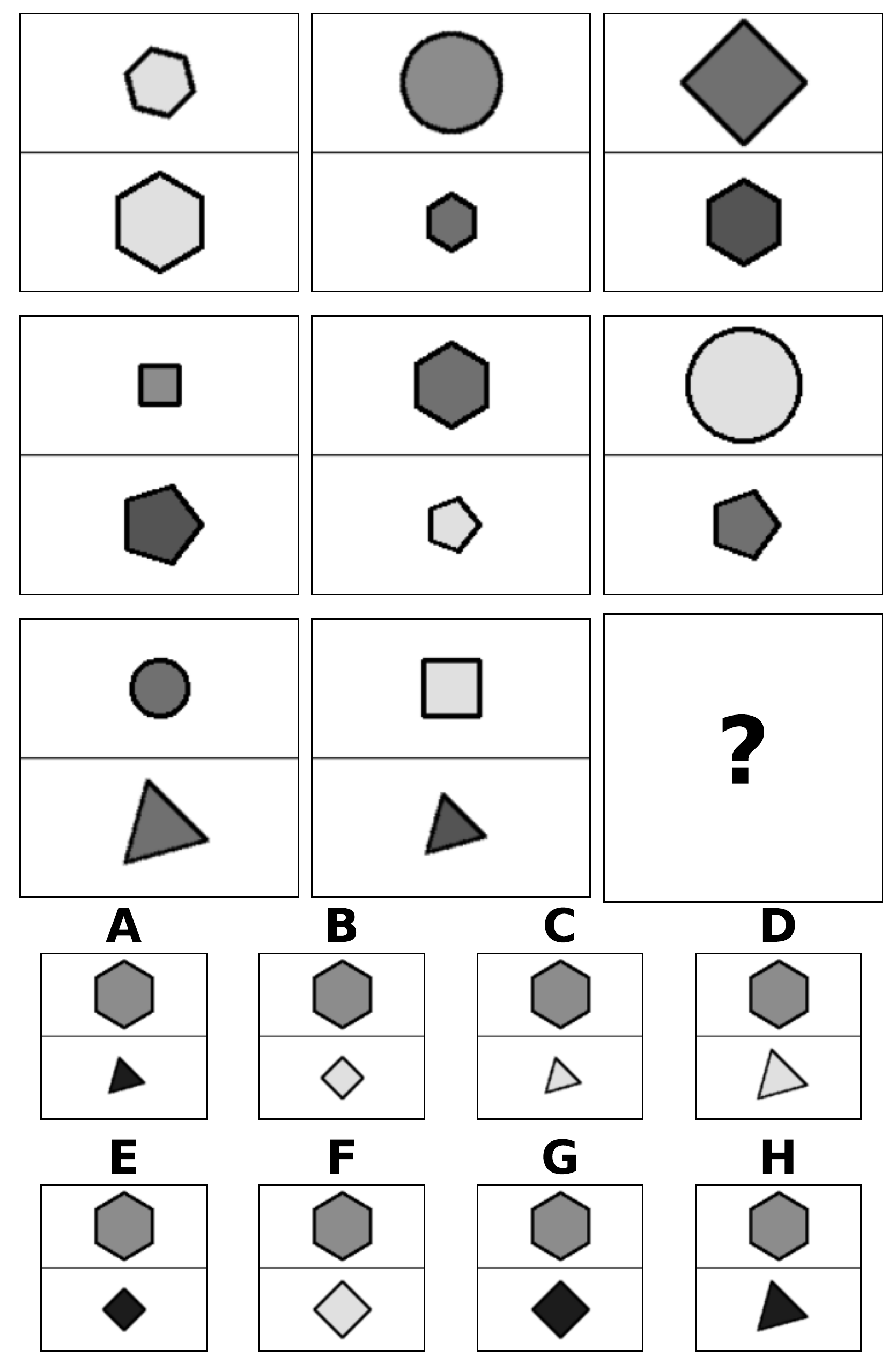}
            \caption{Base}
        \end{subfigure}
        ~
        \begin{subfigure}{.20\textwidth}
            \includegraphics[width=\textwidth]{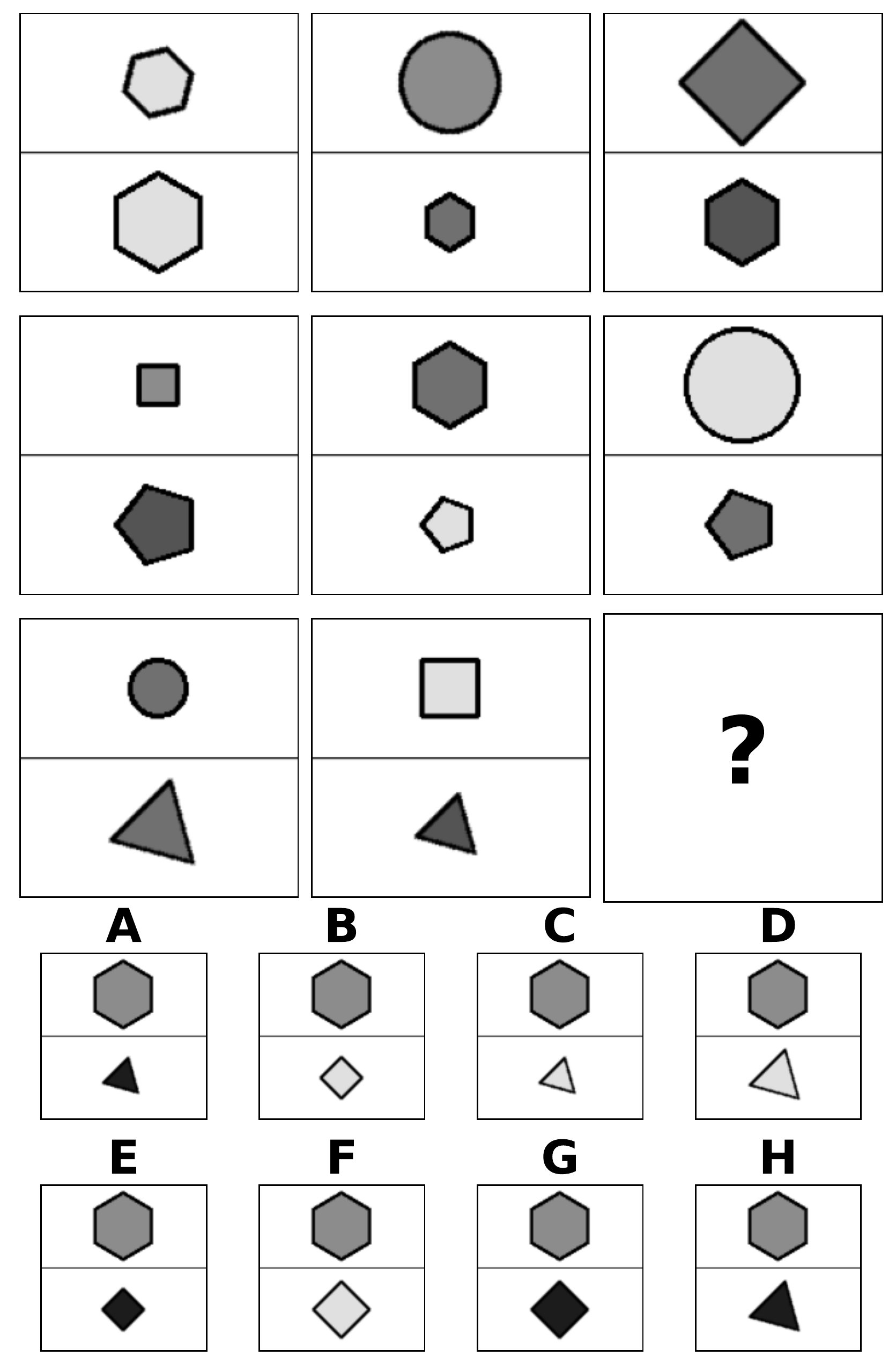}
            \caption{Horizontal flip}
        \end{subfigure}
        ~
        \begin{subfigure}{.20\textwidth}
            \includegraphics[width=\textwidth]{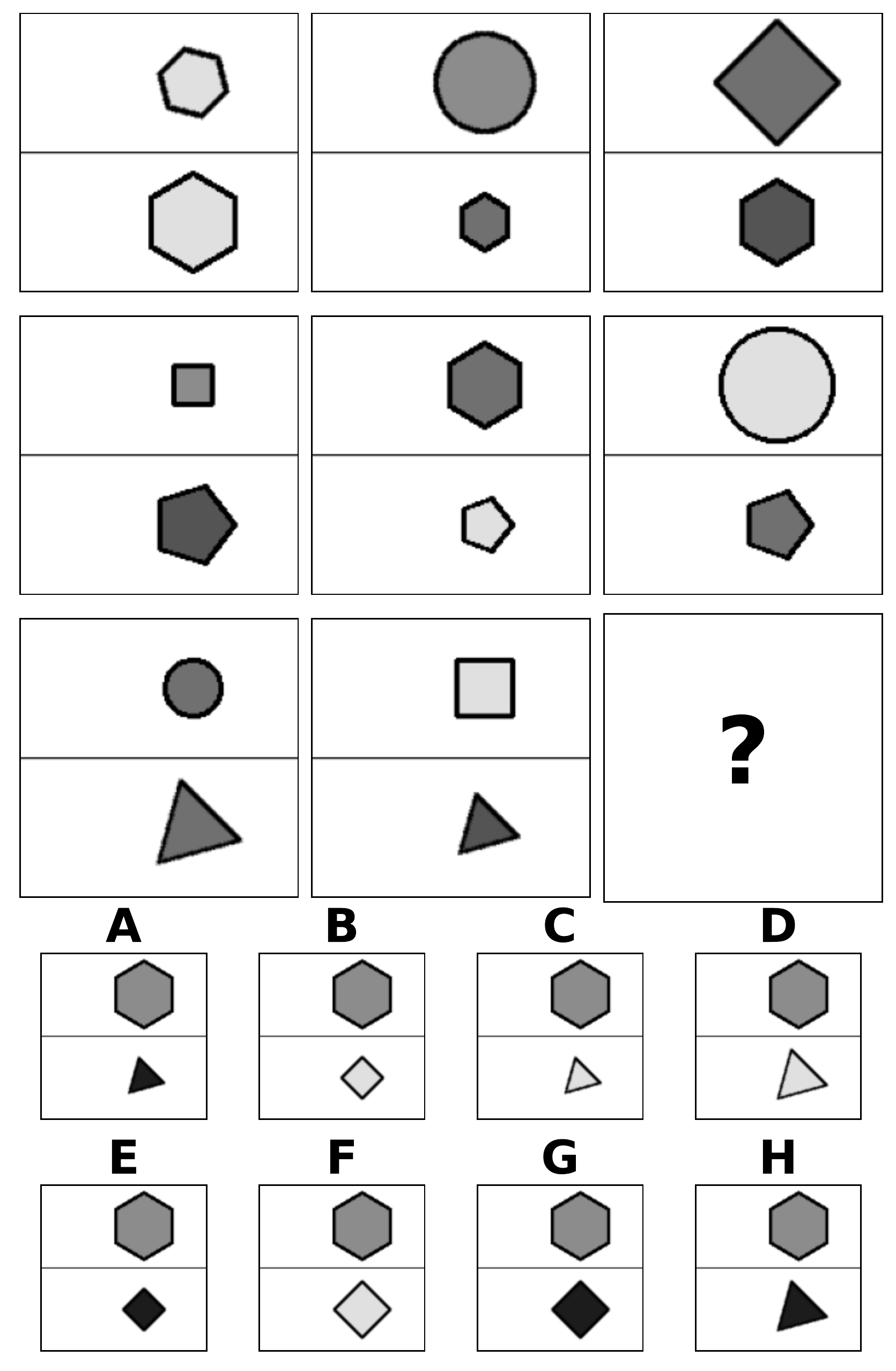}
            \caption{Horizontal roll}
        \end{subfigure}
        ~
        \begin{subfigure}{.20\textwidth}
            \includegraphics[width=\textwidth]{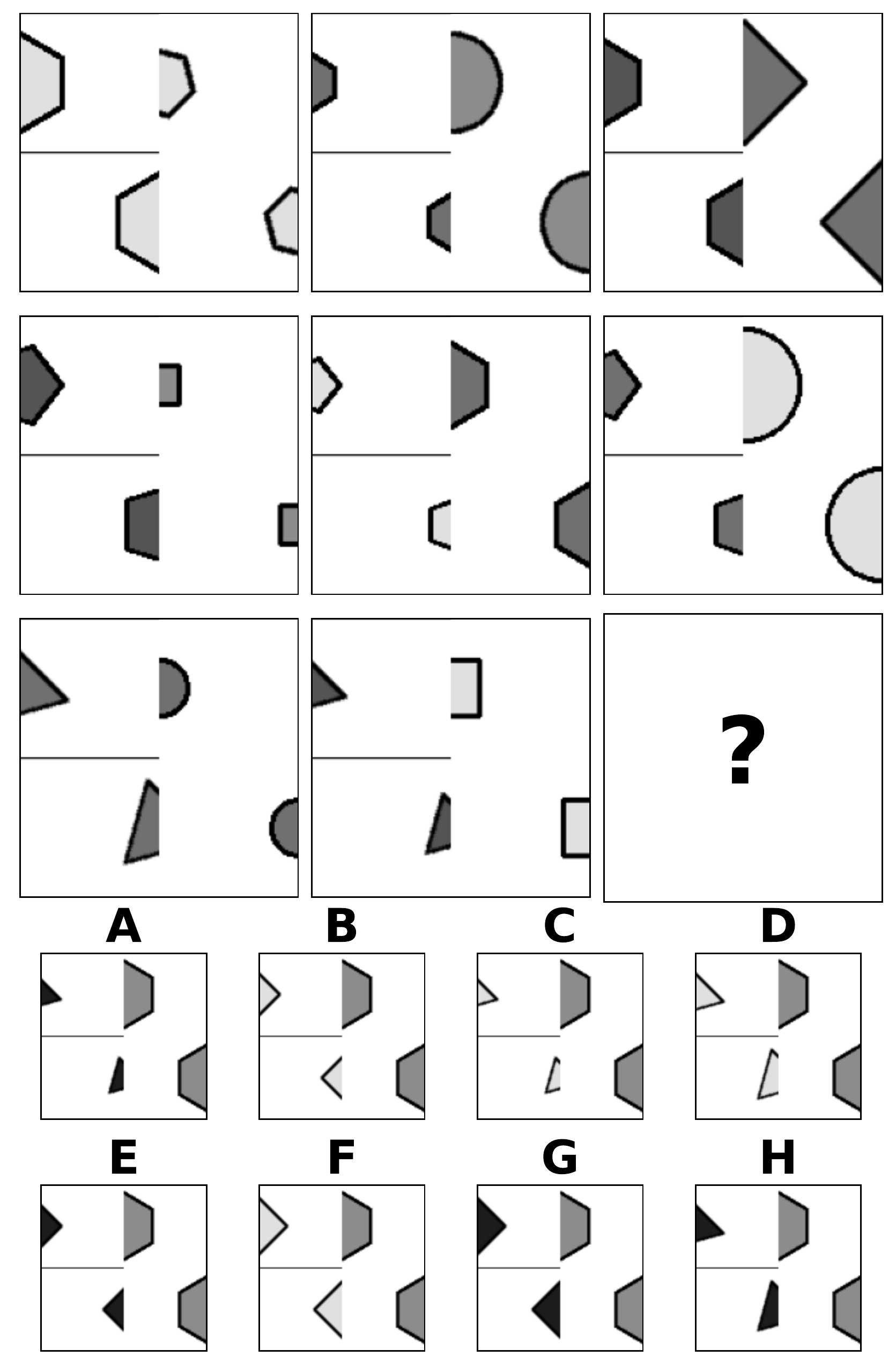}
            \caption{Shuffle 2x2}
        \end{subfigure}
        \caption{
        Augmented RPM from the Balanced-RAVEN dataset with configuration \texttt{U-D}.
        Similarly to configuration \texttt{L-R}, both upper and lower parts can be governed by different sets of rules.
        The abstract structure of upper image parts is $\mathcal{S}_{\mathrm{upper}}$ $=$ $\{[\texttt{Constant},\texttt{Number}],$ $[\texttt{Constant},\texttt{Position}],$ $[\texttt{Distribute\_Three},\texttt{Type}],$ $[\texttt{Progression},\texttt{Size}],$ $[\texttt{Distribute\_Three},\texttt{Color}]\}$ and the set of rules of the lower parts is $\mathcal{S}_{\mathrm{lower}}$ $=$ $\{[\texttt{Constant},\texttt{Number}],$ $[\texttt{Constant},\texttt{Position}],$ $[\texttt{Constant},\texttt{Type}],$ $[\texttt{Arithmetic},\texttt{Size}],$ $[\texttt{Distribute\_Three},\texttt{Color}]\}$.
        The final set of rules $\mathcal{S}$ is defined as $\mathcal{S}$ $=$ $\mathcal{S}_{\mathrm{upper}}$ $+$ $\mathcal{S}_{\mathrm{lower}}$. The correct answer is C.
        }
        \label{fig:augmentation-updown}
    \end{figure*}

\begin{figure*}[t]
        \centering
        \begin{subfigure}{.20\textwidth}
            \includegraphics[width=\textwidth]{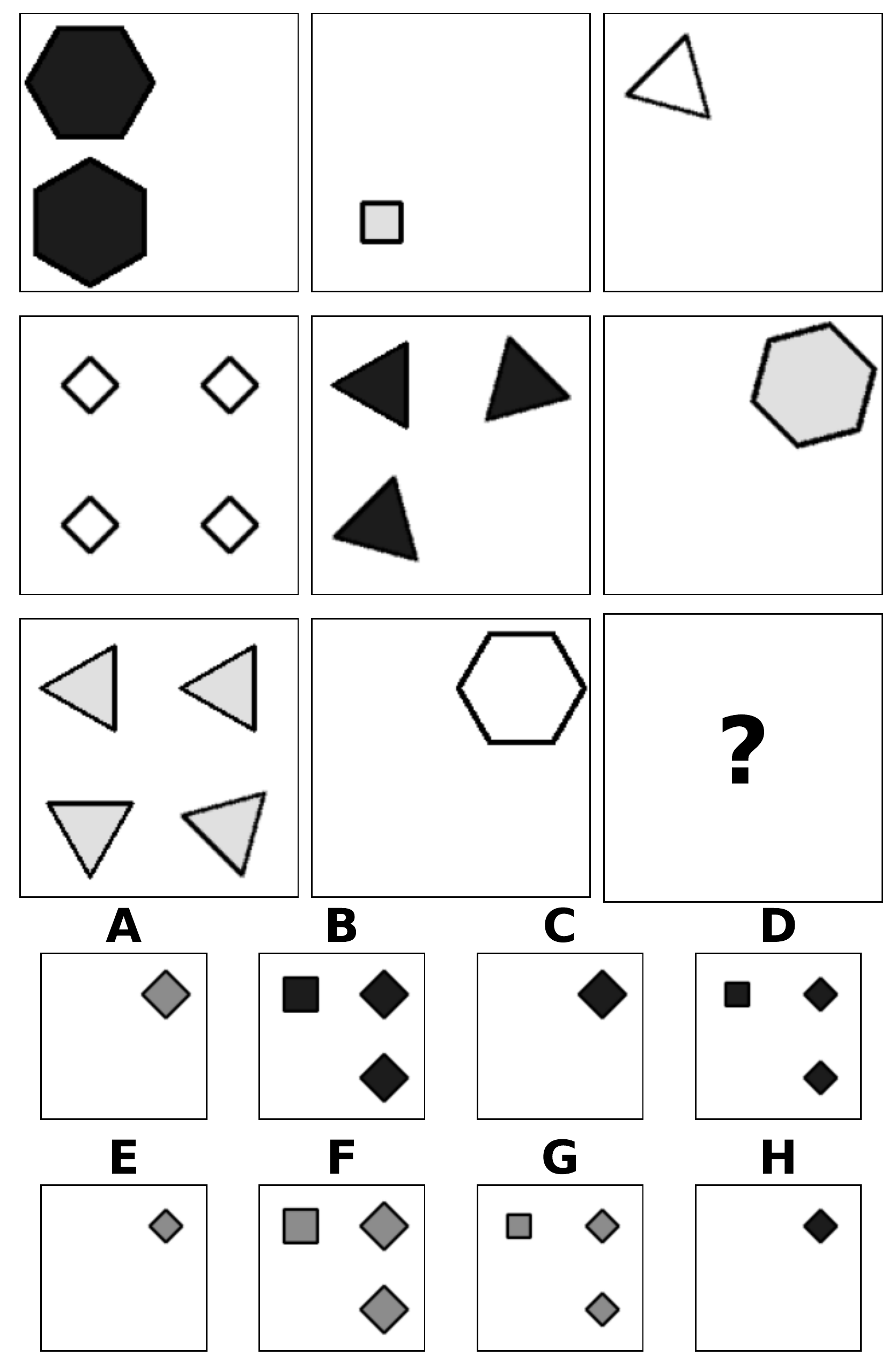}
            \caption{Base}
        \end{subfigure}
        ~
        \begin{subfigure}{.20\textwidth}
            \includegraphics[width=\textwidth]{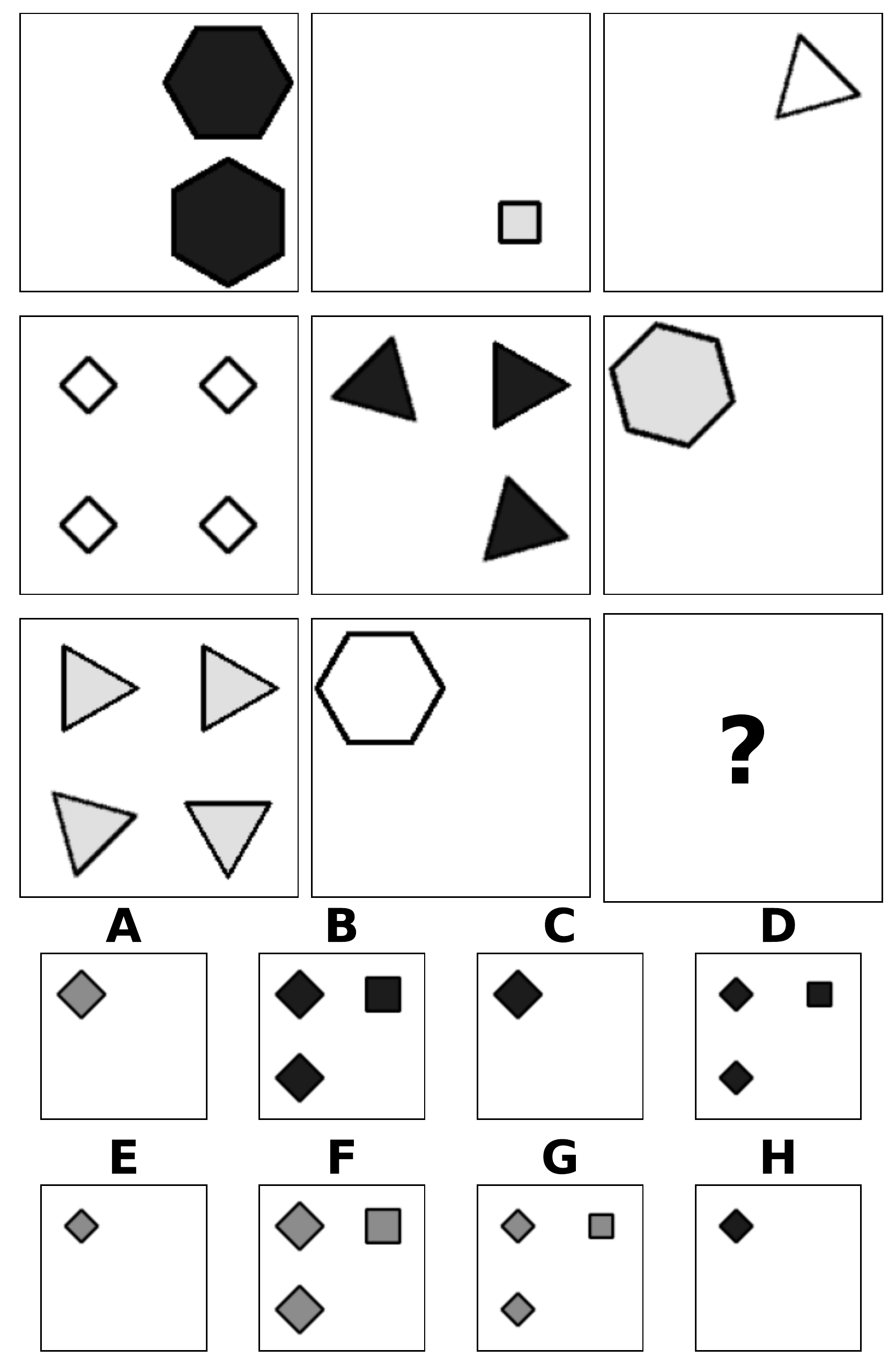}
            \caption{Horizontal flip}
        \end{subfigure}
        ~
        \begin{subfigure}{.20\textwidth}
            \includegraphics[width=\textwidth]{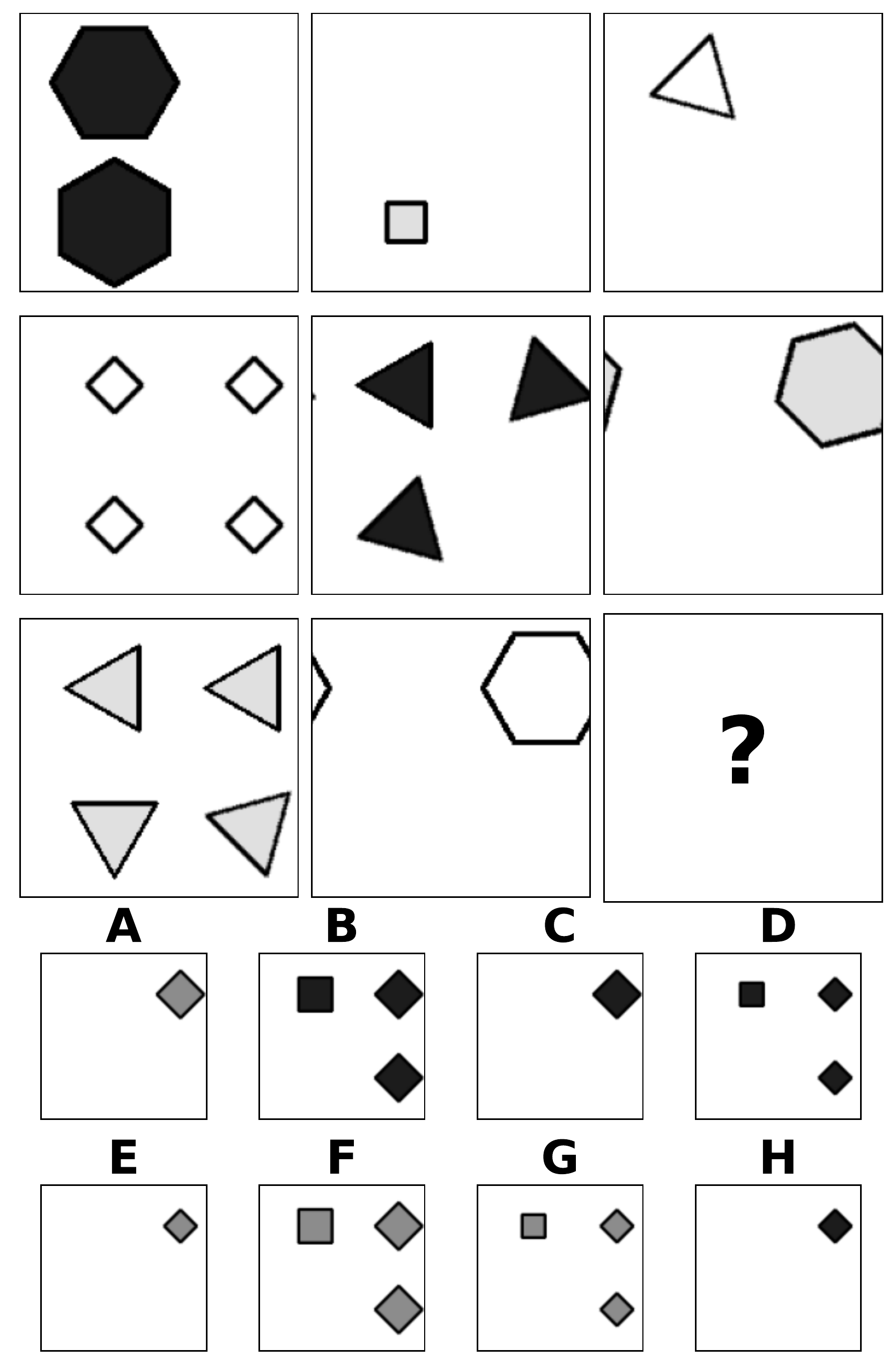}
            \caption{Horizontal roll}
        \end{subfigure}
        ~
        \begin{subfigure}{.20\textwidth}
            \includegraphics[width=\textwidth]{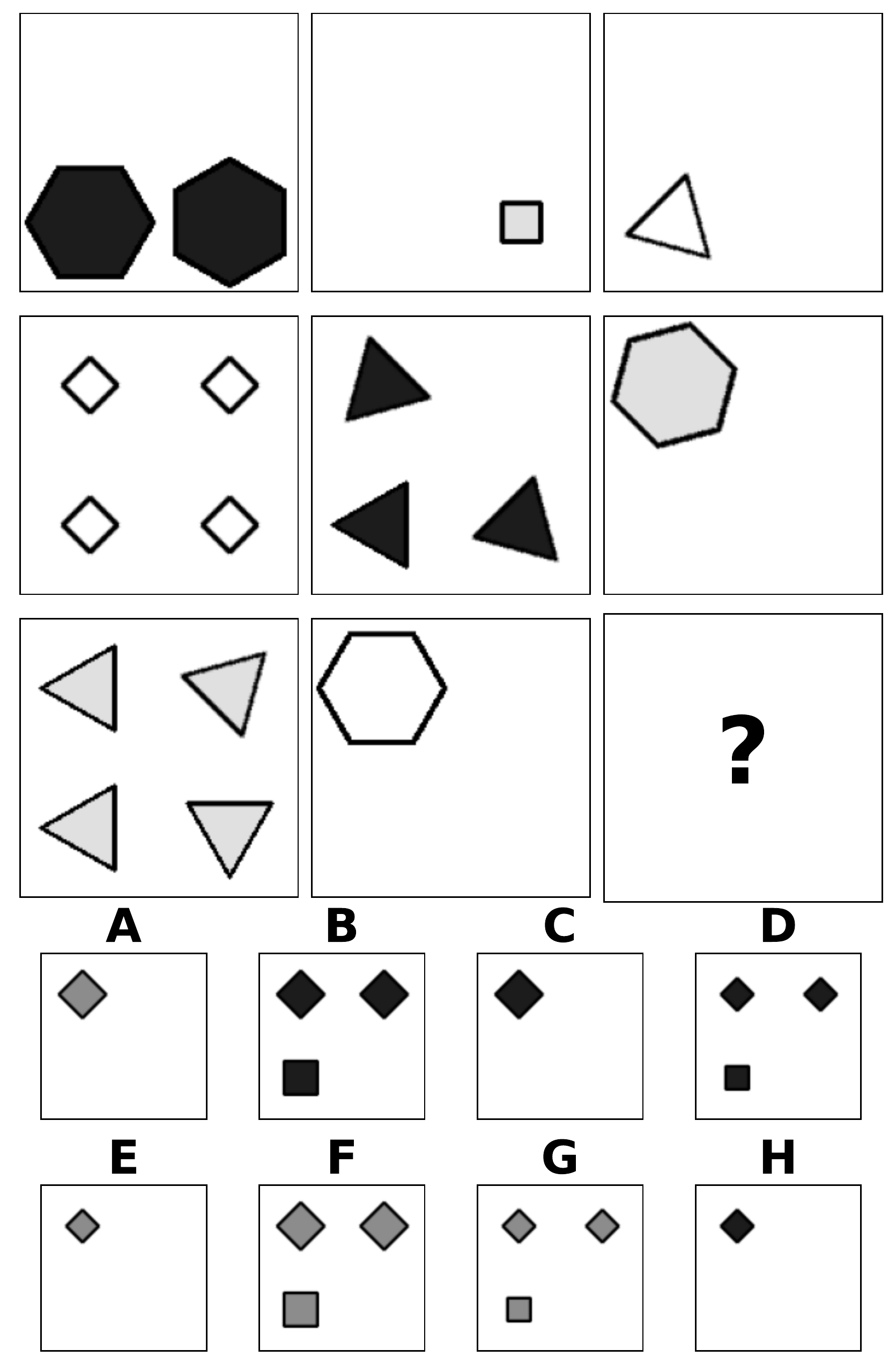}
            \caption{Shuffle 2x2}
        \end{subfigure}
        \caption{
        Augmented RPM from the Balanced-RAVEN dataset with configuration \texttt{2x2Grid}.
        The number of objects in the third column can be calculated by subtracting the number of objects in the second column from the number of objects present in the first column ($[\texttt{Arithmetic},\texttt{Number}]$).
        In rows each panel contains objects with one out of three unique types, sizes and colors ($[\texttt{Distribute\_Three},\texttt{Type}],$ $[\texttt{Distribute\_Three},\texttt{Size}],$ $[\texttt{Distribute\_Three},\texttt{Color}]$).
        These relations lead to an underlying abstract structure defined as $\mathcal{S}$ $=$ $\{[\texttt{Arithmetic},\texttt{Number}],$ $[\texttt{Distribute\_Three},\texttt{Type}],$ $[\texttt{Distribute\_Three},\texttt{Size}],$ $[\texttt{Distribute\_Three},\texttt{Color}]\}$.
        The only correct answer which satisfies all the above rules is D.
        }
        \label{fig:augmentation-2x2}
    \end{figure*}

\begin{figure*}[t]
        \centering
        \begin{subfigure}{.20\textwidth}
            \includegraphics[width=\textwidth]{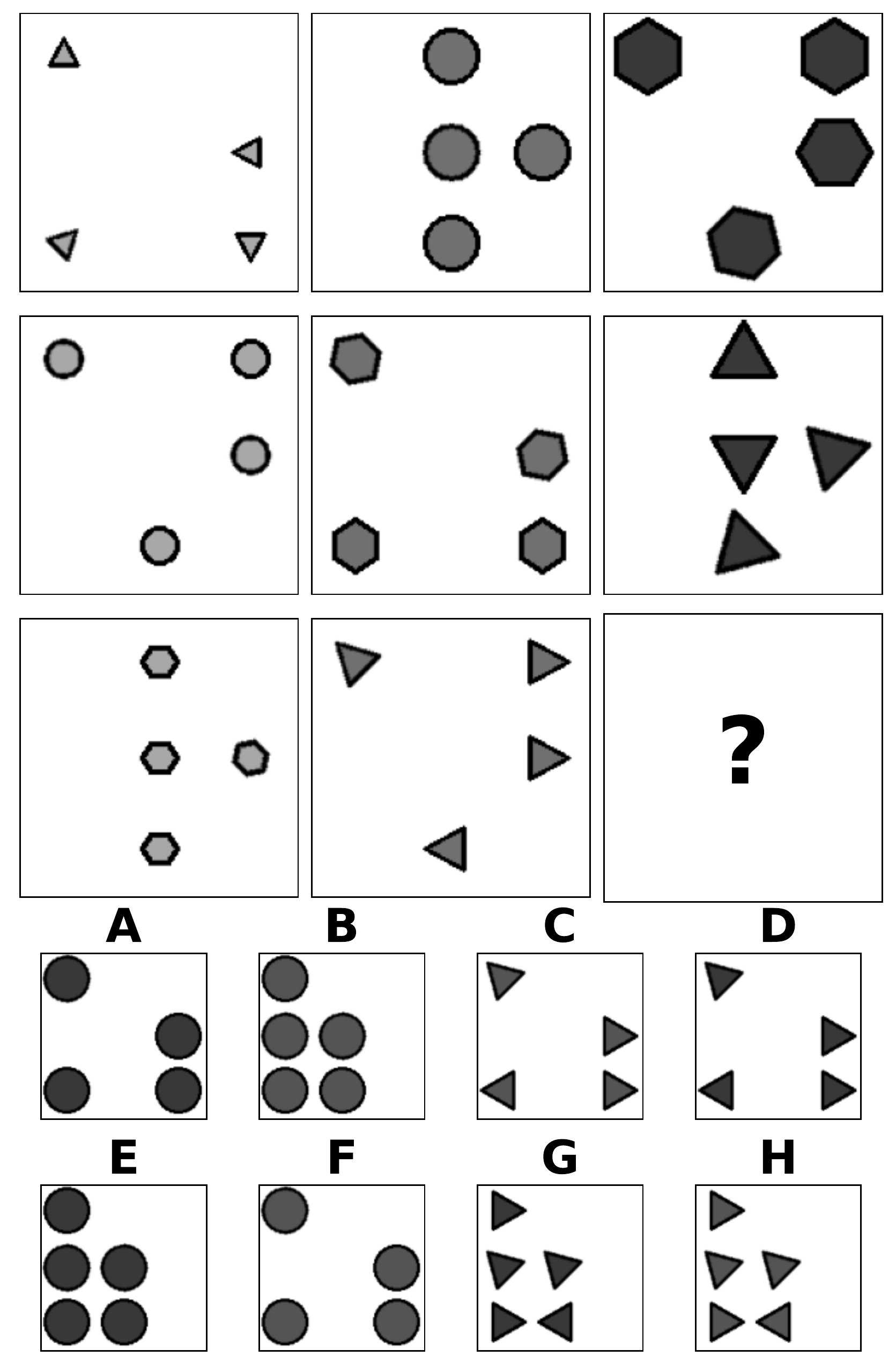}
            \caption{Base}
        \end{subfigure}
        ~
        \begin{subfigure}{.20\textwidth}
            \includegraphics[width=\textwidth]{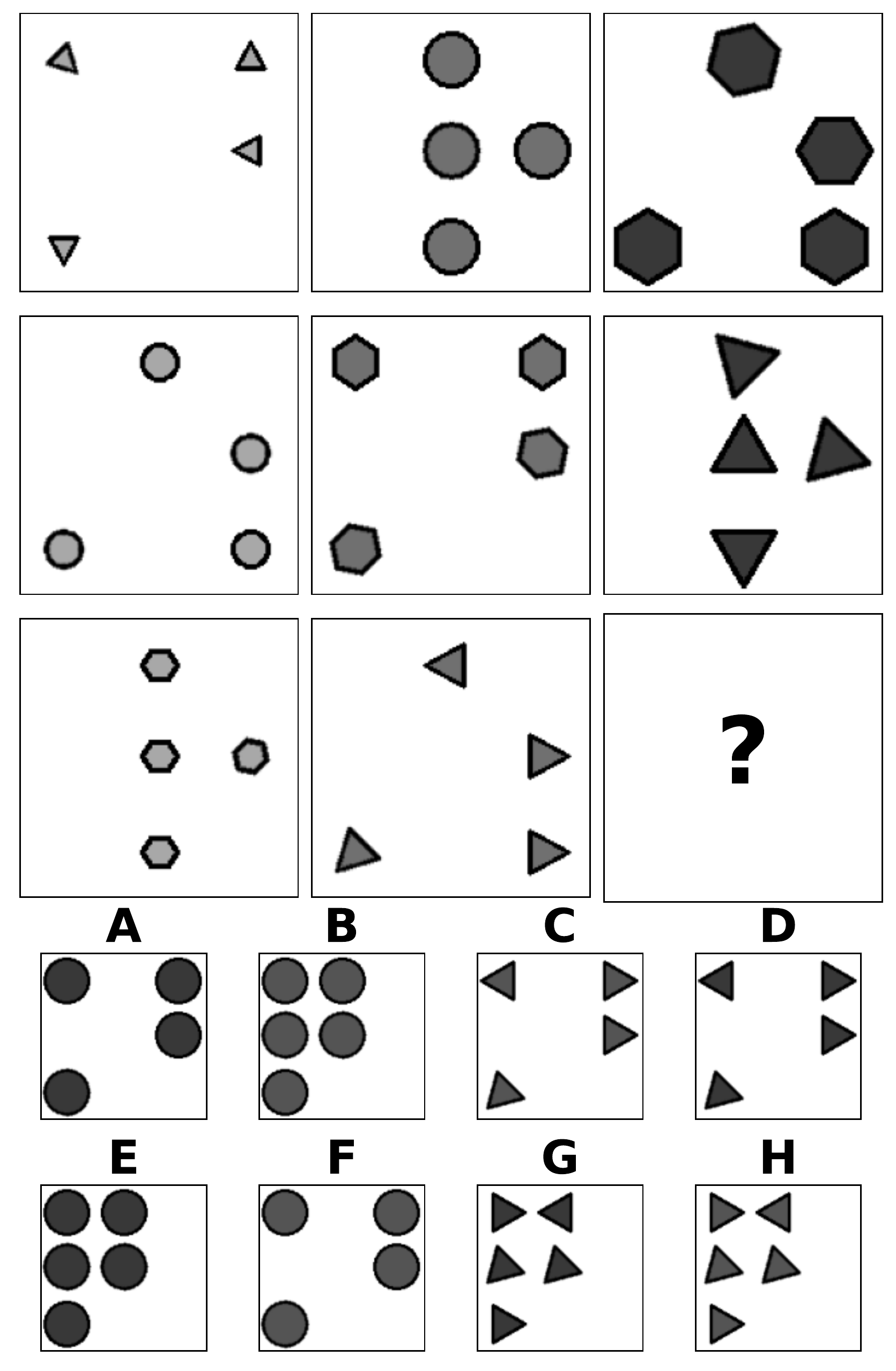}
            \caption{Vertical flip}
        \end{subfigure}
        ~
        \begin{subfigure}{.20\textwidth}
            \includegraphics[width=\textwidth]{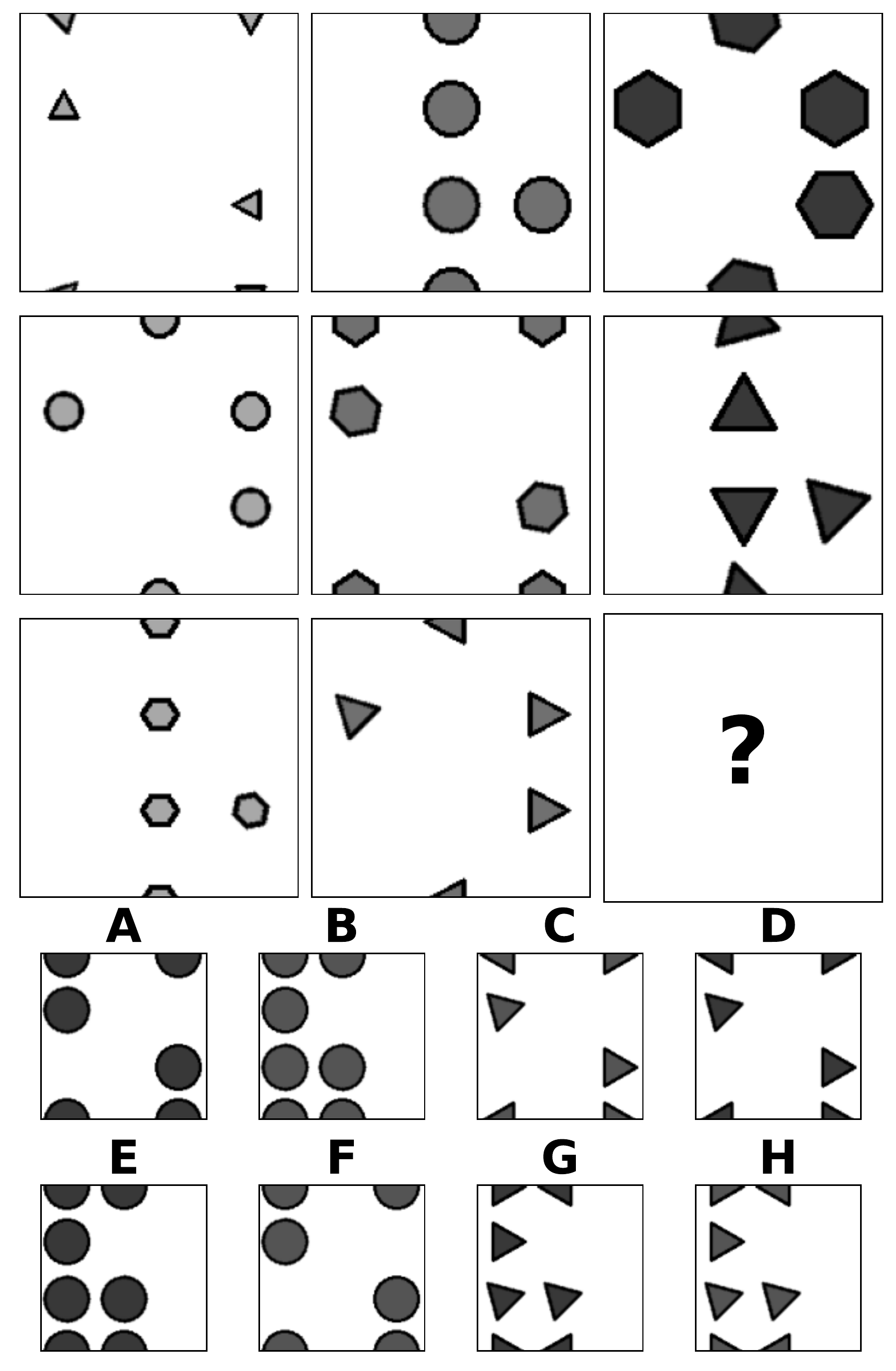}
            \caption{Vertical roll}
        \end{subfigure}
        ~
        \begin{subfigure}{.20\textwidth}
            \includegraphics[width=\textwidth]{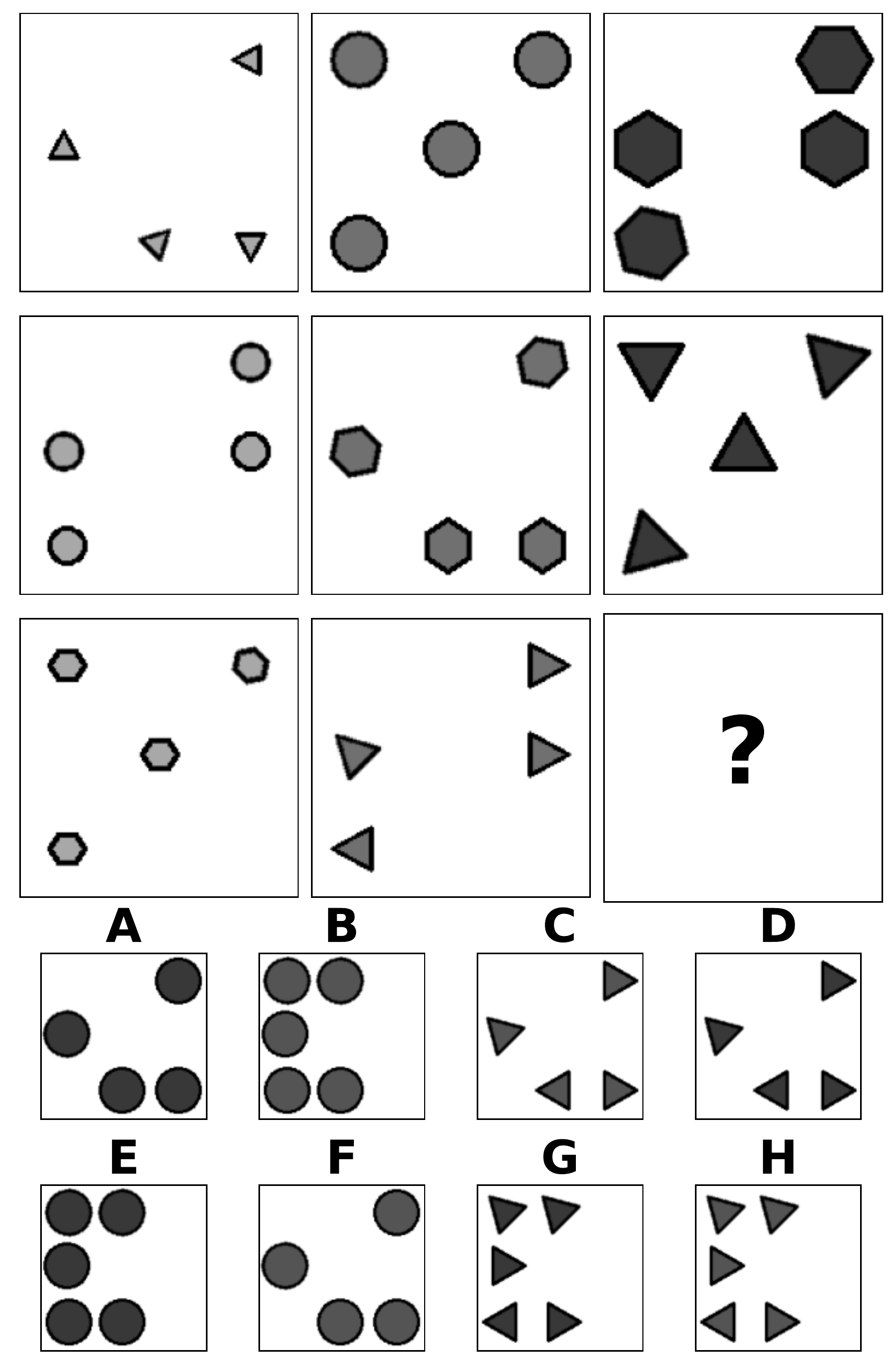}
            \caption{Shuffle 3x3}
        \end{subfigure}
        \caption{
        Augmented RPM from the Balanced-RAVEN dataset with configuration \texttt{3x3Grid}.
        In each row, the objects are arranged in one out of three available positions and each image contains a unique type of objects ($[\texttt{Distribute\_Three},\texttt{Position}]$ and $[\texttt{Distribute\_Three},\texttt{Type}]$).
        Moreover, in each row, from left to right, the objects both increase in size and become darker ($[\texttt{Progression},\texttt{Size}]$ and $[\texttt{Progression},\texttt{Color}]$).
        Therefore, the underlying abstract structure is of the form $\mathcal{S}$ $=$ $\{[\texttt{Distribute\_Three},\texttt{Position}], [\texttt{Distribute\_Three},\texttt{Type}],$ $[\texttt{Progression},\texttt{Size}],$ $[\texttt{Progression},\texttt{Color}]\}$ and the correct answer is A.
        }
        \label{fig:augmentation-3x3}
    \end{figure*}

\begin{figure*}[t]
        \centering
        \begin{subfigure}{.20\textwidth}
            \includegraphics[width=\textwidth]{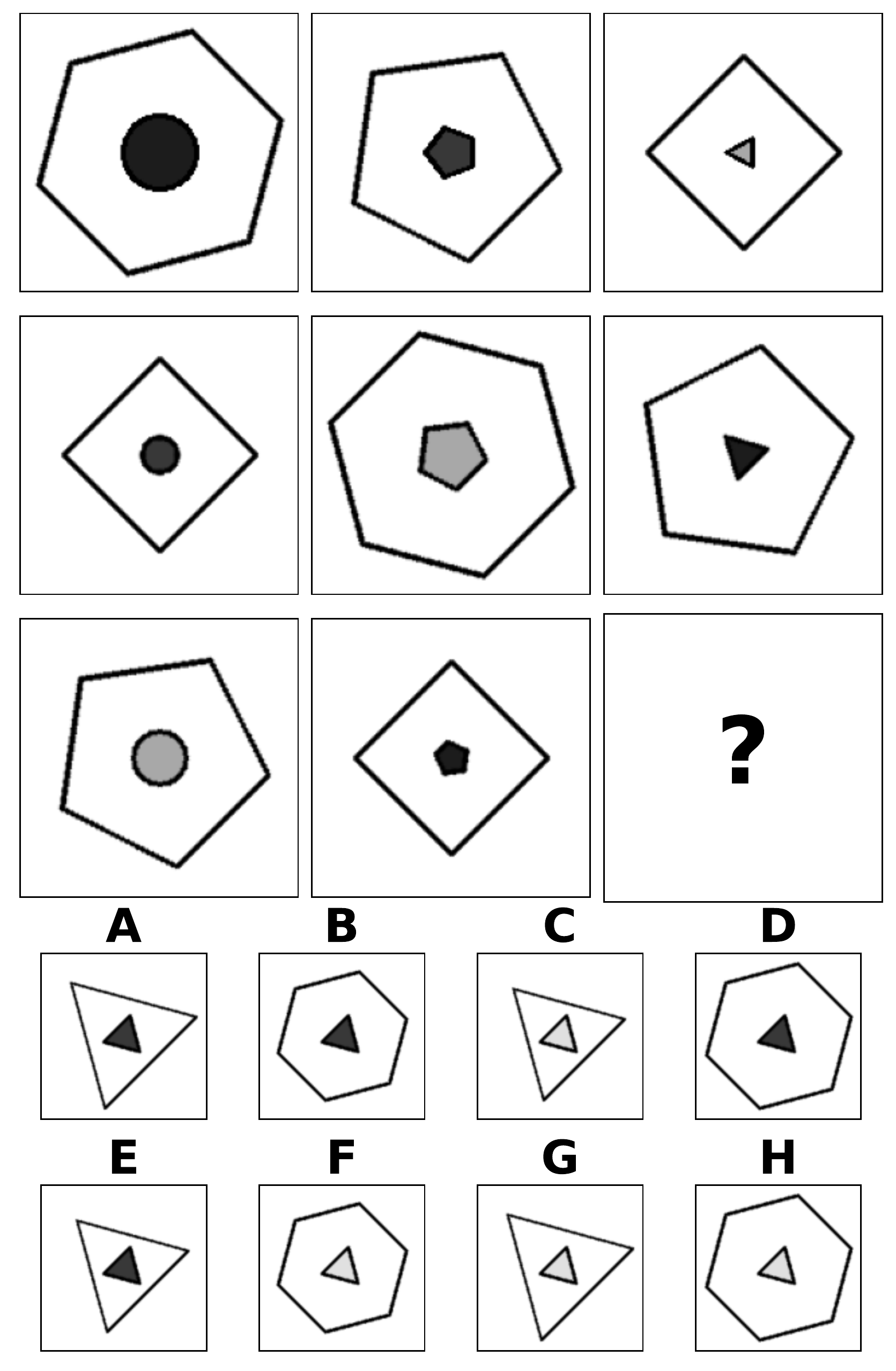}
            \caption{Base}
        \end{subfigure}
        ~
        \begin{subfigure}{.20\textwidth}
            \includegraphics[width=\textwidth]{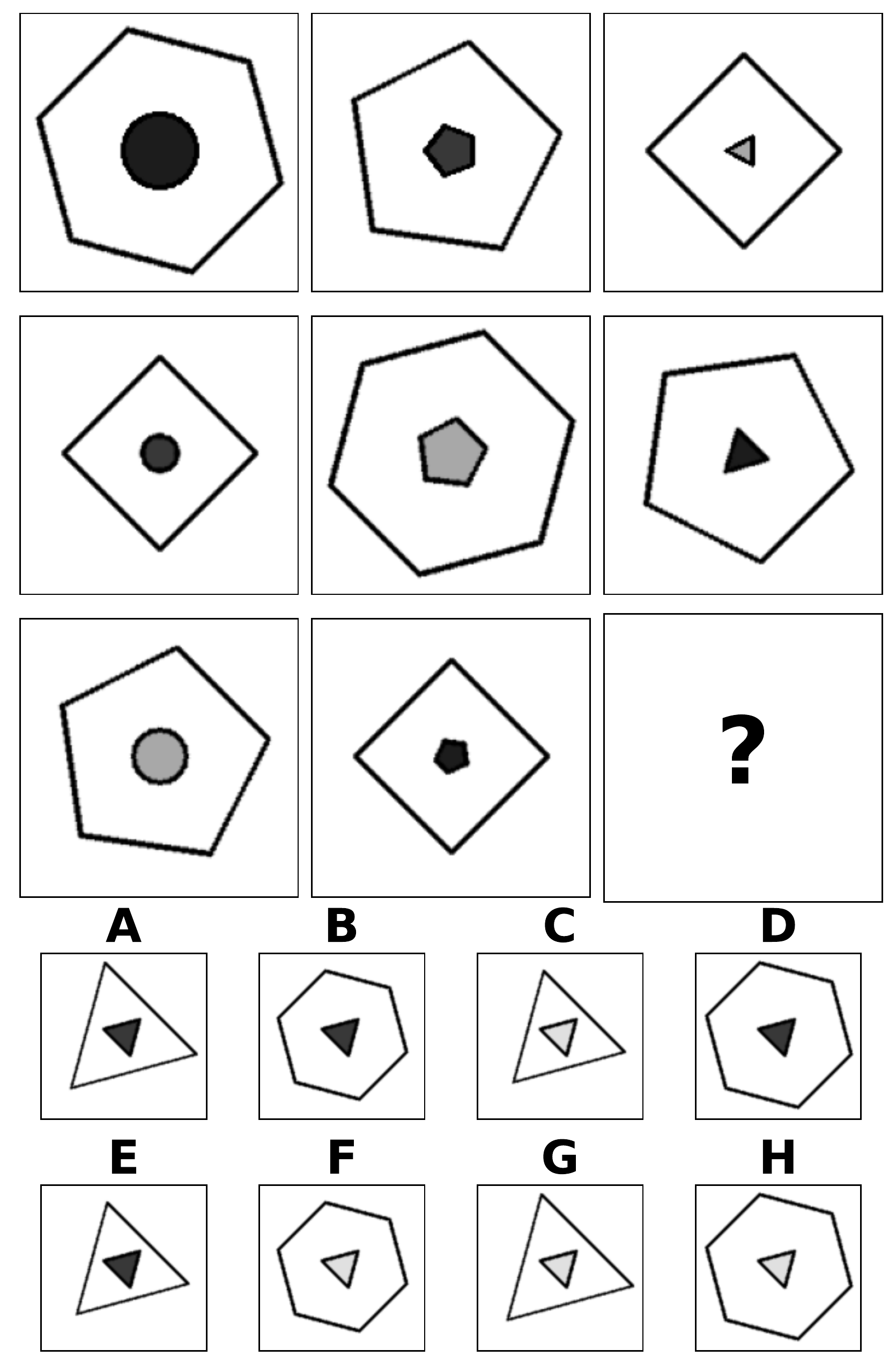}
            \caption{Vertical flip}
        \end{subfigure}
        ~
        \begin{subfigure}{.20\textwidth}
            \includegraphics[width=\textwidth]{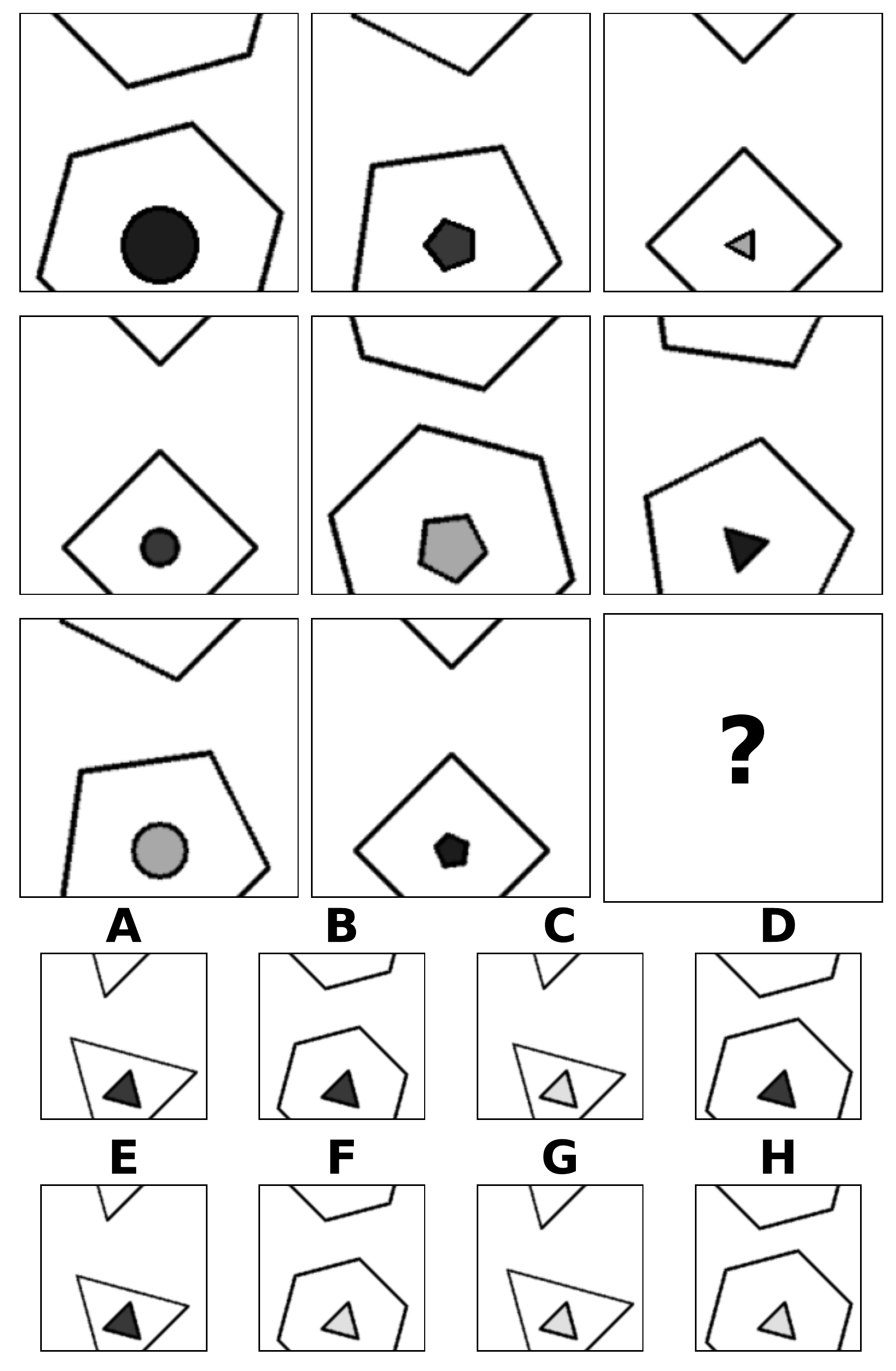}
            \caption{Vertical roll}
        \end{subfigure}
        ~
        \begin{subfigure}{.20\textwidth}
            \includegraphics[width=\textwidth]{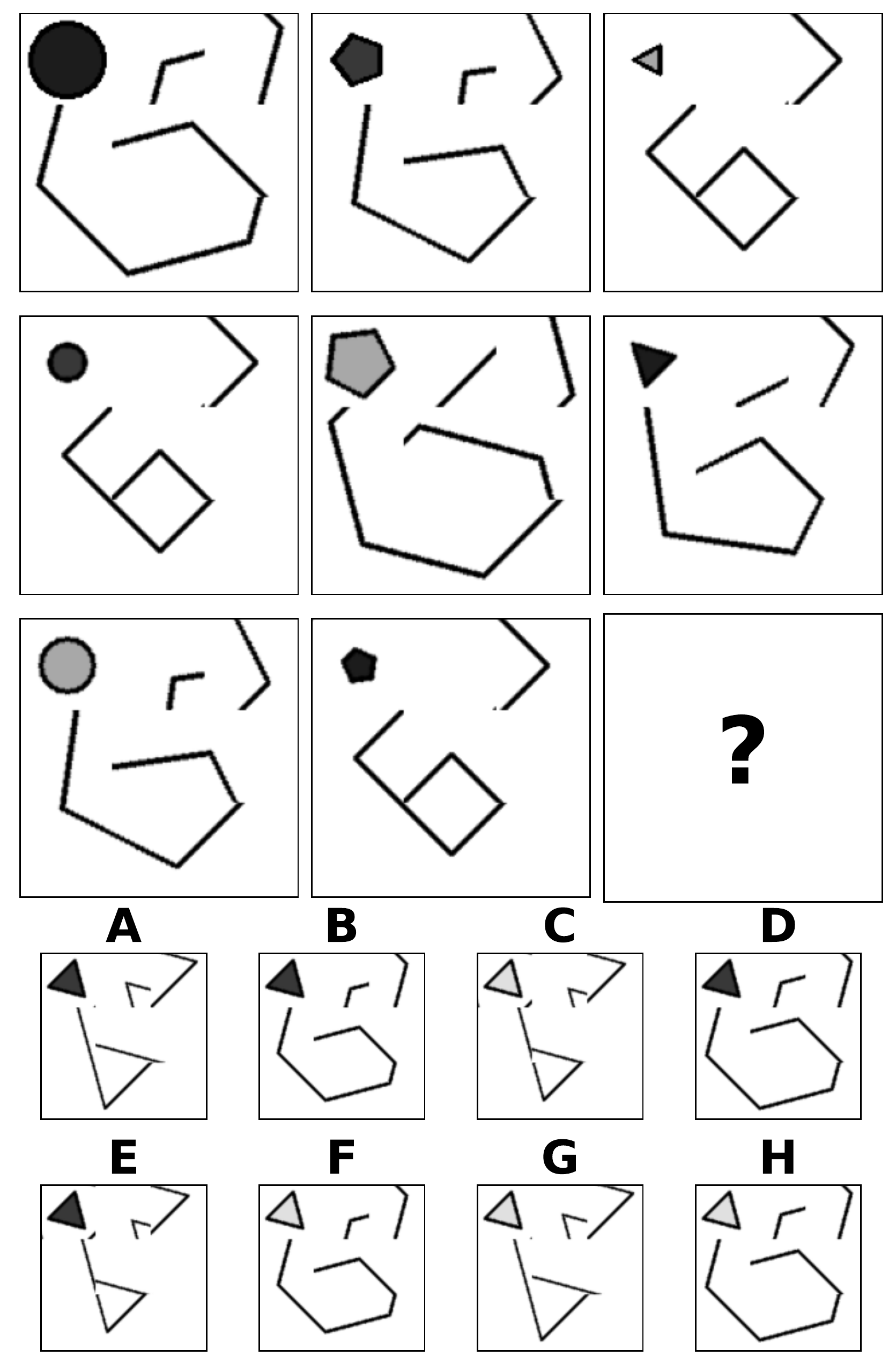}
            \caption{Shuffle 3x3}
        \end{subfigure}
        \caption{
        Augmented RPM from the Balanced-RAVEN dataset with configuration \texttt{O-IC}.
        In this configuration, both inner and outer structures can be governed by different sets of rules.
        Looking at the outer structures, one can see that each panel is composed of a single object ($[\texttt{Constant},\texttt{Number}]$) at the central position ($[\texttt{Constant},\texttt{Position}]$) of the same color ($[\texttt{Constant},\texttt{Color}]$).
        In each row, the outer structures have one out of three possible sizes and types -- hexagon, pentagon or a square ($[\texttt{Distribute\_Three},\texttt{Size}]$ and $[\texttt{Distribute\_Three},\texttt{Type}]$).
        This gives an outer structure defined as $\mathcal{S}_{\mathrm{outer}}=\{[\texttt{Constant},\texttt{Number}],$ $[\texttt{Constant},\texttt{Position}],$ $[\texttt{Distribute\_Three},\texttt{Type}],$ $[\texttt{Distribute\_Three},\texttt{Size}],$ $[\texttt{Constant},\texttt{Color}]\}$.
        Similarly, the inner structure of each image consists of single objects ($[\texttt{Constant},\texttt{Number}]$) located in the centre ($[\texttt{Constant},\texttt{Position}]$).
        The objects differ in type, with circles in the left column, pentagons in the middle one and triangles in the right column ($[\texttt{Progression},\texttt{Type}]$).
        The objects in inner structures of each row have 3 different sizes and colors ($[\texttt{Distribute\_Three},\texttt{Size}]$ and $[\texttt{Distribute\_Three},\texttt{Color}]\}$), which leads to a final outer structure $\mathcal{S}_{\mathrm{outer}}=\{[\texttt{Constant},\texttt{Number}],$ $[\texttt{Constant},\texttt{Position}],$ $[\texttt{Progression},\texttt{Type}],$ $[\texttt{Distribute\_Three},\texttt{Size}],$ $[\texttt{Distribute\_Three},\texttt{Color}]\}$.
        The only answer which matches all rules for both inner and outer structures is D.
        }
        \label{fig:augmentation-oic}
    \end{figure*}

\begin{figure*}[t]
        \centering
        \begin{subfigure}{.20\textwidth}
            \includegraphics[width=\textwidth]{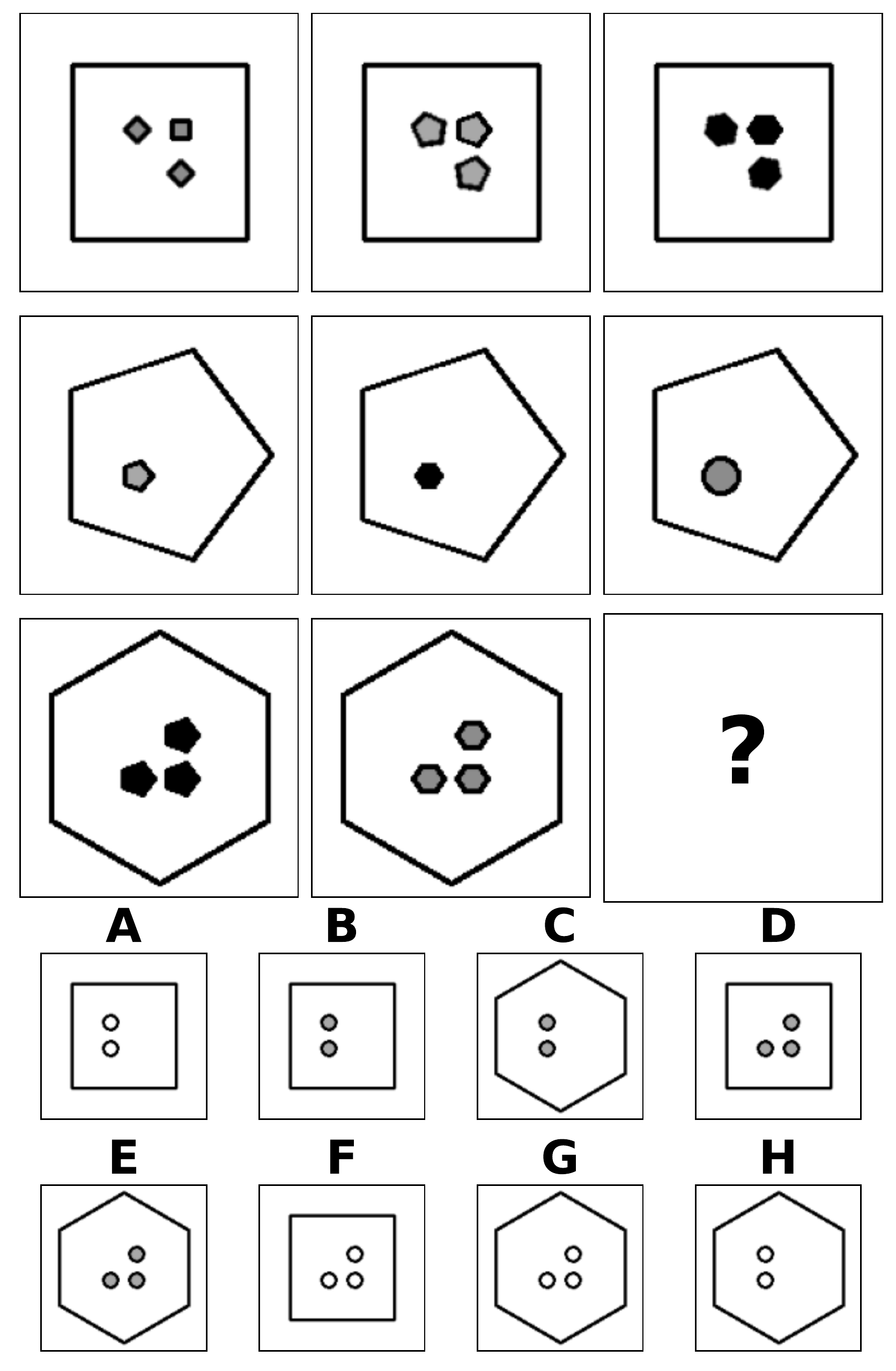}
            \caption{Base}
        \end{subfigure}
        ~
        \begin{subfigure}{.20\textwidth}
            \includegraphics[width=\textwidth]{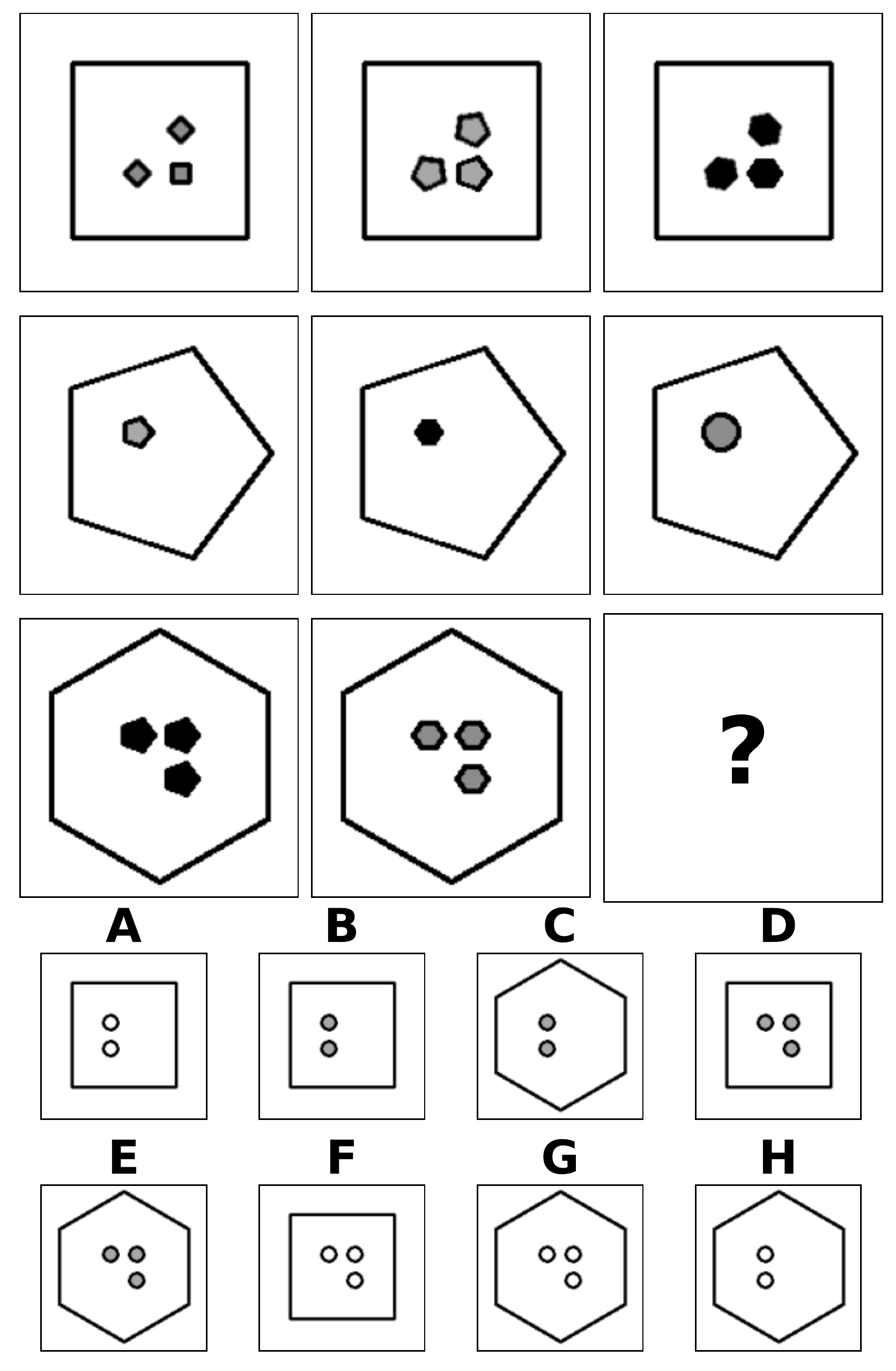}
            \caption{Vertical flip}
        \end{subfigure}
        ~
        \begin{subfigure}{.20\textwidth}
            \includegraphics[width=\textwidth]{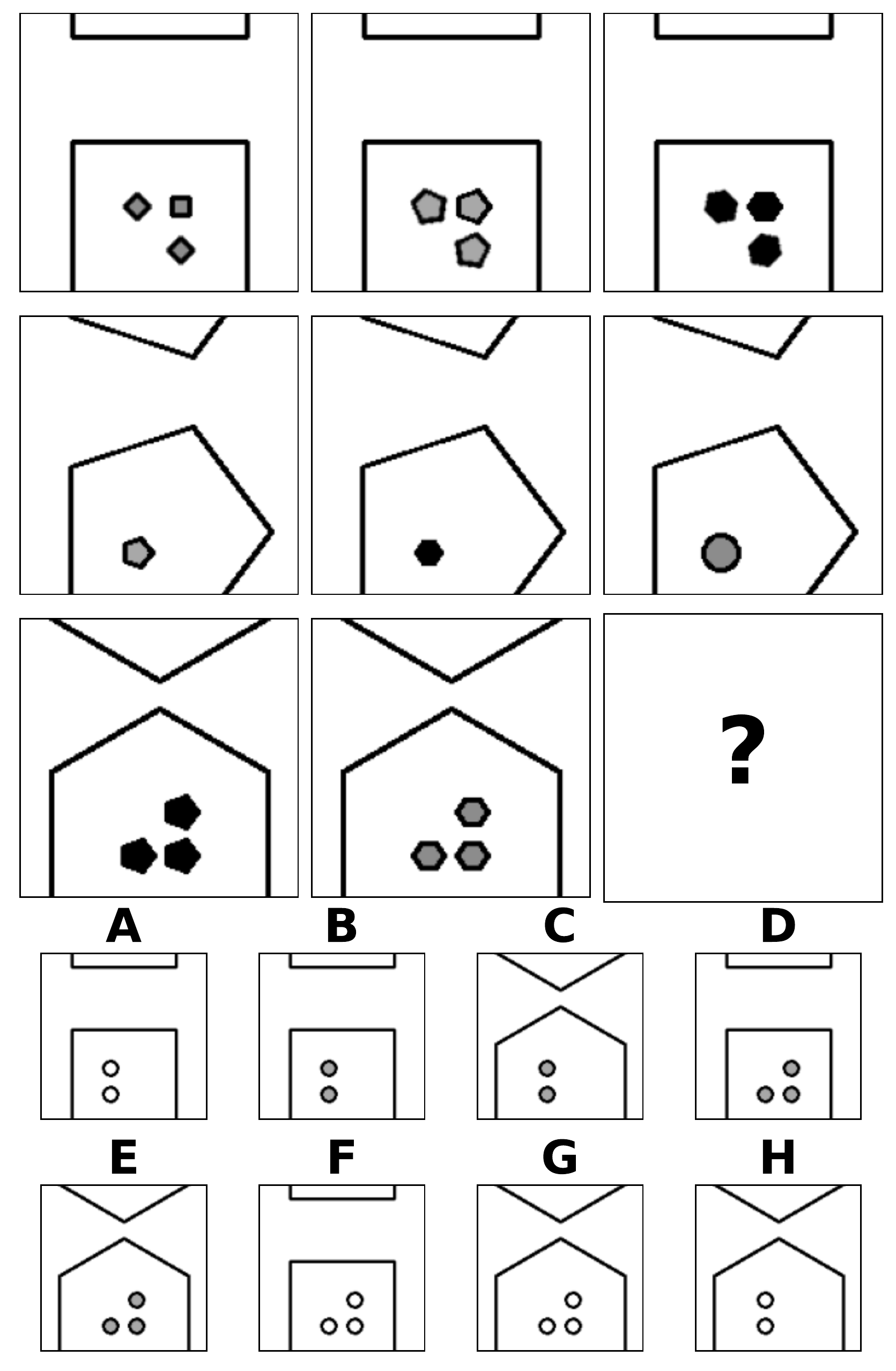}
            \caption{Vertical roll}
        \end{subfigure}
        ~
        \begin{subfigure}{.20\textwidth}
            \includegraphics[width=\textwidth]{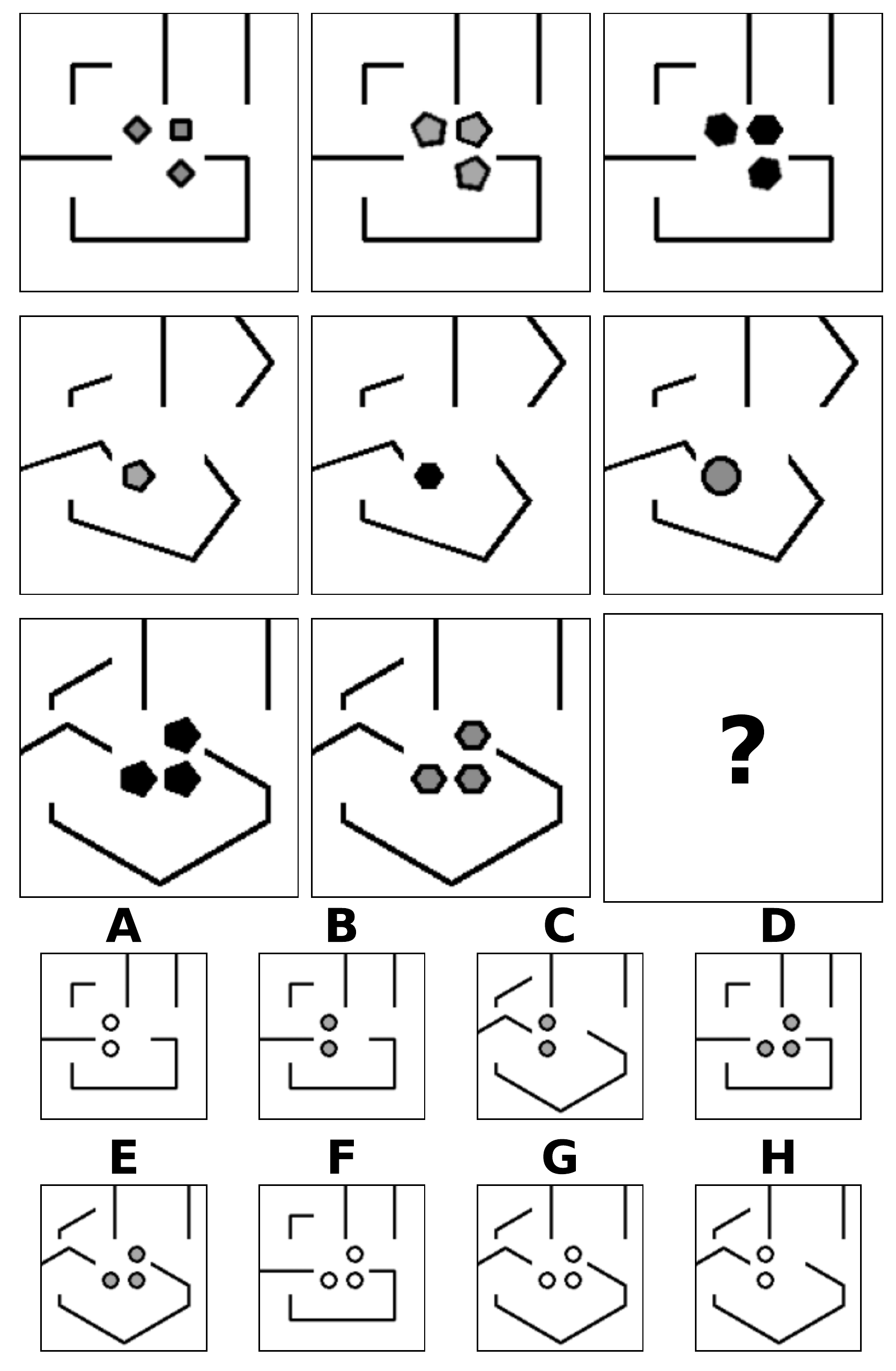}
            \caption{Shuffle 3x3}
        \end{subfigure}
        \caption{
        Augmented RPM from the Balanced-RAVEN dataset with configuration \texttt{O-IG}.
        Similarly to \texttt{O-IC} configuration, the rules of \texttt{O-IG} are applied to both inner and outer structures.
        Here, the outer structure is defined as $\mathcal{S}_{\mathrm{outer}}=\{[\texttt{Constant},\texttt{Number}], [\texttt{Constant},\texttt{Position}],$ $[\texttt{Constant},\texttt{Type}],$ $[\texttt{Constant},\texttt{Size}],$ $[\texttt{Constant},\texttt{Color}]\}$ and the inner structure as $\mathcal{S}_{\mathrm{inner}}=\{[\texttt{Constant},\texttt{Number}],$ $[\texttt{Constant},\texttt{Position}],$ $[\texttt{Progression},\texttt{Type}],$ $[\texttt{Distribute\_Three},\texttt{Size}],$ $[\texttt{Distribute\_Three},\texttt{Color}]\}$.
        E is the correct answer.
        }
        \label{fig:augmentation-oig}
    \end{figure*}

\begin{figure*}[t]
        \centering
        \begin{subfigure}{.20\textwidth}
            \includegraphics[width=\textwidth]{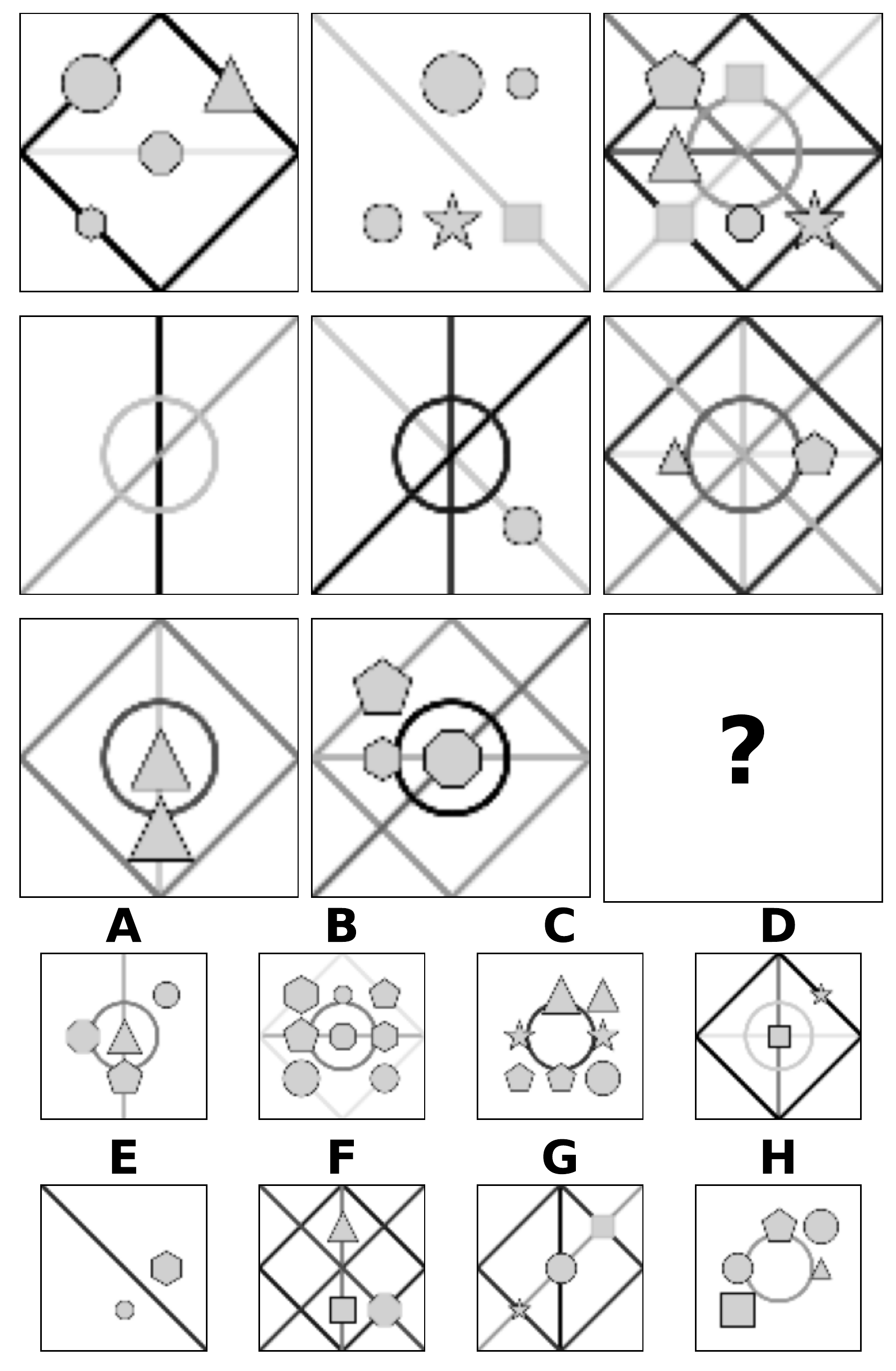}
            \caption{Base}
        \end{subfigure}
        ~
        \begin{subfigure}{.20\textwidth}
            \includegraphics[width=\textwidth]{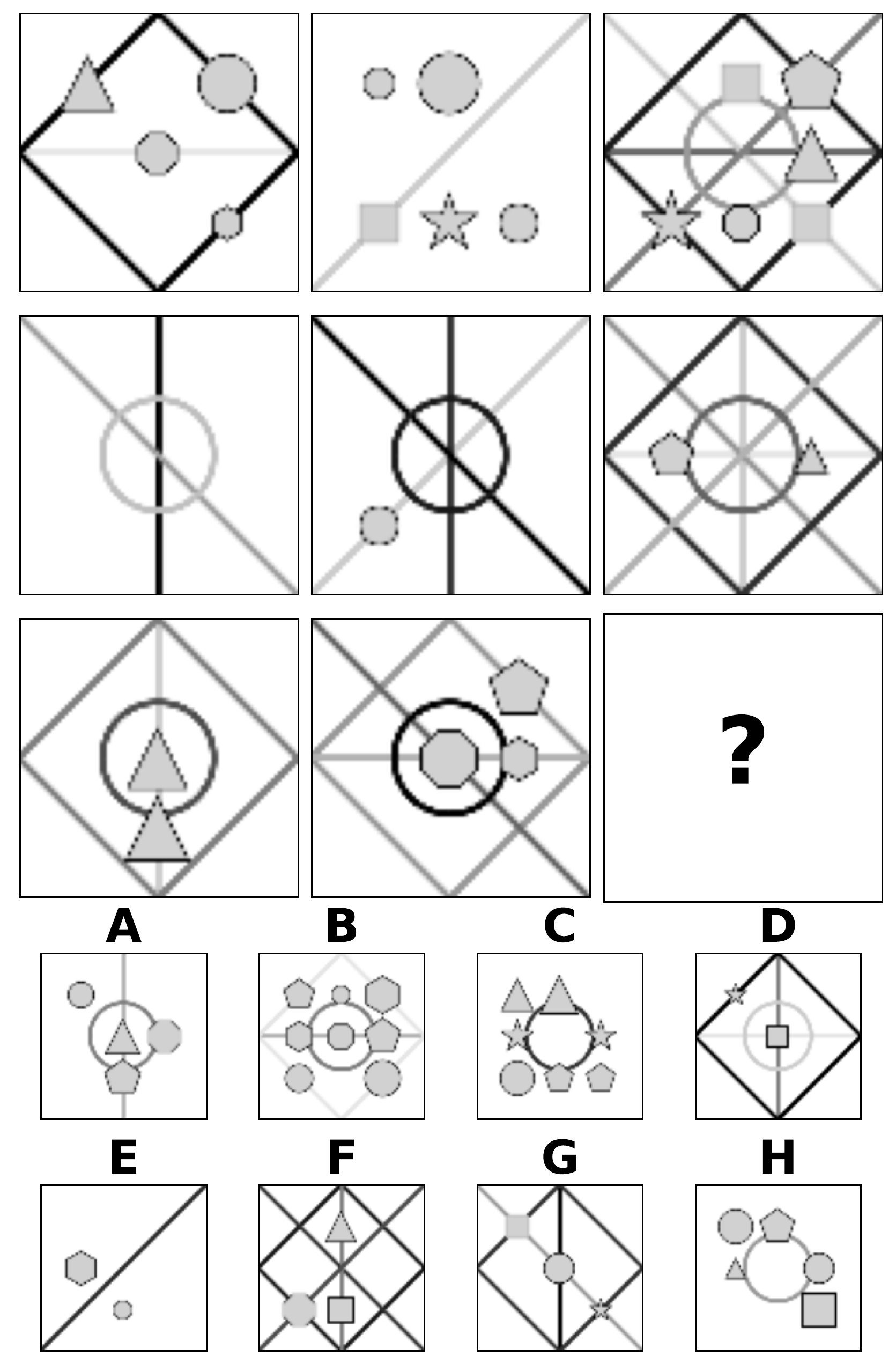}
            \caption{Horizontal flip}
        \end{subfigure}
        ~
        \begin{subfigure}{.20\textwidth}
            \includegraphics[width=\textwidth]{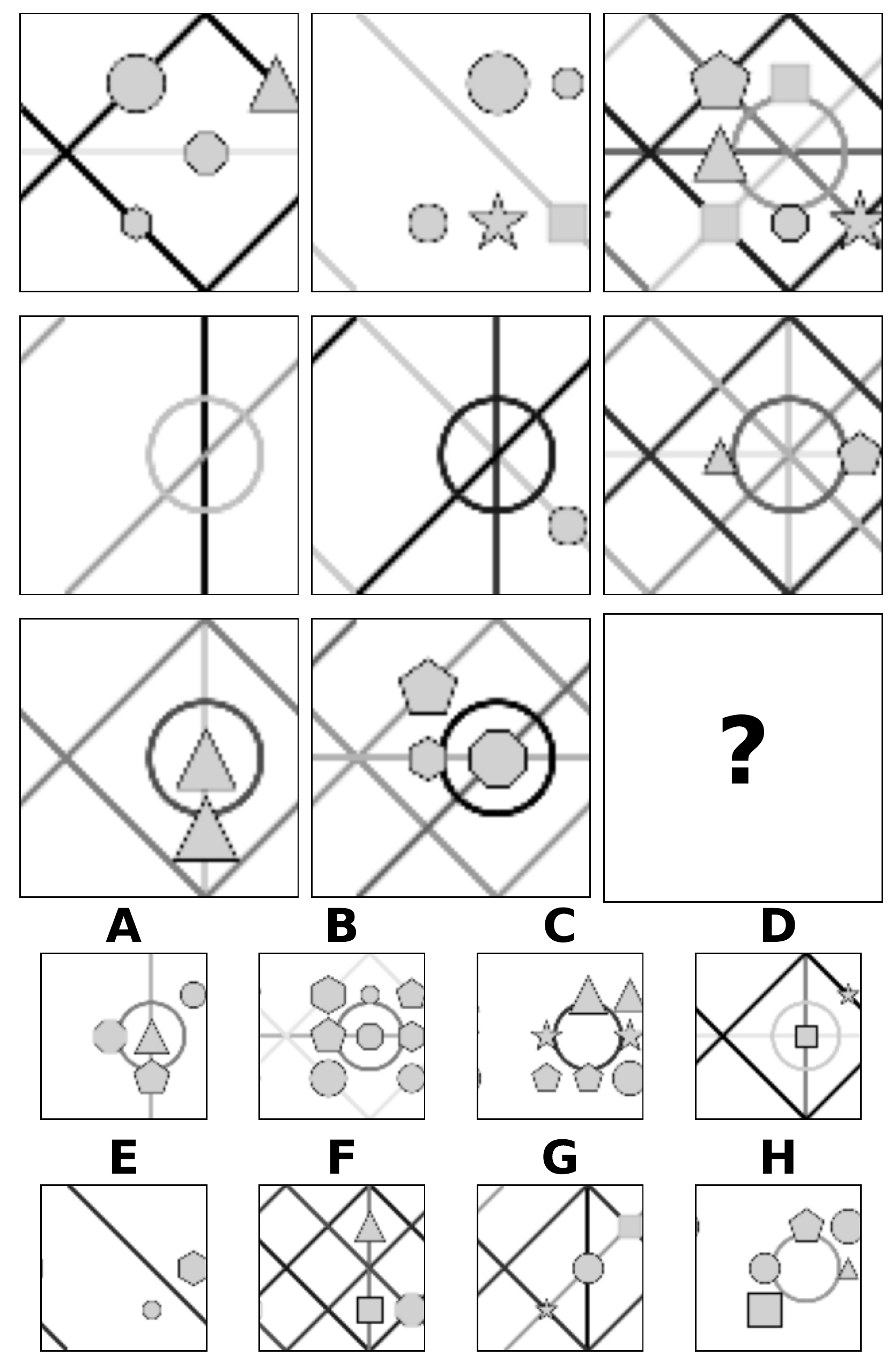}
            \caption{Horizontal roll}
        \end{subfigure}
        ~
        \begin{subfigure}{.20\textwidth}
            \includegraphics[width=\textwidth]{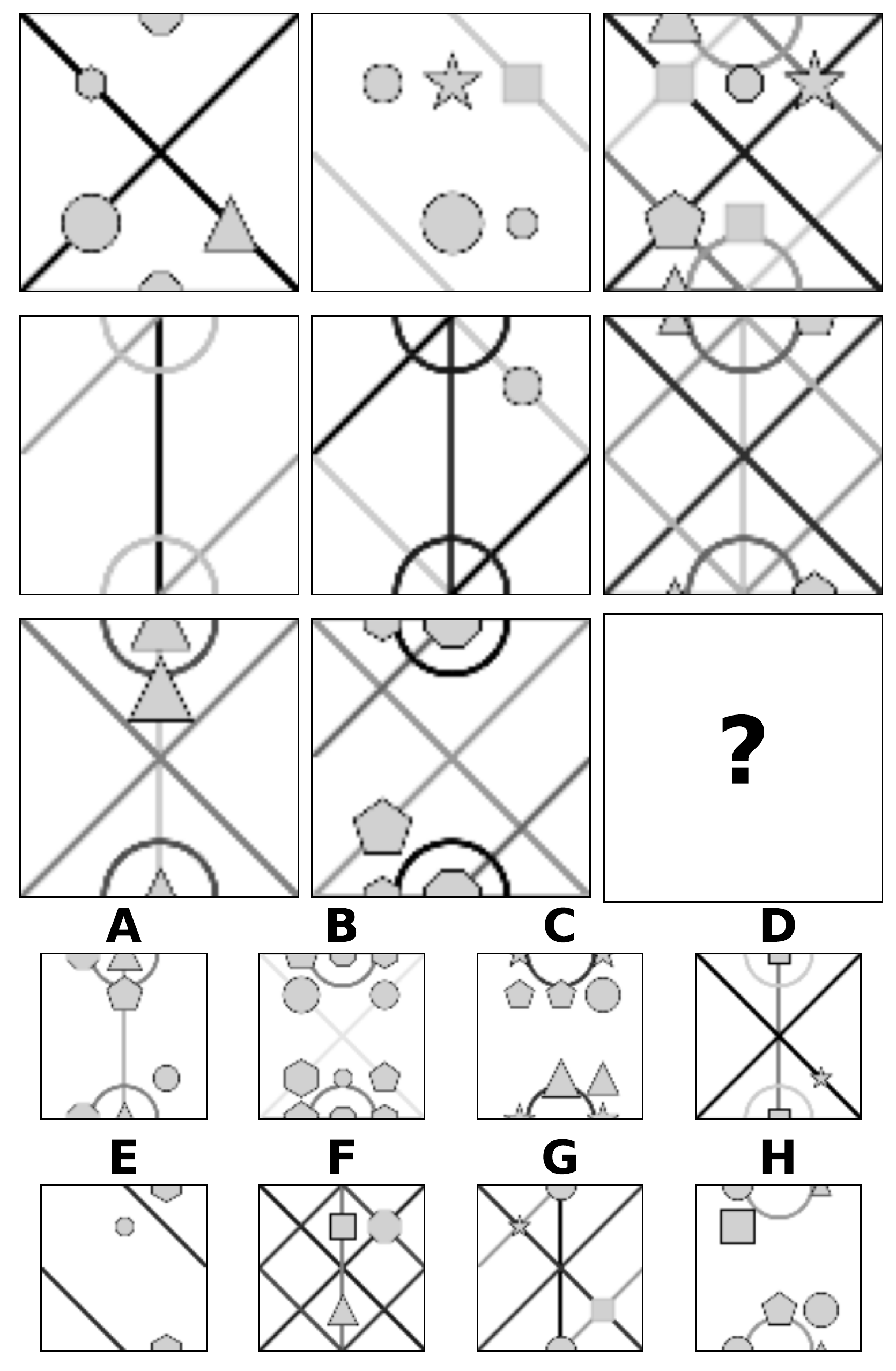}
            \caption{Shuffle 2x2}
        \end{subfigure}
        \caption{
        Augmented RPM from the PGM dataset.
        In each row, the number of shapes increases by one in each consecutive panel from left to right.
        Namely, the first row contains images with 4, 5 and 6 shapes, respectively, whereas the second row with 0, 1 and 2 shapes, respectively.
        Following this pattern, we expect the missing image in the bottom row to be composed of four shapes, which is realised by choosing the answer A.
        The underlying abstract structure is defined as $\mathcal{S}$ $=$ $\{[\texttt{shape},\texttt{number},\texttt{progression}]\}$.
        }
        \label{fig:augmentation-pgm}
    \end{figure*}

\begin{figure*}[t]
        \centering
        \begin{subfigure}{.20\textwidth}
            \includegraphics[width=\textwidth]{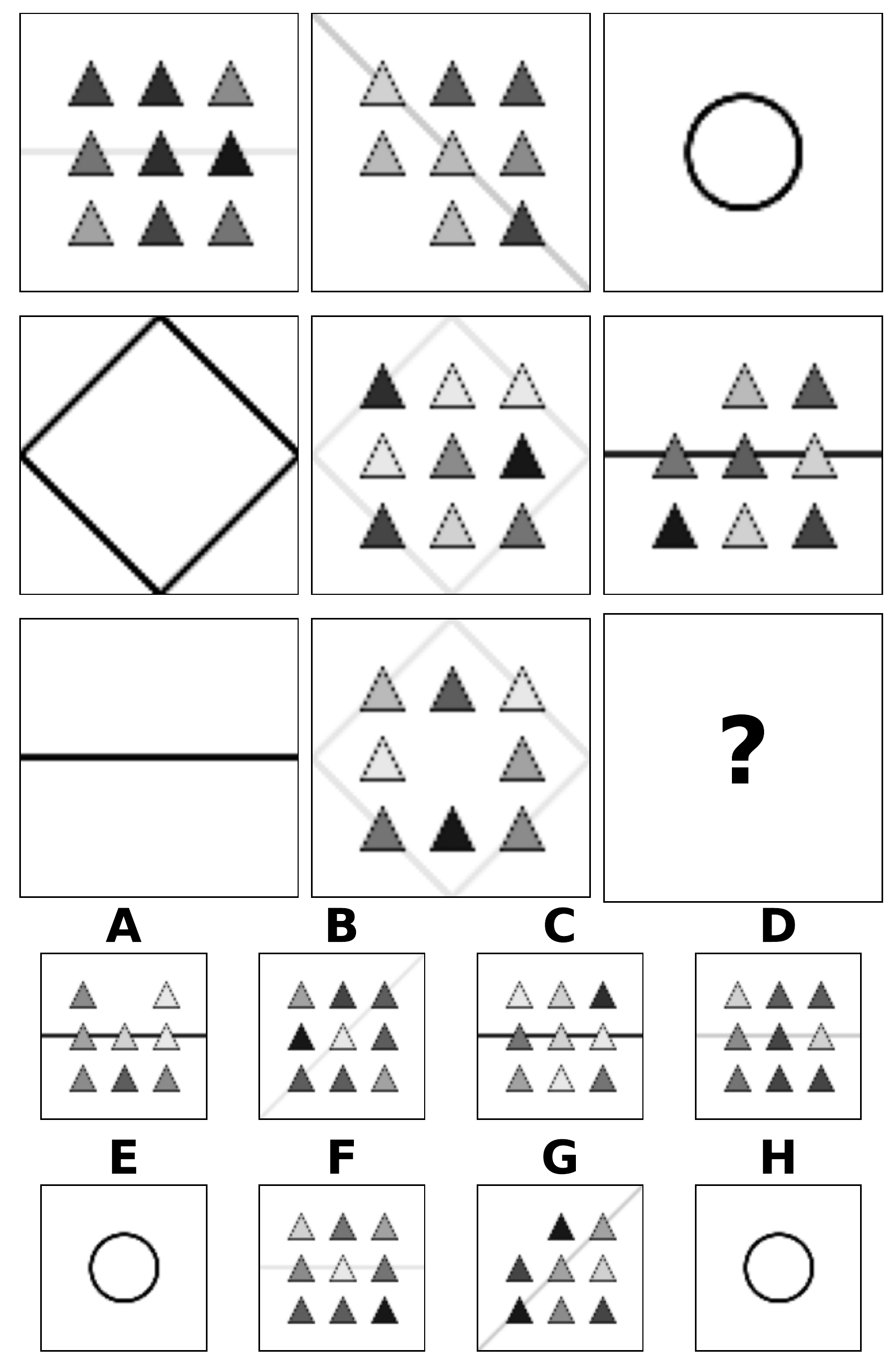}
            \caption{Base}
        \end{subfigure}
        ~
        \begin{subfigure}{.20\textwidth}
            \includegraphics[width=\textwidth]{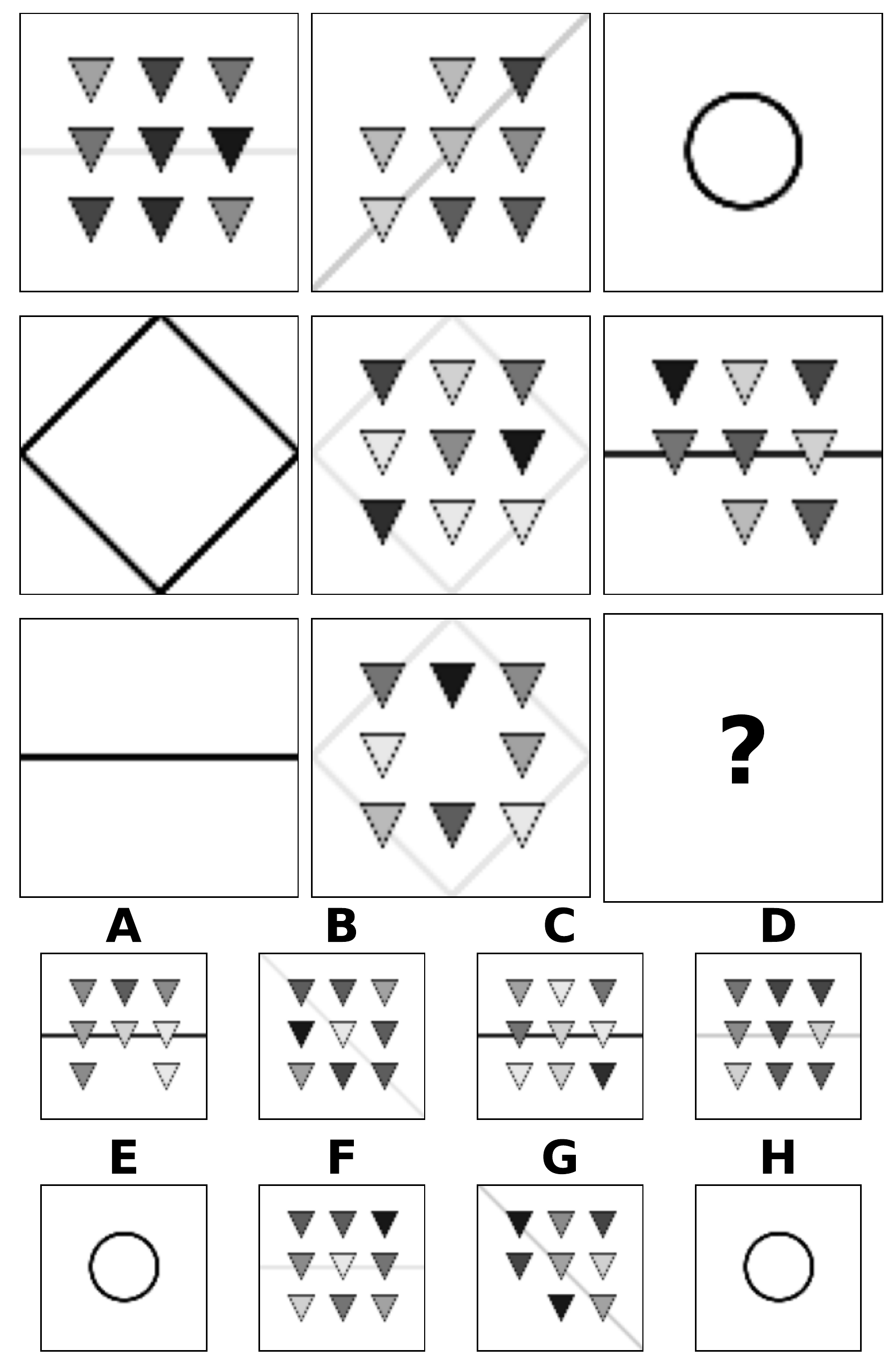}
            \caption{Vertical flip}
        \end{subfigure}
        ~
        \begin{subfigure}{.20\textwidth}
            \includegraphics[width=\textwidth]{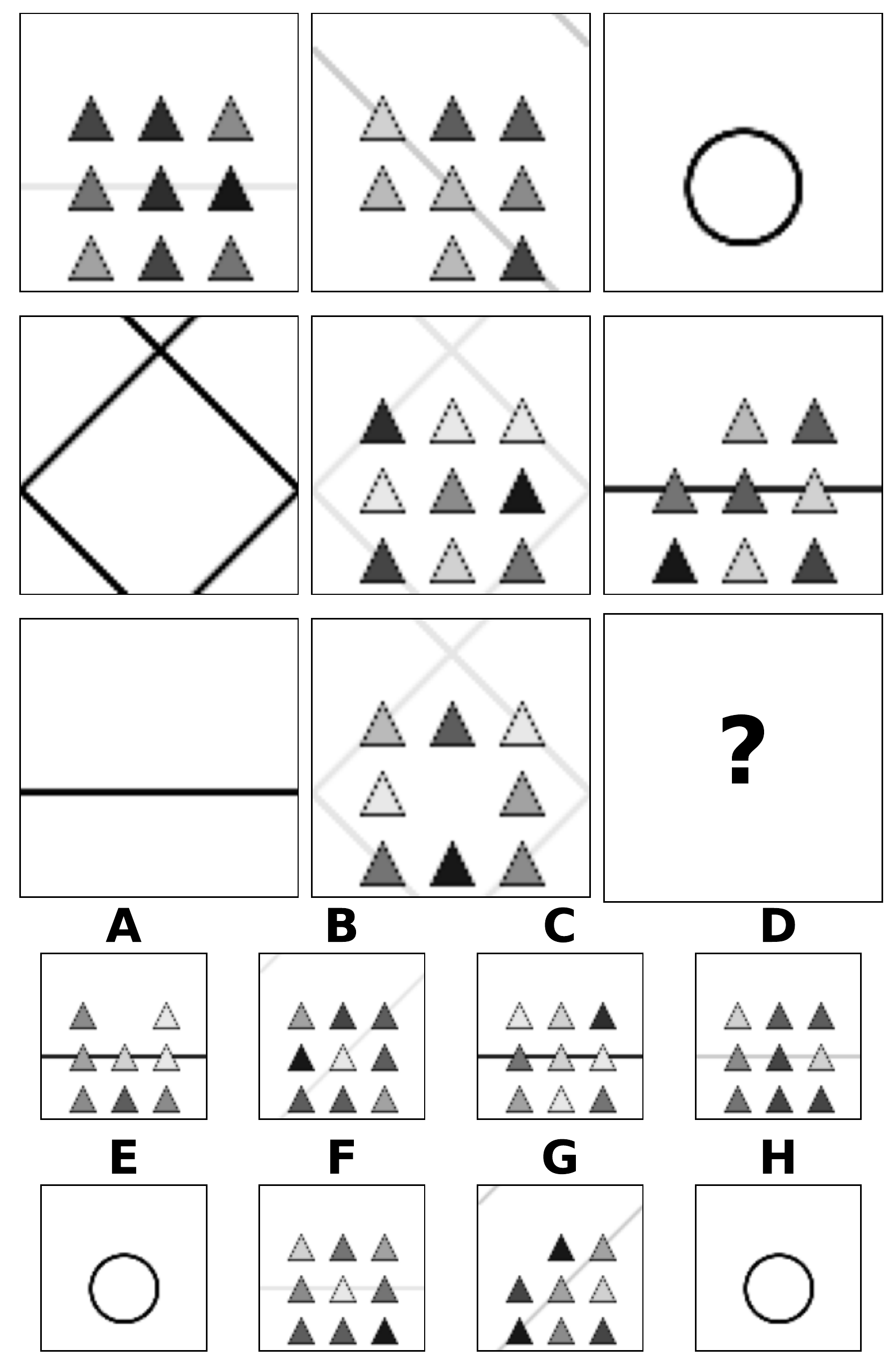}
            \caption{Vertical roll}
        \end{subfigure}
        ~
        \begin{subfigure}{.20\textwidth}
            \includegraphics[width=\textwidth]{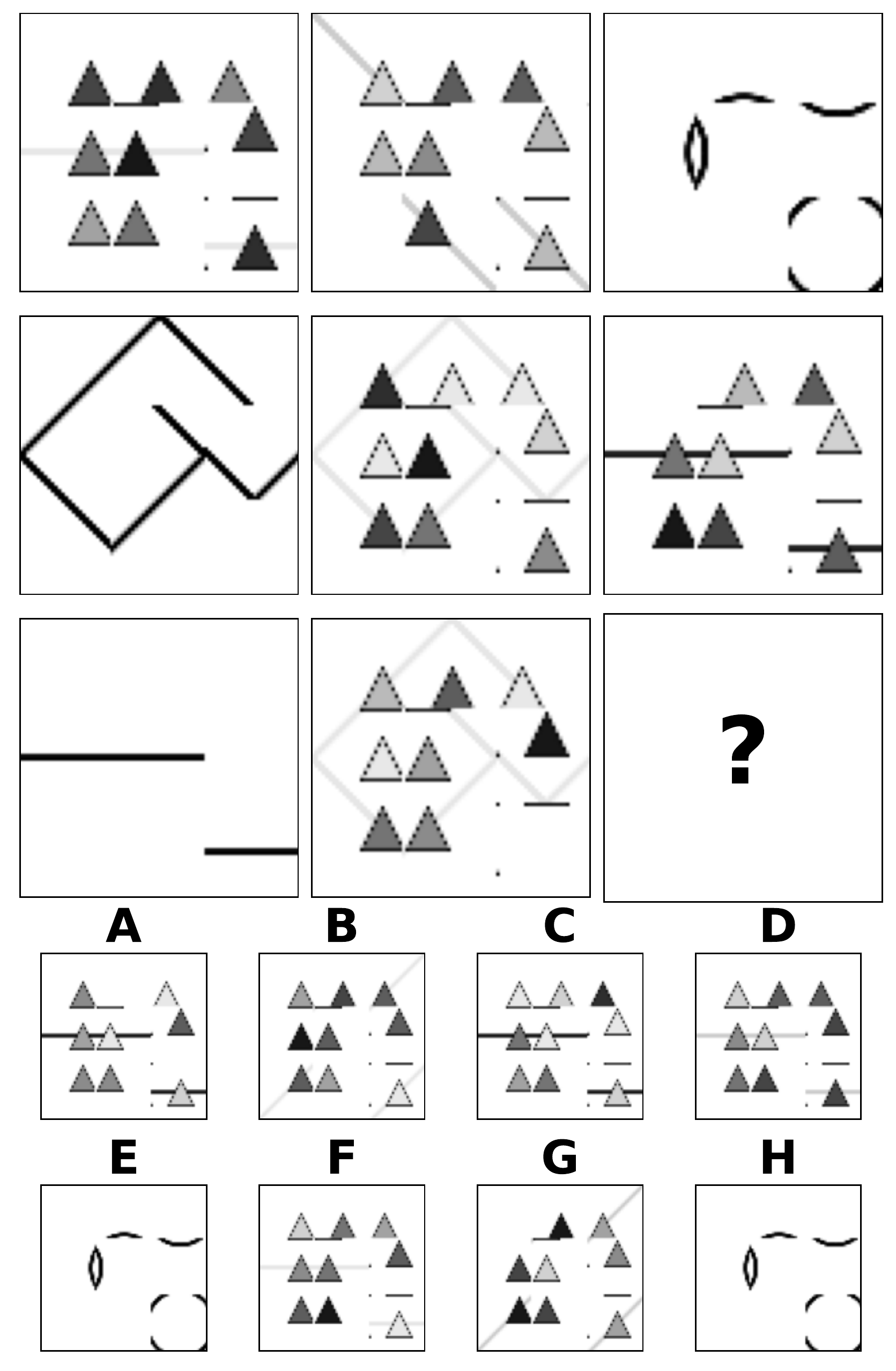}
            \caption{Shuffle 3x3}
        \end{subfigure}
        \caption{
        Augmented RPM from the PGM dataset, with two rules applied to two different object types -- shape and line.
        Firstly, note that the first two rows have consistent unions of shape numbers.
        That is, the first row contains panels with 9, 8 and 0 shapes, whereas the second row with 0, 9 and 8 shapes.
This suggests that the missing panel should contain 9 shapes.
However, there are multiple choices which satisfy this condition (B, C, D and F).
Therefore, it is necessary to further notice the consistent union of line colors in both completed rows.
        This results in an abstract structure for this RPM defined as $\mathcal{S}$ $=$ $\{[\texttt{shape},\texttt{number},\texttt{consistent union}],$ $[\texttt{line},\texttt{color},\texttt{consistent union}]\}$ and D being the correct answer.
        }
        \label{fig:augmentation-pgm2}
    \end{figure*}

    Examples of augmented RPMs are presented in Figs.~\ref{fig:augmentation-center}-\ref{fig:augmentation-pgm2}. Each figure, from left to right, presents (a) the base RPM from either Balanced-RAVEN or PGM, (b) the RPM after applying the horizontal/vertical flip augmentation, (c) the RPM after horizontal/vertical roll, and (d) the RPM after 2x2 or 3x3 grid shuffle.
    The biggest structural changes are introduced to the images after applying the roll operation (the 3rd column in each figure) and the grid shuffle transformation (the last column in each figure).
    Even though the images become more difficult to decipher for human solvers, they greatly diversify the training set and increase the overall number of training instances.

    \section{Implementation details}
    For the sake od speeding-up the experiments, in each hierarchy of HriNet we have replaced the ResNet backbone with a simpler CNN\@.
    This CNN was composed of 4 layers, each with 32 convolutional kernels of size 3x3 and stride equal to 2.
    Each layer was followed by batch normalization and ReLU activation.
    All hyperparameters reported in the paper were chosen without any extensive search, based on manual convergence analysis of a limited number of runs.

    In all experiments on both Balanced-RAVEN and PGM datasets, we have trained models on images rescaled to the size 80x80, following the setup from~\cite{santoro2018measuring}.
    This allowed to overcome hardware memory limitations, train with larger batches and conduct extensive comparisons.
    For the ease of reproducibility, in the source code attachment the experiments are structured as PyTorch Lightning modules.

    \section{Batch size}
    We have further validated the influence of batch sizes on the results of MLCL training \emph{with} and \emph{without} data augmentation (see Figure~\ref{fig:additional-ablations}).
    We have varied batch sizes in the contrastive pre-training stage and used a constant size of 128 for linear evaluation.

    The best performance for all models was achieved with small and middle-size batches, ranging from 32 to 128.
    We hypothesize that although larger batches allow for increasing the contrast strength of the proposed Multi-Label Contrastive Loss (eq.~(8) in the paper), at the same time they hinder the convergence ability of the auxiliary loss.

    This empirical evaluation stresses the importance of balanced interplay between the auxiliary and contrastive losses in our learning framework.

    \section{Loss coefficients}
    In all experiments reported in the paper we set $\gamma=1$ and $\beta=10$ as the balancing factors used for the Multi-Label Contrastive Loss (eq.~(8)).
    In order to verify the accuracy of this choice, we run additional experiments with $\gamma$ fixed to 1 and various values of $\beta$.
    As presented in Figure~\ref{fig:additional-ablations}, a correct choice of these hyperparameters is critical for the final performance. In the experiments without data augmentation, too small, as well as too big values of $\beta$ have a negative impact on the accuracy of all models and the best results are achieved with $\beta=10$. When data augmentation is applied, small values of $\beta$ once again hinder the performance, however, the results for bigger values of $\beta$ are less diverse than in the former case, and depending on the model the best results are accomplished with either $\beta=10$ or $\beta=15$.

\end{document}